%% file: arxiv.tex
\documentclass[sigconf,nonacm]{aamas}
\input{preamble}
\title{\texorpdfstring{\AkuleMasterLogo\methodname: Fast and Scalable Multi-Robot Motion Planning via Sparse Interaction Diffusion}{Akule: Fast and Scalable Multi-Robot Motion Planning via Sparse Interaction Diffusion}}

\author{Manav Vora}
\author{Gokul Puthumanaillam}
\author{Aditya Penumarti}
\author{Jane Shin}
\author{Melkior Ornik}
\begin{abstract}
Diffusion models provide expressive trajectory priors for
multi-robot motion planning (MRMP), but scaling them to large
teams requires coordinating individual motions without
expensive joint computation. We introduce
\AkuleInlineLogo\method, a sparse interaction diffusion
framework that augments a frozen single-robot prior with
three learned modules: a reusable pairwise correction model
$R_{\phi}$, a contextual selector $U_{\psi}$ that determines
which corrections to evaluate, and a composer $G_{\omega}$
that combines their effects through context-dependent signed
weights. This decomposition separates which interactions
merit computation from how their effects should be combined
into coordinated denoising updates. We train $R_{\phi}$
using matched pairwise trajectory optimizations, then learn
$G_{\omega}$ and $U_{\psi}$ from physically optimized
joint trajectories without requiring a prescribed
ground-truth interaction graph. The resulting joint
proposals are passed to a downstream search-based planner
for validation and repair. Across diverse MRMP benchmarks,
\method achieves competitive planning success,
complete-planner speedups of up to $25\times$, zero-shot
population transfer, and successful planning with up to
$256$ robots. Ablations further demonstrate that learned
selection and composition reduce pre-repair conflicts
and downstream repair effort relative to heuristic
selection and uniform summation.
Project website: \href{https://akule-diff.github.io/}{\textcolor{akBlue}{https://akule-diff.github.io/}}
\end{abstract}
\hypersetup{keeppdfinfo=true,pdfcreator={LaTeX},pdfsubject={Multi-robot motion planning},pdfauthor={Manav Vora, Gokul Puthumanaillam, Aditya Penumarti, Jane Shin, Melkior Ornik},pdfkeywords={Diffusion Models, Multi-Agent Path Planning, Robotics}}
\begin{document}
\maketitle
\pagestyle{standardpagestyle}
\hypersetup{keeppdfinfo=true,pdfcreator={LaTeX},pdfsubject={Multi-robot motion planning},pdfauthor={Manav Vora, Gokul Puthumanaillam, Aditya Penumarti, Jane Shin, Melkior Ornik},pdfkeywords={Diffusion Models, Multi-Agent Path Planning, Robotics}}
\input{introduction}
\input{related}
\input{problem_formulation}
\input{method}
\input{experiments}
\bibliographystyle{ACM-Reference-Format}
\bibliography{references}
\clearpage
\appendix
\twocolumn[\section*{Appendix}]
\input{supplement/sections/architecture}
\input{supplement/sections/learning}
\input{supplement/sections/protocol}
\input{supplement/sections/quantitative}
\input{supplement/sections/ablations}
\input{supplement/sections/analysis}

\FloatBarrier
\input{supplement/sections/qualitative}

\FloatBarrier
\input{supplement/sections/additional}
\FloatBarrier
\end{document}

%% file: preamble.tex
\usepackage{booktabs}
\usepackage{xspace}
\usepackage{xcolor}

\usepackage{colortbl}
\definecolor{akBlue}{HTML}{1850A0}
\definecolor{akAccent}{HTML}{5888E0}
\definecolor{akGroup}{HTML}{B3C8EE}
\definecolor{akRule}{HTML}{DFE5EF}
\definecolor{akTint}{HTML}{EAF1FC}
\newcommand{\akH}[1]{\textcolor{akBlue}{\bfseries\boldmath #1}}
\newcommand{\akM}[1]{\textcolor{akBlue}{\bfseries #1}}
\newcommand{\akOurs}{\rowcolor{akTint}}
\newcommand{\akLine}{\specialrule{0.25pt}{0pt}{0pt}}
\newcommand{\akGroupLine}{\arrayrulecolor{akGroup}\specialrule{0.5pt}{0pt}{0pt}\arrayrulecolor{akRule}}
\newcommand{\akAccentLine}{\arrayrulecolor{akAccent}\specialrule{0.65pt}{0pt}{0pt}\arrayrulecolor{akRule}}
\newcommand{\akStyle}{\sffamily\ifdim\linewidth<\textwidth\fontsize{6}{7.2}\else\fontsize{7}{8.4}\fi\selectfont\renewcommand{\arraystretch}{1.16}\setlength{\aboverulesep}{0pt}\setlength{\belowrulesep}{0pt}}
\newlength{\akSlack}
\newsavebox{\akBoxA}
\newsavebox{\akBoxB}
\newcommand{\akFit}[1]{\arrayrulecolor{akRule}\setlength{\akSlack}{0pt}\setlength{\tabcolsep}{0pt}\sbox{\akBoxA}{#1}\setlength{\tabcolsep}{1pt}\sbox{\akBoxB}{#1}\setlength{\tabcolsep}{\dimexpr(\linewidth-\wd\akBoxA)/\numexpr(\wd\akBoxB-\wd\akBoxA)/65536\relax\relax}\ifdim\tabcolsep>6pt\setlength{\tabcolsep}{6pt}\fi
  \ifdim\tabcolsep<1.5pt\setlength{\tabcolsep}{1.5pt}\fi
  \setlength{\akSlack}{\dimexpr\linewidth-\wd\akBoxA-\tabcolsep*\numexpr(\wd\akBoxB-\wd\akBoxA)/65536\relax\relax}\ifdim\akSlack<0pt\setlength{\akSlack}{0pt}\fi
  #1\arrayrulecolor{black}}

\AtEndPreamble{\hypersetup{
    colorlinks=true,
    citecolor = akBlue,
    linkcolor=akBlue,
    urlcolor=black,
    }}
\newcommand{\methodname}{Akule}
\newcommand{\method}{\methodname\xspace}

\newcommand{\AkuleInlineLogo}{\raisebox{-0.20em}{\includegraphics[height=1.15em]{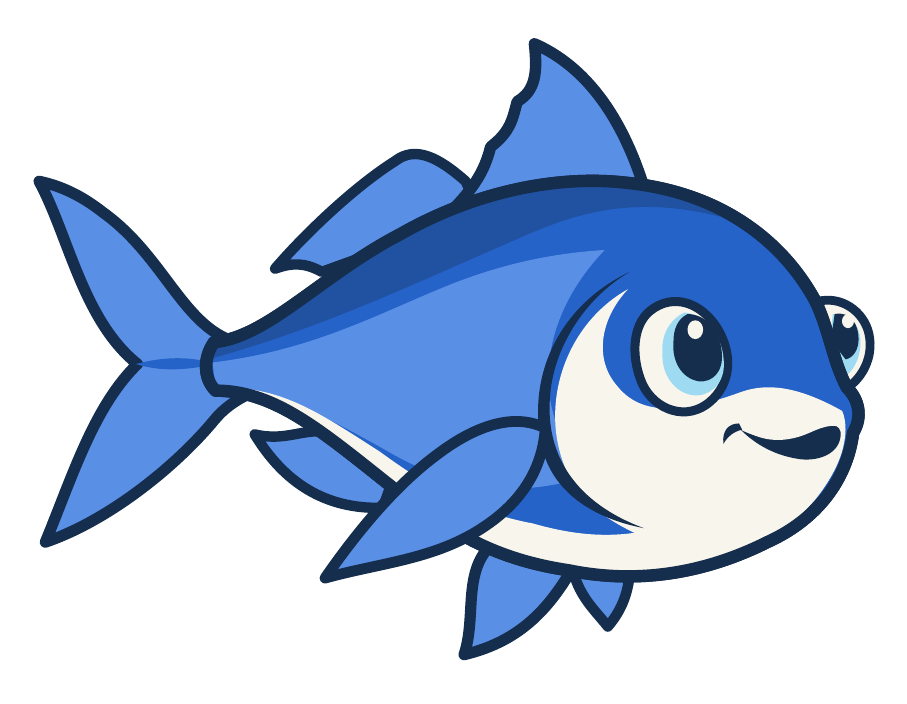}}\hspace{3pt}}

\newcommand{\AkuleMasterLogo}{\raisebox{-0.25em}{\includegraphics[height=1.20em]{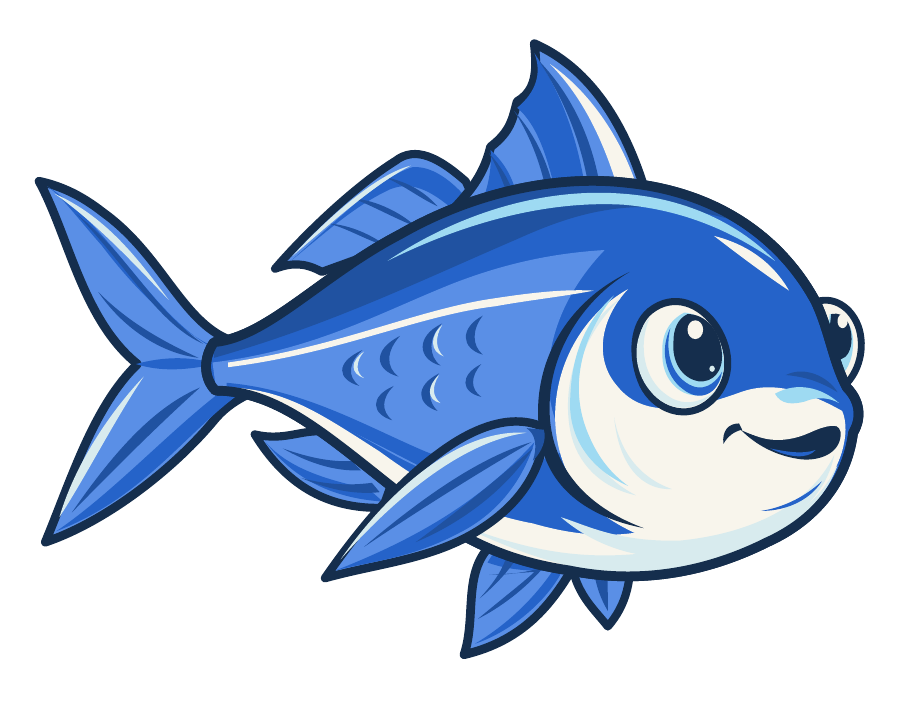}}\hspace{3pt}}

\newcommand{\akQualOther}[1]{{\setlength{\fboxsep}{1.5pt}\setlength{\fboxrule}{0.9pt}\fcolorbox{white}{white}{\includegraphics[width=\akPanel]{#1}}}}
\newcommand{\akQualOurs}[1]{{\setlength{\fboxsep}{1.5pt}\setlength{\fboxrule}{0.9pt}\fcolorbox{akBlue}{akTint}{\includegraphics[width=\akPanel]{#1}}}}
\newcommand{\akQualFail}{{\bfseries\color{red!80!black}[FAIL]}}

\newcommand{\FloatBarrier}{\clearpage}
\makeatletter
\AtBeginDocument{
\fancypagestyle{standardpagestyle}{
  \fancyhf{}
  \fancyfoot[C]{\normalfont\footnotesize\thepage}

}
\fancypagestyle{firstpagestyle}{
  \fancyhf{}
  \fancyfoot[C]{\normalfont\footnotesize\thepage}

}
}
\newcommand{\UIUCMark}{\raisebox{0.15em}{\includegraphics[height=0.72em]{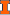}}}
\newcommand{\UFMark}{\raisebox{0.15em}{\includegraphics[height=0.58em]{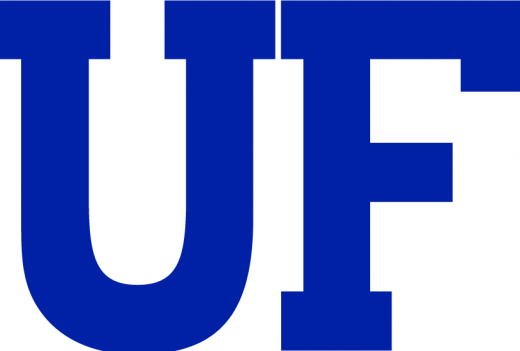}}}
\newcommand{\SNUMark}{\raisebox{0.08em}{\includegraphics[height=0.85em]{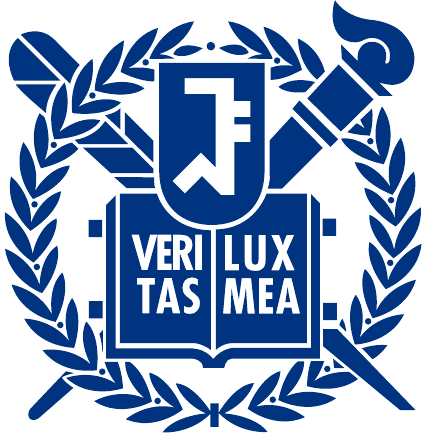}}}
\def\@mkauthors{\begingroup
  \hsize=\textwidth
  \global\setbox\mktitle@bx=\vbox{\noindent
    \unvbox\mktitle@bx\par\medskip
    \centering
    {\large\sffamily\bfseries
    Manav Vora\,\UIUCMark\quad
    Gokul Puthumanaillam\,\UIUCMark\quad
    Aditya Penumarti\,\UFMark\par
    Jane Shin\,\SNUMark\quad
    Melkior Ornik\,\UIUCMark\par}
    \smallskip
    {\normalsize
    \UIUCMark\,University of Illinois Urbana-Champaign\quad
    \UFMark\,University of Florida\par
    \SNUMark\,Seoul National University\par}
    \smallskip
    {\small
    \href{mailto:mkvora2@illinois.edu}{mkvora2@illinois.edu}\quad
    \href{mailto:gokulp2@illinois.edu}{gokulp2@illinois.edu}\quad
    \href{mailto:apenumarti@ufl.edu}{apenumarti@ufl.edu}\par
    \href{mailto:janeshin@snu.ac.kr}{janeshin@snu.ac.kr}\quad
    \href{mailto:mornik@illinois.edu}{mornik@illinois.edu}\par}
    \bigskip}
  \endgroup}
\makeatother

%% file: introduction.tex
\section{Introduction}
\label{sec:introduction}

Diffusion-based trajectory planners learn expressive motion priors from
offline data and generate goal-conditioned trajectories by iteratively
denoising entire motion sequences
\cite{janner2022diffuser,carvalho2023mpd}. Extending these models to
multi-robot motion planning (MRMP) introduces a coordination
challenge: \emph{trajectories that are individually plausible may become
infeasible when planned together}. Avoiding inter-robot collisions
therefore requires accounting for interactions between robots while
generating jointly compatible motions
\cite{shaoul2025mmd,liang2025smd}.

Existing diffusion-based MRMP methods introduce coordination
at different stages. Multi-robot Multi-model planning Diffusion
(MMD) retains single-robot priors and uses conflict-based
search to regenerate constrained trajectories
\cite{shaoul2025mmd}. Simultaneous MRMP Diffusion (SMD)
enforces inter-robot constraints during joint sampling
\cite{liang2025smd}, while Discrete-Guided Diffusion
(DGD) uses discrete multi-robot paths to guide constrained
generation \cite{liang2025dgd}. These approaches coordinate
through constrained replanning, joint optimization, or
discrete guidance. Yet interaction relevance can differ
across robot pairs and change as planned trajectories
evolve. This motivates our question: \emph{can diffusion
allocate expensive interaction reasoning to the pairs
that matter during denoising?}

One way to achieve this is to retain a single-robot motion prior
and represent the influence of another robot as a reusable
\emph{pairwise interaction correction} to its denoising prediction,
drawing on relational and compositional modeling
\cite{battaglia2016interaction,mishraGenerativeFactorChaining2025}.
For a team of $N$ robots however,  evaluating such corrections densely requires considering
$N(N-1)$ ordered interactions at every reverse step.
Simply selecting fewer interactions is not sufficient either.
Corrections that are individually useful may become incompatible
when several robots are coordinated simultaneously. Our key insight is to decompose interaction-aware diffusion
into two distinct decisions: \emph{which} pairwise corrections
are worth computing, and \emph{how} their effects should be combined
into a coherent joint update. Both decisions depend on the
surrounding team and the evolving trajectory estimates.

We introduce \AkuleInlineLogo\method, which realizes this
decomposition through three learned modules. A shared
pairwise model $R_{\phi}$ learns reusable interaction
corrections; a contextual selector $U_{\psi}$ uses
inexpensive pairwise descriptors and pooled team
information to choose a directed interaction graph
before evaluating $R_{\phi}$; and a composer
$G_{\omega}$ combines the selected corrections using
context-dependent signed weights. Shared parameters
and symmetric aggregation \cite{zaheer2017deepsets}
make the architecture permutation equivariant:
relabeling the robots correspondingly relabels its
outputs, allowing the same modules to operate across
variable team sizes. We train these modules through offline trajectory
optimization. Matched pairwise optimizations supervise
$R_{\phi}$, joint trajectory optimization provides
composition targets for $G_{\omega}$, and sparse-support
learning trains $U_{\psi}$ without prescribing a
ground-truth interaction graph. Complete denoising
rollouts further refine the learned coordination.
At inference, diffusion produces a coordinated joint
proposal that downstream search validates and repairs
as needed.

We evaluate \method on obstacle-rich navigation and
congested position-exchange benchmarks
\cite{shaoul2025mmd,liang2025smd,liang2025dgd}.
\method achieves competitive or superior planning
success with complete-planner speedups of up to
$25\times$, alongside zero-shot population transfer
across team sizes and successful planning with up
to $256$ robots. Ablations further show that
learned selection and composition reduce pre-repair
conflicts and downstream repair effort relative to
budget-matched heuristic selection and uniform
summation.

\noindent\textcolor{akBlue}{\textbf{Contributions.}} Our main contributions are:
\textbf{(i)} A permutation-equivariant diffusion architecture
that coordinates variable-sized robot teams by selectively
evaluating reusable pairwise interaction corrections and
composing their effects through context-dependent signed weights;
\textbf{(ii)} a trajectory-optimization-based learning procedure
that supervises pairwise interaction corrections and their
composition using physically optimized trajectories, without
requiring a unique ground-truth interaction graph; and
\textbf{(iii)} extensive benchmark evaluations demonstrating
competitive planning success with complete-planner speedups
of up to $25\times$, zero-shot population transfer, and
scalability to hundreds of robots.

%% file: related.tex
\section{Related Work}
Prior work addresses inter-robot coupling through
classical search, constrained or conditional diffusion,
and compositional models built from reusable factors.
\method instead learns reusable pairwise corrections,
which ones to evaluate, and how to combine their
effects in the context of the team.

\noindent\textcolor{akBlue}{\textbf{Diffusion trajectory models ---}}
Learning-based motion planners exploit learned sampling
distributions and motion priors to accelerate planning
\cite{ichter2018cvae,qureshi2019mpnet}. Diffusion-based
planners generate complete trajectories through iterative
denoising \cite{janner2022diffuser,ajay2023decisiondiffuser},
learning from expert demonstrations
\cite{chi2023diffusion}, trajectories produced by classical
planners or optimization \cite{carvalho2023mpd,luo2024PDMP},
or navigation data \cite{yu2024trajdiffusionnav}.
These approaches provide expressive single-agent motion
priors, but independently generated trajectories do not
account for inter-robot constraints.

\noindent\textcolor{akBlue}{\textbf{Classical multi-robot planning ---}}
Inter-robot collision constraints couple individual
plans, making joint-space search expensive as team
size grows. Decoupled and prioritized planners
reduce computation but can fail on solvable instances
\cite{erdmannMultipleMovingObjects1987,
silver_cooperative_2005,sanchez_delaying_2002}.
Subdimensional expansion couples robots only when
interactions require it \cite{wagner2011mstar},
while conflict-based search (CBS) resolves conflicts
through constrained replanning
\cite{sharonConflictBasedSearchOptimal2012}.
Both exploit selective coupling, but search can remain
expensive in heavily interacting scenes. Bounded-suboptimal
CBS variants accelerate search while preserving cost bounds
\cite{barerSuboptimalVariantsConflictBased2014,
liEECBSBoundedSuboptimalSearch2021}.
Large-neighborhood repair \cite{liMAPFLNS2FastRepairing2022},
priority inheritance \cite{okumura2019pibt}, and lazy
constraint addition \cite{okumura2023lacam} offer other
scalable strategies with differing guarantees.
Discrete planning followed by continuous refinement
\cite{honig_trajectory_2018} and reactive collision
avoidance \cite{vandenbergReciprocalNBodyCollision2011}
provide further alternatives. These methods exploit
interaction structure during planning or execution,
rather than learning reusable denoising-time corrections.

\noindent\textcolor{akBlue}{\textbf{Multi-agent diffusion planning ---}}
Offline multi-agent diffusion models learn coordination
through cross-agent attention \cite{zhu2024madiff}
or sparse coordination graphs \cite{zeng2025mcgd},
but address different control objectives from
collision-constrained MRMP. For MRMP, MMD retains
single-robot diffusion priors but resolves coupling
through conflict-driven search and constrained
regeneration \cite{shaoul2025mmd}. SMD enforces
joint constraints through optimization during sampling
\cite{liang2025smd}, while DGD uses discrete MAPF
guidance to organize constrained generation and repair
\cite{liang2025dgd}. Other approaches use decentralized
neighbor prediction \cite{liang2026sid}, local
conditional denoising \cite{lee2026dmbd}, or model-based
diffusion with conflict-based search \cite{He2026MDOC}.
These methods account for interactions through search,
joint optimization, or conditioning on other agents'
motions. \method instead makes the evaluation and
composition of reusable pairwise denoising corrections
separate learned operations.

\noindent\textcolor{akBlue}{\textbf{Compositional diffusion and sparse interactions ---}}
Compositional diffusion constructs complex distributions
from reusable scores or energy-based factors
\cite{liu2022composable,duReduceReuseRecycle2023}.
Graph-structured approaches organize local models
through prescribed graphical relationships
\cite{zhangDiffCollageParallelGeneration2023,
weilbach2023graphically}. In planning, related work
composes skills and task factors
\cite{mishraGenerativeSkillChaining2023,
mishraGenerativeFactorChaining2025}, locally valid
trajectory predictions and reusable score factors
\cite{sun2026ecd,mitra2026factored}, or obstacle and
map-region potentials \cite{luo2024PDMP,shaoul2025mmd}.
These approaches demonstrate the value of reusable
components, but their factorization or composition
rule is often prescribed by task structure.
Sparse interaction relationships can also be learned
for multi-agent coordination \cite{zeng2025mcgd}
and humanoid control \cite{li2026interagent},
although these address different planning objectives.
\method instead learns a time-varying directed support
and a context-dependent signed composition of
pairwise corrections to a frozen motion prior.

%% file: problem_formulation.tex
\section{Preliminaries and Problem Formulation}
\label{sec:problem}

\noindent\textcolor{akBlue}{\textbf{Multi-robot motion planning.}}
We consider $N$ robots operating in a shared workspace
$\mathcal{W}$ \cite{shaoul2025mmd,liang2025smd}.
Robot $i$ has state $s_h^i\in\mathcal{S}^i$ at physical
time index $h$, with configuration $q_h^i$ determining its occupied region $\mathcal{R}^i(q_h^i)$. Given start and goal states $s_{\mathrm{s}}^i$ and
$s_{\mathrm{g}}^i$, we seek trajectories
$\tau^i=(s_1^i,\ldots,s_H^i)$ satisfying
$s_1^i=s_{\mathrm{s}}^i$ and
$s_H^i=s_{\mathrm{g}}^i$.
The joint trajectory is
$\boldsymbol{\tau}=(\tau^1,\ldots,\tau^N)$.
Let $\Omega$ contain joint trajectories satisfying
these endpoints, static-obstacle avoidance,
benchmark-specific motion constraints, and inter-robot
collision avoidance:
    $\mathcal{R}^i(q_h^i)\cap
    \mathcal{R}^j(q_h^j)=\emptyset,
    \text{ } \forall i\neq j,\;h=1,\ldots,H.$
The objective is to find a feasible
$\boldsymbol{\tau}^{\star}\in\Omega$.

\noindent\textcolor{akBlue}{\textbf{Frozen unary diffusion prior.}}
We build on a pretrained single-robot diffusion planner
\cite{janner2022diffuser,shaoul2025mmd}.
Let $x_t^i$ denote robot $i$'s noisy trajectory
representation at reverse-diffusion step $t$, and
$c^i$ its start, goal, and environment context.
The frozen denoiser $f_\theta$ defines
$
    x_{t-1}^i =
    \Phi_t\!\left(
        x_t^i,f_\theta(x_t^i,t,c^i),\xi_t^i
    \right),
$
where $\Phi_t$ is the sampler update and $\xi_t^i$
its stochastic component. Endpoints are hard-conditioned
throughout sampling. Applying $f_\theta$ independently
yields the unary joint prior
$p_\theta^{\mathrm{unary}}(\boldsymbol{\tau}\mid\mathbf c)
=\prod_{i=1}^{N}p_\theta(\tau^i\mid c^i)$,
which supports variable team sizes but does not model
inter-robot dependencies.

\noindent\textcolor{akBlue}{\textbf{Coordinated proposal and repair.}}
We augment the unary reverse process to generate a
joint proposal $\hat{\boldsymbol{\tau}}=\mathcal{P}(\mathbf c)$.
A downstream conflict-based search planner
$\mathcal{H}$ then repairs remaining conflicts:
$
    \boldsymbol{\tau}^{\mathrm{sol}}
    =\mathcal{H}(\hat{\boldsymbol{\tau}}).
$
A returned solution is accepted only if
$\boldsymbol{\tau}^{\mathrm{sol}}\in\Omega$;
planning may also terminate without a solution.
The central challenge is therefore to capture sufficient inter-robot coordination to
produce high-quality proposals without requiring dense pairwise coordination
computation at every reverse-diffusion step. We address this by separating
\emph{which} interactions should be modeled from \emph{how} their effects
should be composed.

%% file: method.tex
\begin{figure*}[!t]
    \centering
    \includegraphics[
    width=0.75\textwidth,
    trim=2bp 4bp 2bp 3bp,
    clip
]{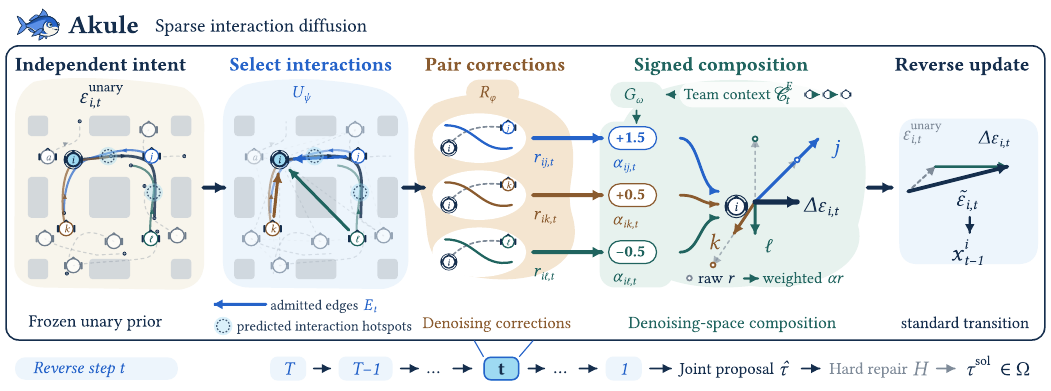}
    \caption{\textbf{\method architecture.}
    At each reverse-diffusion step, the frozen unary prior produces independent
    intent, $U_{\psi}$ selects a directed interaction graph, $R_{\phi}$ computes
    corrections only for the selected pairs, and $G_{\omega}$ composes them
    using context-dependent signed weights. Repeating this update produces a
    joint trajectory proposal that is passed to downstream hard repair.}
    \label{fig:akule_architecture}
    \Description{Overview of the Akule planning architecture. A frozen unary
    diffusion prior generates independent robot intent. A selector chooses
    directed robot interactions, pairwise interaction corrections are computed
    only for selected pairs, and a context-aware composer combines them with
    signed weights. The coordinated correction is applied during reverse
    diffusion, producing a joint trajectory proposal that is passed to a hard
    repair planner.}
\end{figure*}

\section{\texorpdfstring{\AkuleMasterLogo\methodname}{Akule}}
\label{sec:method}
We now introduce \method, which preserves the frozen single-robot diffusion
prior from Sec.~\ref{sec:problem} and adds the interaction reasoning missing
from independent planning. \method consists of three learned modules: \textbf{(i)} A reusable pairwise model
$R_{\phi}$ that learns how one robot should alter another's
denoising prediction; \textbf{(ii)} A contextual selector $U_{\psi}$ that determines which pairwise corrections are worth computing; and \textbf{(iii)} A composer $G_{\omega}$ which learns how the selected corrections should be combined. At inference, $U_{\psi}$ selects
interactions before the expensive $R_{\phi}$ evaluations,
and $G_{\omega}$ composes the admitted corrections into
a coordinated update. All three modules share parameters
across robots, yielding a permutation-equivariant
architecture that is independent of team size. We learn them sequentially through physical
trajectory optimization, in the order
$R_{\phi}\rightarrow G_{\omega}\rightarrow U_{\psi}$
(Sec.~\ref{sec:method_learning}). The resulting interaction-aware prediction for robot $i$ is
\begin{equation}
    \widetilde{\epsilon}_{i,t}
    =
    \epsilon_{i,t}^{\mathrm{unary}}
    +
    \sum_{j\neq i}
    m_{ij,t}\,
    \alpha_{ij,t}\,
    R_{\phi}
    \left(
        x_t^i,x_t^j,t,c^i,c^j
    \right),
    \label{eq:akule_master}
\end{equation}
where $m_{ij,t}\in\{0,1\}$ determines whether the interaction between robots
$i$ and $j$ is evaluated, $R_{\phi}$ produces the corresponding pairwise
correction, and $\alpha_{ij,t}\in\mathbb{R}$ determines how that correction
contributes in the context of the full team. Figure~\ref{fig:akule_architecture} summarizes the resulting reverse-step
pipeline; the remainder of this section develops its components and learning
procedure.

\subsection{Reusable Pairwise Interaction Corrections}
\label{sec:method_factorization}

The single-robot diffusion model produces
$\epsilon_{i,t}^{\mathrm{unary}}=f_{\theta}(x_t^i,t,c^i)$
for robot $i$ at reverse-diffusion step $t$. Applied independently across
robots, these predictions capture goal-directed motion but do not account for
how another robot should alter the denoising update. We capture this missing interaction through a shared temporal-convolutional model
$R_{\phi}$. For an ordered pair $(i,j)$, where robot $j$ influences the update
of robot $i$,
$    r_{ij,t}
    =
    R_{\phi}
    \left(
        x_t^i,x_t^j,t,c^i,c^j
    \right)$
is a correction in the same space as the unary denoising prediction. Conceptually, considering only the interaction with robot $j$ gives the pair-aware prediction
$\epsilon_{i\mid j,t}^{\mathrm{pair}}
=\epsilon_{i,t}^{\mathrm{unary}}+r_{ij,t}$. Thus, $R_{\phi}$ need not reproduce robot $i$'s goal-directed trajectory
prior; it only represents how that prediction should change because robot $j$
is present. These reusable corrections provide a common backbone for inter-robot
coordination. Evaluating that backbone densely, however, would require
$N(N-1)$ calls to $R_{\phi}$ at every reverse step. The first learned decision
in \method is to identify which pairwise corrections are worth
computing from information available \emph{before} $R_{\phi}$ is evaluated.

\subsection{Sparse Contextual Interaction Selection}
\label{sec:method_selector}

\begin{figure}[!t]
    \centering
    \includegraphics[
    width=0.8\columnwidth,
    trim=2bp 6bp 2bp 1bp,
    clip
]{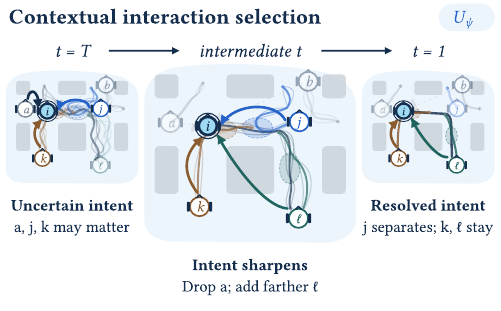}
    \caption{\textbf{Contextual interaction selection.}
    As reverse diffusion sharpens trajectory intent, $U_{\psi}$ can revise the
    directed interaction graph, removing interactions that become irrelevant
    while admitting others that become important.}
    \label{fig:akule_selector}
    \Description{Three snapshots of the same multi-robot planning scene during
reverse diffusion. Early in diffusion several directed interactions are
selected because robot intent is uncertain. As trajectories become more
defined, the selector removes interactions that become irrelevant and admits
others that become important.}
\end{figure}

A useful interaction cannot always be identified from pairwise proximity
alone, and the relevant interactions can change as trajectory intent sharpens
through reverse diffusion (Fig.~\ref{fig:akule_selector}). For example, whether robot $j$ should influence robot $i$ may depend on
the motion of both robots, their intended trajectories, and whether either is
simultaneously interacting with other robots. For every ordered
candidate pair $(i,j)$, we therefore construct an inexpensive pair descriptor
$\chi_{ij,t}$ using: the current noisy trajectories $x_t^i$ and $x_t^j$, their planning
contexts, relative trajectory geometry, unary denoising features, and the
diffusion timestep.

Pair-local information is supplemented with context from the rest of the
team. We use Deep-Sets-style symmetric aggregation~\cite{zaheer2017deepsets},
denoted by $\operatorname{Pool}(\cdot)$, so that the resulting representation
does not depend on robot ordering or population size. For candidate pair
$(i,j)$, we form
    $\rho_{i,t}^{\mathrm{ego}}
    =\operatorname{Pool}_{k\neq i}\chi_{ik,t},
    \text{ }
    \rho_{j,t}^{\mathrm{partner}}
    =\operatorname{Pool}_{k\neq j}\chi_{kj,t},
    \text{ }
    \rho_t^{\mathrm{global}}
    =\operatorname{Pool}_{a\neq b}\chi_{ab,t}.$
The first two summaries describe the other candidate interactions associated
with robots $i$ and $j$, while $\rho_t^{\mathrm{global}}$ summarizes the
interaction context of the full team. A shared selector $U_{\psi}$ uses these quantities to predict a real-valued interaction relevance logit,\\
$
    z_{ij,t}
    =
    U_{\psi}
    \left(
        \chi_{ij,t},
        \rho_{i,t}^{\mathrm{ego}},
        \rho_{j,t}^{\mathrm{partner}},
        \rho_t^{\mathrm{global}}
    \right).$
The sigmoid $p_{ij,t}=\sigma(z_{ij,t})$ provides a
differentiable relevance score during learning.
At inference, each directed pair is either admitted
or rejected using
\begin{equation}
    m_{ij,t}
    =
    \mathbf{1}
    \left[
        p_{ij,t}>\tfrac{1}{2}
    \right]
    =
    \mathbf{1}
    \left[
        z_{ij,t}>0
    \right],
    \label{eq:akule_selector}
\end{equation}
and the resulting mask $m_{ij,t}$ defines the directed interaction graph
$    E_t
    =
    \left\{
        (i,j):
        i\neq j,\;
        m_{ij,t}=1
    \right\}.$
The graph may change across reverse diffusion steps, and the two directions of
an interaction are scored independently. Crucially, $R_{\phi}$ is evaluated
only for $(i,j)\in E_t$. Because the selector uses shared pair processing, no
robot identities, and symmetric set aggregation, reindexing the robots
correspondingly reindexes the selected interactions. The same $U_{\psi}$ can
therefore be applied without architectural modification as $N$ changes.

\subsection{Context-Aware Signed Composition}
\label{sec:method_composer}

\begin{figure}[!t]
    \centering
    \includegraphics[width=0.8\columnwidth]{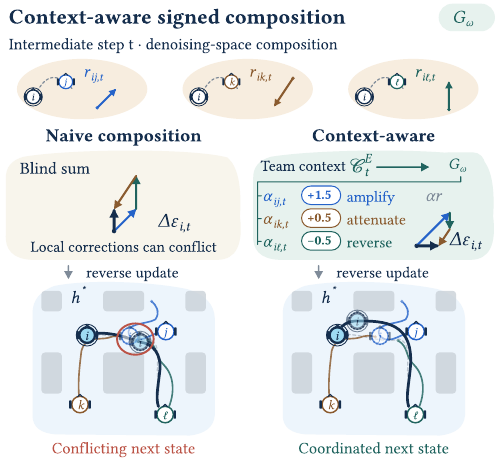}
    \caption{\textbf{Context-aware signed composition.}
    A blind sum of useful pairwise corrections can produce a
    conflicting joint update. $G_{\omega}$ uses team context to amplify,
    attenuate, or reverse selected corrections before the reverse update.}
    \label{fig:akule_composer}
    \Description{Comparison of naive and context-aware composition for one robot.
Several pairwise interaction corrections are shown. Naively summing them
produces a conflicting next state, whereas the context-aware composer assigns
positive and negative weights that amplify, attenuate, or reverse individual
corrections to produce a coordinated next state.}
\end{figure}

Selection determines which pairwise corrections are worth computing, but it
does not determine how several simultaneously relevant corrections should be
used together. As illustrated in Fig.~\ref{fig:akule_composer}, simply summing them treats
every admitted interaction as equally compatible, even though corrections
that are individually useful may reinforce, cancel, or conflict when
considered jointly.

After $R_{\phi}$ is evaluated on the selected edge set, we construct
permutation-invariant context $\mathcal{C}_{t}^{E}$ from the admitted
interactions and the current joint reverse state. This context
allows the effect assigned to one pairwise correction to depend on the other
corrections that are simultaneously active. A shared composer
$G_{\omega}$ assigns each admitted interaction a signed coefficient, $\alpha_{ij,t}
    =
    G_{\omega}
    \left(
        r_{ij,t},
        \chi_{ij,t},
        \mathcal{C}_{t}^{E}
    \right)$, with $
    (i,j)\in E_t.$
The coefficients are unconstrained in sign: depending on the surrounding
interactions, a primitive can be amplified, attenuated, or applied in the
opposite direction. Substituting these coefficients and the hard selection
mask into Eq.~\eqref{eq:akule_master} gives the coordinated denoising
prediction $\widetilde{\epsilon}_{i,t}$. The standard reverse-diffusion transition from Sec.~\ref{sec:problem} is then
applied to $\widetilde{\epsilon}_{i,t}$, with the prescribed start and goal states hard-conditioned throughout the
reverse process.

\subsection{Learning from Trajectory Optimization}
\label{sec:method_learning}

We learn coordination from physically optimized
trajectories rather than requiring binary labels
specifying which interactions are correct. Different supports and coefficient combinations can lead to
equally valid behavior, making a single hand-designed binary graph an
artificial supervision target. We therefore obtain supervision from
trajectory-level physical performance rather than prescribing a particular
interaction graph. Learning proceeds in the same dependency order as the
model: we first learn the reusable pairwise primitive $R_{\phi}$, then use
offline trajectory optimization to generate high-quality composition targets
for $G_{\omega}$, and finally learn which interactions can be removed through
$U_{\psi}$.

\noindent\textcolor{akBlue}{\textbf{Learning the pairwise primitive.}}
We construct the $R_{\phi}$ supervision through
pairwise trajectory optimization. For each ordered pair $(i,j)$, starting
from the same unary estimate, we solve
\begin{equation}
    \hat{\boldsymbol{x}}_{0,t}^{(\gamma)}
    =
    \arg\min_{\boldsymbol{x}}
    \left[
        J_{\mathrm{base}}(\boldsymbol{x})
        +
        \gamma J_{ij}^{\mathrm{int}}(\boldsymbol{x})
    \right],
    \qquad
    \gamma\in\{0,1\},
    \label{eq:akule_pair_optimization}
\end{equation}
where $J_{\mathrm{base}}$ contains the shared trajectory-quality and feasibility
terms, while $J_{ij}^{\mathrm{int}}$ penalizes physical interaction between
robots $i$ and $j$. The two solves use identical initialization, endpoint
constraints, and optimization settings; $\gamma=1$ gives the
interaction-aware solution and $\gamma=0$ the matched null solution.
Denoting their robot-$i$ trajectories by
$\hat{x}_{0,t}^{i,\mathrm{real}}$ and
$\hat{x}_{0,t}^{i,\mathrm{null}}$, respectively, the standard DDPM
correspondence~\cite{ho2020ddpm} gives the pairwise diffusion target
\begin{equation}
    \Delta\epsilon_{i\leftarrow j,t}^{\star}
    =
    -
    \sqrt{
        \frac{\bar{\alpha}_t}
             {1-\bar{\alpha}_t}
    }
    \left(
        \hat{x}_{0,t}^{i,\mathrm{real}}
        -
        \hat{x}_{0,t}^{i,\mathrm{null}}
    \right).
    \label{eq:akule_pair_target}
\end{equation}
Because the two optimizations differ only in whether interaction $(i,j)$ is
modeled, $\Delta\epsilon_{i\leftarrow j,t}^{\star}$ isolates how robot $j$
should modify robot $i$'s unary diffusion prediction. We train $R_{\phi}$ by
regressing to this target over qualified pairs in both directions, and
subsequently freeze it.

\noindent\textcolor{akBlue}{\textbf{Offline composition targets.}}
With $R_{\phi}$ fixed, we next solve the multi-robot composition
problem offline. For each training state, we optimize signed coefficients
$\beta_{ij,t}\in[-s_t,s_t]$ in the dense composition
$\epsilon_{i,t}^{\boldsymbol{\beta}}
=\epsilon_{i,t}^{\mathrm{unary}}
+\sum_{j\neq i}\beta_{ij,t}r_{ij,t}$.
The bound $s_t$ acts as a trust region, preventing the optimizer from arbitrarily amplifying a pairwise primitive and keeping the optimized composition within the regime represented by the frozen interaction model. Let $\hat{\boldsymbol{\tau}}(\boldsymbol{\beta})$ denote the resulting composed joint trajectory. We solve 
\begin{equation}
    \boldsymbol{\beta}^{\star}
    =
    \arg\min_{\lvert\beta_{ij,t}\rvert\leq s_t}
    J_{\mathrm{opt}}
    \left(
        \hat{\boldsymbol{\tau}}(\boldsymbol{\beta})
    \right),
    \label{eq:akule_teacher_optimization}
\end{equation}
where $J_{\mathrm{opt}}$ combines swept inter-robot and obstacle clearance,
trajectory regularity, path quality, trust to the unary motion prior, and
workspace feasibility. Exact objective terms and weights are given in
Appendix~\ref{app:dense_teachers}. We denote the resulting optimized trajectory by
$\boldsymbol{\tau}^{\star}$ and use it as an offline trajectory teacher for
the learned coordination modules. Because the optimization acts only through
the frozen pairwise primitives, these targets remain within the
representational span available to the deployed model.

\noindent\textcolor{akBlue}{\textbf{Learning the composer.}}
We train $G_{\omega}$ to amortize the optimization described above. Under dense interaction support,
the composer is first trained to reproduce the optimized composition through
the transition-matching objective
\begin{equation}
    \mathcal{L}_{G}^{\mathrm{trans}}
    =
    \sum_{t}
    w_t
    \sum_{i=1}^{N}
    \left\|
        \widetilde{\epsilon}_{i,t}
        -
        \epsilon_{i,t}^{\boldsymbol{\beta}^{\star}}
    \right\|_2^2.
    \label{eq:akule_g_transition}
\end{equation}
Here, $w_t$ is a fixed timestep-dependent weight that compensates for the
different noise scales across the diffusion
schedule. Local transition matching provides a useful initialization but does not guarantee that the learned composer remains effective when its own predictions
are propagated through the entire diffusion chain. We therefore refine
$G_{\omega}$ through complete reverse-diffusion rollouts. If
$\hat{\boldsymbol{\tau}}_{G}$ is the resulting trajectory, the rollout
objective is
\begin{equation}
    \mathcal{L}_{G}^{\mathrm{roll}}
    =
    \mathcal{L}_{\mathrm{match}}
    \left(
        \hat{\boldsymbol{\tau}}_{G},
        \boldsymbol{\tau}^{\star}
    \right)
    +
    \lambda_{\mathrm{clear}}
    \mathcal{L}_{\mathrm{clear}}
    \left(
        \hat{\boldsymbol{\tau}}_{G}
    \right)
    +
    \lambda_{\mathrm{trans}}
    \mathcal{L}_{G}^{\mathrm{trans}}.
    \label{eq:akule_g_rollout}
\end{equation}
Here, $\mathcal{L}_{\mathrm{match}}$ keeps the rollout near the optimized
trajectory target, $\mathcal{L}_{\mathrm{clear}}$ penalizes insufficient
inter-robot separation, and the small transition auxiliary stabilizes local
denoising behavior.

\noindent\textcolor{akBlue}{\textbf{Learning the selector.}}
Once composition is fixed, we learn which interactions can be removed without
sacrificing trajectory quality. We first optimize per-edge selector logits
offline using the same hard decisions employed at deployment, while the
sigmoid scores provide differentiable gradients. Sparsity is encouraged
through the expected mean directed degree,
    $\mathcal{R}_{\mathrm{sp}}(z)
    =
    \frac{1}{TN}
    \sum_{t=1}^{T}
    \sum_{i\neq j}
    \sigma(z_{ij,t}).$
The resulting physically successful sparse supports provide bootstrap
supervision for the shared selector $U_{\psi}$ without declaring any one
binary graph to be uniquely correct. 
The bootstrapped selector is then refined through complete reverse diffusion
rollouts with $f_{\theta}$, $R_{\phi}$, and $G_{\omega}$ fixed. Forward
rollouts use the deployed hard decisions from Eq.~\eqref{eq:akule_selector},
while gradients are propagated through the corresponding sigmoid scores using
a straight-through estimator. The rollout objective is augmented by
$\lambda_{\mathrm{sp}}\mathcal{R}_{\mathrm{sp}}$, directly trading trajectory
quality against interaction density.
Additional architectural details, objective definitions, optimization
settings, and the equivalent implementation parameterizations are provided in
Appendices~\ref{sec:architecture} and~\ref{sec:learning}.

\subsection{Complete Planner and Computational Cost}
\label{sec:method_complete_planner}

The final interaction-aware proposal
$\hat{\boldsymbol{\tau}}$ is passed to the downstream
repair planner $\mathcal{H}$ from Sec.~\ref{sec:problem}.
Akule need not eliminate every collision during diffusion;
remaining conflicts are resolved through constrained
search, and only validated solutions are accepted.

The selector still scores all $N(N-1)$ directed candidate
pairs, retaining $O(N^2)$ candidate-scoring complexity.
Its computational benefit comes from avoiding expensive
$R_{\phi}$ evaluations and composition on rejected
edges. With mean selected degree
$\bar d_t=|E_t|/N$, only $N\bar d_t$ corrections
are evaluated rather than $N(N-1)$, saving the fraction $\mathrm{Savings}_t
    =
    1-
    \frac{\bar d_t}{N-1}$
of dense pairwise-correction evaluations. Sec.~\ref{sec:experiments} evaluates
how this sparsity, proposal quality, and complete-planner cost evolve with
team size.

%% file: experiments.tex
\section{Experiments}
\label{sec:experiments}
We ask whether sparse interaction selection preserves planning quality,
whether one coordination model transfers across team sizes, whether the
learned proposal reduces complete-planner computation, and whether these
properties hold across map families and very large teams.

\noindent\textcolor{akBlue}{\textbf{Benchmarks.}}
\emph{Canonical Weave} \cite{shaoul2025mmd} is a high-congestion
position-exchange task ($N\in\{10,20,28\}$); \emph{Basic}, \emph{Dense},
\emph{Shelf}, and \emph{Room} \cite{liang2025smd} are obstacle maps of
increasing clutter ($N\in\{3,\ldots,18\}$, 25 instances per cell);
\emph{Highways} and \emph{Conveyor} \cite{shaoul2025mmd} are the MMD
obstacle environments ($N\in\{3,\ldots,20\}$, 10 tasks per $N$); and
\emph{ScaledWeave} scales the Weave geometry to very large teams
($N\in\{40,\ldots,256\}$, 10 scenes per $N$).
In every family, \method is trained or adapted \emph{only} at the largest
population ($N=28$, $18$, $20$, and $256$, respectively) and evaluated
zero-shot at all smaller populations.

\noindent\textcolor{akBlue}{\textbf{Baselines.}}
MPD \cite{carvalho2023mpd} applies the same frozen single-robot prior
independently and relies entirely on downstream repair, isolating the effect
of coordination; MMD \cite{shaoul2025mmd} couples single-robot diffusion
with conflict-based search (we use its strongest variant, xECBS, rerun on our
panels); SMD \cite{liang2025smd} and DGD \cite{liang2025dgd} enforce
inter-robot constraints during joint sampling and via discrete guidance,
respectively.
On the map families, baseline values are the published DGD-benchmark
results and are not same-hardware timings.

\noindent\textcolor{akBlue}{\textbf{Protocol and metrics.}}
\method and MPD share the same hard-repair planner $\mathcal{H}$ (MMD's
obstacle-aware xECBS) and differ only in the proposal.
We report \emph{success} (validated collision-free plans within the time
limit), \emph{planning latency} (wall-clock time including repair, averaged
over all runs with timeouts counted at the limit), and \emph{quality} (mean
path length and acceleration over successful plans).
Full benchmark, timing, and hardware details are in Appendices~\ref{sec:protocol} and~\ref{sec:analysis}.

\subsection{Quantitative Results}
\label{sec:exp_results}

\input{tables/tab_weave}
\input{tables/tab_smd_maps}
\input{tables/tab_mmd_maps}
\input{tables/tab_scaledweave}

\noindent\textcolor{akBlue}{\textbf{Canonical Weave.}}
All three planners solve every panel (1280/1280, Table~\ref{tab:weave_main}),
so the comparison is one of computation and quality.
\method is $2.8$--$4.7\times$ faster than MMD-xECBS and up to $3.8\times$
faster than MPD across $N\in\{10,20,28\}$, with path lengths within 4\% of
MMD-xECBS and acceleration RMS 6--32\% higher.
On the $N=28$ task shown, the independent MPD proposal is natively
infeasible, while MMD-xECBS and \method both complete the exchange.

\noindent\textcolor{akBlue}{\textbf{Projected-diffusion map families.}}
With one model trained at $N=18$, \method succeeds on 595/600 instances
across Basic, Dense, Shelf, and Room (Table~\ref{tab:smd_basic}), against
112/600 for MPD, 238/600 for MMD, and 584/600 for DGD; SMD solves 298/300
where reported ($N\le9$) but is 50--170$\times$ slower at $N=9$.
At $N=18$, \method is $1.3$--$5.4\times$ faster than DGD (published values,
not same-hardware).
The hardest cells remain hard: success dips to 22--24/25 on Shelf and Room
at $N\ge15$, and DGD is marginally faster on Dense at $N=15$.
Table~\ref{tab:smd_basic} also shows one $N=18$ task per map: SMD fails on Dense
and Room, while \method reaches every goal.

\noindent\textcolor{akBlue}{\textbf{Obstacle environments and ScaledWeave.}}
On Highways and Conveyor (Table~\ref{tab:highways}), a model adapted
only at $N=20$ matches MMD's success at every population (60/60 and 56/60;
MPD: 4/60 and 5/60) while planning $1.1$--$3.8\times$ faster, with path
length within 7\% and acceleration within 17\% of MMD at $N=20$.
On ScaledWeave (Table~\ref{tab:scaledweave}), \method solves every scene
through $N=128$ and 8/10 at $N=256$, whereas MPD and MMD-xECBS time out on
every scene from $N=64$; at $N=40$, the only population all methods
complete, \method is $5.5\times$ faster than MMD-xECBS.

\subsection{Key Findings and Ablation Studies}
\label{sec:key_findings}

Tables~\ref{tab:weave_main}--\ref{tab:akule_ablations} support the central
claim of this work: coordination can be allocated selectively inside the
denoising process without giving up complete-planner success. Four key findings
(KF) explain why, each paired with a figure that isolates one mechanism.

\textcolor{akBlue}{\textbf{(KF\#1) }Sparse selection preserves planning quality, and the share of pairwise work it removes grows with the team.} Fig.~\ref{fig:kf123}a compares sparse \method with the dense variant that evaluates every ordered pair. The sparse planner is consistently faster, and the fraction of $R_\phi$ evaluations it skips grows with $N$, plateauing on Highways and Conveyor. Local congestion affects the pairwise denoising update, not team size; therefore, most of its $N(N-1)$ dense corrections are close to zero, and the selector learns to recognize them using cheap pairwise descriptors. As the team grows, irrelevant pairs multiply faster than relevant ones, increasing the discarded share while the trajectory remains largely unchanged.
The exception is extreme congestion (Shelf at $N=18$, Table~\ref{tab:akule_ablations}), where borderline pairs are dropped, and the sparse planner loses a few time-limited instances.

\input{tables/fig_kf123}

\textcolor{akBlue}{\textbf{(KF\#2)} One coordination model transfers zero-shot across team
sizes.}
Fig.~\ref{fig:kf123}b plots success against population for every family, normalized by the single training population; everything to the left of $N_{\rm train}$ is zero-shot.
\method's curves stay flat where MMD's fall away.
Nothing in the architecture sees $N$; the selector and composer operate on a set of pairwise descriptors and a permutation-invariant summary of the team, so a smaller team is simply a smaller set whose interactions are sub-configurations of those encountered in a crowded scene at the training population. Training at the largest feasible population is therefore the right default: it exposes the modules to the densest interactions they will meet, and the smaller teams come for free.

\textcolor{akBlue}{\textbf{(KF\#3)} The learned proposal hands repair far fewer conflicts, and
that is where the speed-up comes from.}
Fig.~\ref{fig:kf123}c counts sampled collisions in the unary prior's joint proposal before and after \method's corrections and before any smoothing or repair, on matched scenes and noise.
Across all populations and all seven environments, every condition remains well below the diagonal by roughly the same fraction.
Conflict-based repair pays for each remaining conflict with a constraint tree
branch and low-level replans, so with an identical backend, this explains the
speed-ups over MPD and MMD (Tables~\ref{tab:weave_main},~\ref{tab:highways});
in the ablation (Table~\ref{tab:akule_ablations}), variants leaving more
conflicts take proportionally longer.
The reduction is smallest where the prior already yields few conflicts
(Weave, $N=10$) and on Conveyor, whose narrow lanes leave corrections little
room.

\textcolor{akBlue}{\textbf{(KF\#4)} The same formulation covers different map families and
scales to hundreds of robots.}
The learned modules are shared across robots and conditioned on each robot's own planning context, so a new map changes the unary prior; they correct not the coordination mechanism, carrying from free-space exchanges to cluttered maps, shelves, rooms, and the MMD environments (Tables~\ref{tab:smd_basic} and~\ref{tab:highways}). On ScaledWeave (Fig.~\ref{fig:kf4}), from 40 to several hundred robots, planning time grows more slowly than the team and per-robot time stays roughly constant, while the search-based baselines hit the time limit from the second population onward. Pair scoring, the quadratic part, is cheap; the expensive evaluation and composition scale with the selected degree, and the discarded share keeps rising with $N$ (Fig.~\ref{fig:kf123}a). Conflict-based search faces a conflict count that grows faster than $N$ and lacks such a lever.

\input{tables/fig_kf4}

\noindent \textcolor{akBlue}{\textbf{Ablation Studies}}
We ablate interaction selection and composition on Canonical
Weave ($N=28$, 30 instances) and Shelf ($N=18$, 25 instances),
using the largest population in each benchmark family.
All variants share the frozen unary prior, pairwise model
$R_{\phi}$, diffusion schedule, and downstream repair planner.

\noindent {\textbf{Ablation variants.}}
\textit{Dense Akule:} evaluates all directed pairwise corrections
using the learned composer.
\textit{Raw sum:} replaces $G_{\omega}$ with uniform weights
($\alpha=1$), retaining $U_{\psi}$.
\textit{Calibrated sum:} selects a constant coefficient from
$\{0,0.25,0.5,0.75,1,1.5,2\}$ using full diffusion rollouts
on held-out validation instances, prioritizing inter-robot conflicts on Weave and
obstacle feasibility on Shelf ($\alpha=1$ and
$\alpha=0$, respectively).
\textit{Geometric selector:} ranks pairs by Euclidean distance
between their start positions.
\textit{Collision-risk selector:} ranks pairs by minimum
time-aligned separation along the current step's estimated
denoised unary trajectories. Both heuristics select the
global top-$K_t$ directed pairs, where $K_t$ matches the
rounded mean interaction count of $U_{\psi}$ at each
diffusion step.

\noindent\textcolor{akBlue}{\textbf{Results.}}
Table~\ref{tab:akule_ablations} shows that learned
composition reduces pre-repair conflicts and downstream
repair relative to uniform summation, while learned
selection outperforms both geometric and collision-risk
heuristics at comparable interaction budgets.
On Weave, \method produces $32.1$ conflicts compared
with $94.2$ for uniform summation and $59.2$/$63.0$
for the heuristic selectors, with similar trends on
Shelf. Relative to dense Akule, sparse selection avoids
$29.4\%$ and $80.0\%$ of pairwise corrections on Weave
and Shelf, respectively, while reducing planning latency.
All variants solve every Weave instance, although dense
Akule achieves slightly higher success on Shelf
($24/25$ versus $22/25$), illustrating the
efficiency--reliability trade-off under greater congestion. 

\input{tables/tab_ablations}

\section{Conclusion and Limitations}

We introduced \AkuleInlineLogo \method, which augments a frozen
single-robot diffusion prior with reusable pairwise
corrections, learned sparse interaction selection,
and context-aware signed composition for MRMP. Trained from
physically optimized trajectories, these modules
coordinate variable-sized robot teams without
prescribed interaction graphs. Across diverse MRMP
benchmarks, \method achieves competitive planning
success with $25\times$ computational savings,
zero-shot population transfer, and successful
planning with up to $256$ robots. Ablations demonstrate
the benefits of learned selection and composition
in reducing the coordination burden before hard repair.

\noindent\textcolor{akBlue}{\textbf{Limitations and Future Work.}}
Although sparse selection avoids expensive pairwise
correction evaluations, the selector still scores
$O(N^2)$ candidate interactions. Hierarchical candidate
screening could reduce this overhead while
preserving context-dependent selection. Moreover,
composed updates remain within the span of learned
pairwise corrections, potentially limiting coordination
that requires more complex higher-order interactions.
Extending the correction basis to capture such
interactions is a promising direction.
Finally, learned proposals do not guarantee feasibility,
and downstream repair can dominate computation in
highly congested settings. Incorporating repair directly in learning, alongside adaptive sparsity, could improve the trade-off between computational efficiency and planning success.

%% file: tables/tab_weave.tex
\begin{table}[!ht]
    \centering
    \caption{Canonical Weave: complete-planner results on matched panels and the three planners' saved trajectories on one $N{=}28$ task. Scarlet brackets mark selected collisions.}
    \label{tab:weave_main}
    \akStyle
    \def\akPanel{\dimexpr0.29\columnwidth\relax}
    \setlength{\fboxsep}{1.5pt}
    \def\akFail{{\bfseries\color{red!80!black}[FAIL]}}
    \akFit{\begin{tabular}{c l c r c}
        \akH{$N$} & \akH{Method} & \akH{Success} & \akH{Plan (s)} & \akH{Quality (P/A)} \\
        \akAccentLine
        10 & MPD & 512/512 & 3.973 & 1.781 / 1.454 \\ \akLine
        10 & MMD-xECBS & 512/512 & 12.180 & 1.811 / 1.504 \\ \akLine
        \akOurs 10 & \AkuleInlineLogo\akM{Akule} & 512/512 & 3.128 & 1.811 / 1.593 \\
        \akGroupLine
        20 & MPD & 256/256 & 26.673 & 1.828 / 1.942 \\ \akLine
        20 & MMD-xECBS & 256/256 & 33.424 & 1.842 / 1.848 \\ \akLine
        \akOurs 20 & \AkuleInlineLogo\akM{Akule} & 256/256 & 7.106 & 1.910 / 2.447 \\
        \akGroupLine
        28 & MPD & 512/512 & 77.602 & 1.877 / 2.214 \\ \akLine
        28 & MMD-xECBS & 512/512 & 82.909 & 1.889 / 2.196 \\ \akLine
        \akOurs 28 & \AkuleInlineLogo\akM{Akule} & 512/512 & 29.944 & 1.949 / 2.813 \\
        \akAccentLine
    \end{tabular}}
    \par\vspace{5pt}
    \begin{tabular}{@{}c@{\hspace{5pt}}c@{\hspace{5pt}}c@{}}
        \colorbox{white}{\includegraphics[width=\akPanel]{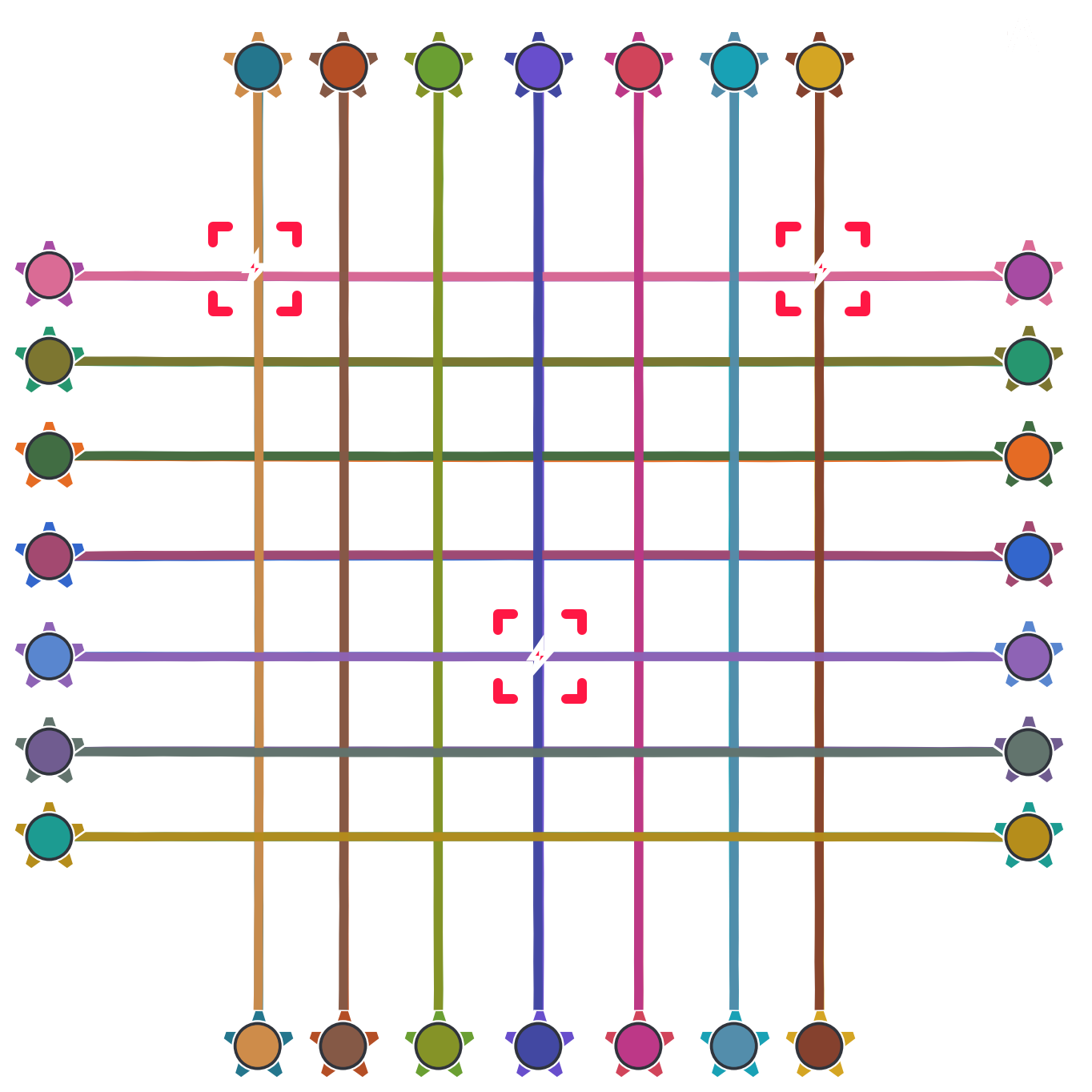}} & \colorbox{white}{\includegraphics[width=\akPanel,trim=5.70 6.07 3.40 3.04,clip]{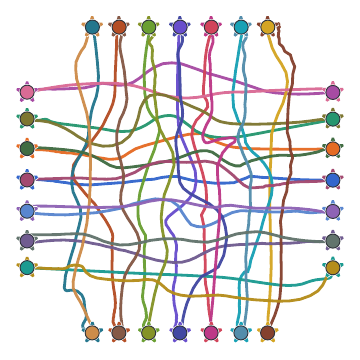}} & \colorbox{akTint}{\includegraphics[width=\akPanel]{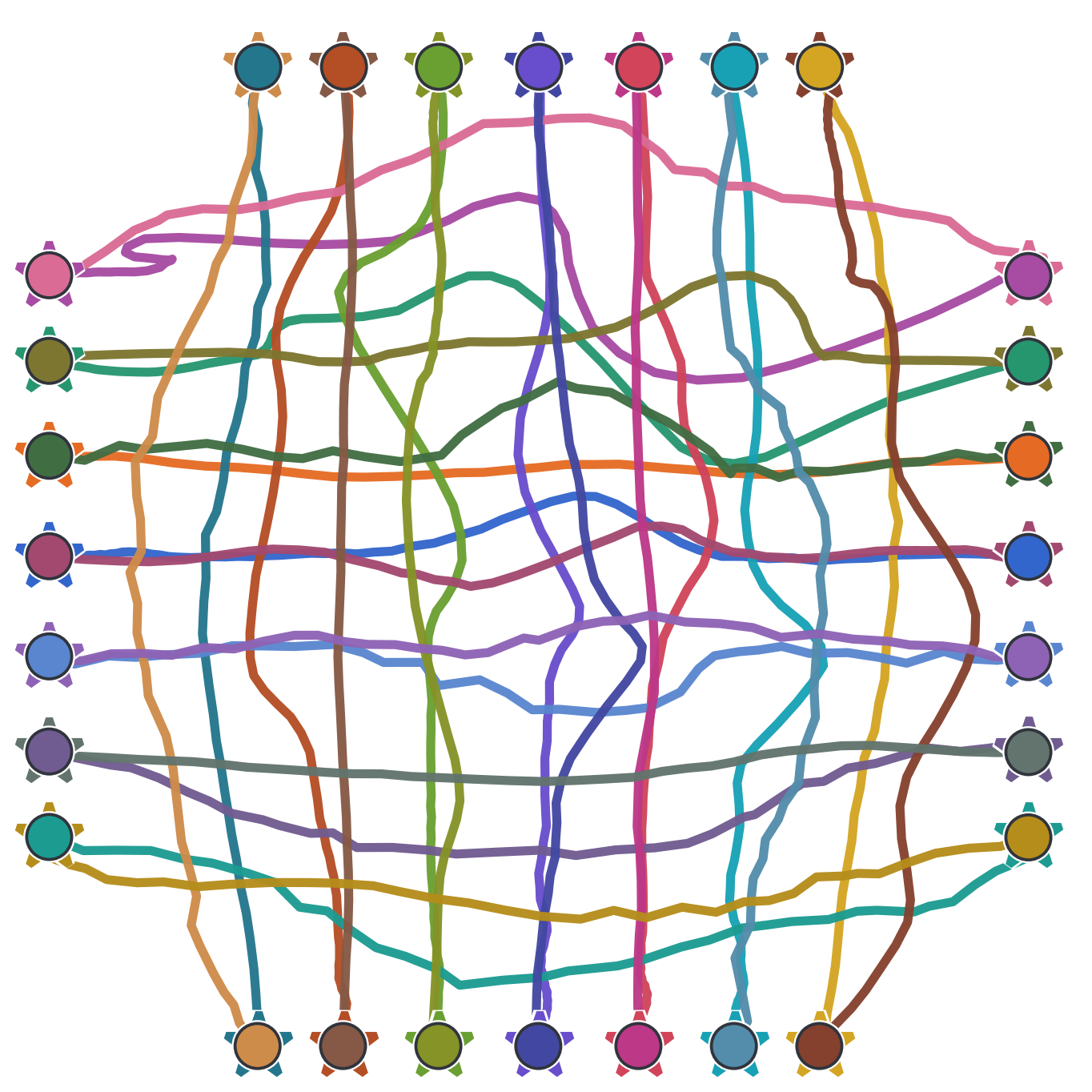}} \\[1pt]
        MPD~\akFail & MMD-xECBS & \AkuleInlineLogo\akM{Akule} \\
    \end{tabular}
\end{table}

%% file: tables/tab_smd_maps.tex
\begin{table*}[t]
    \centering
    \caption{Projected-diffusion map families: MPD, MMD, SMD, and DGD are published DGD-benchmark values. Below, one $N{=}18$ task per map family, the same task for SMD and \method; scarlet brackets mark selected collisions.}
    \label{tab:smd_basic}\label{tab:smd_dense}\label{tab:smd_shelf}\label{tab:smd_room}\label{fig:qual_smd}
    \akStyle
    \def\akFail{{\bfseries\color{red!80!black}[FAIL]}}
    \akFit{\begin{tabular}{c l c r c@{}p{3\tabcolsep}@{}c r c@{}p{3\tabcolsep}@{}c r c@{}p{3\tabcolsep}@{}c r c}
         & & \multicolumn{3}{c}{\akH{Basic}} & & \multicolumn{3}{c}{\akH{Dense}} & & \multicolumn{3}{c}{\akH{Shelf}} & & \multicolumn{3}{c}{\akH{Room}} \\
        \arrayrulecolor{akGroup}\cmidrule(lr){3-5}\cmidrule(lr){7-9}\cmidrule(lr){11-13}\cmidrule(lr){15-17}\arrayrulecolor{akRule}
        \akH{$N$} & \akH{Method} & \akH{Success} & \akH{Time (s)} & \akH{Quality (P/A)} & & \akH{Success} & \akH{Time (s)} & \akH{Quality (P/A)} & & \akH{Success} & \akH{Time (s)} & \akH{Quality (P/A)} & & \akH{Success} & \akH{Time (s)} & \akH{Quality (P/A)} \\
        \akAccentLine
        3 & MPD & 23/25 & 9.300 & 1.110 / 0.006 &  & 22/25 & 8.300 & 1.140 / 0.008 &  & 8/25 & 9.100 & 1.150 / 0.007 &  & 3/25 & 8.800 & 1.120 / 0.007 \\ \akLine
        3 & MMD & 25/25 & 13.100 & 1.120 / 0.006 &  & 15/25 & 19.100 & 1.100 / 0.006 &  & 17/25 & 23.700 & 1.160 / 0.006 &  & 15/25 & 28.200 & 1.220 / 0.007 \\ \akLine
        3 & SMD & 25/25 & 76.600 & 1.100 / 0.005 &  & 25/25 & 81.200 & 1.120 / 0.006 &  & 25/25 & 88.000 & 1.140 / 0.007 &  & 25/25 & 72.600 & 1.170 / 0.008 \\ \akLine
        3 & DGD & 25/25 & 6.500 & 1.160 / 0.008 &  & 25/25 & 6.600 & 1.170 / 0.007 &  & 25/25 & 31.500 & 1.270 / 0.008 &  & 25/25 & 17.100 & 1.360 / 0.008 \\ \akLine
        \akOurs 3 & \AkuleInlineLogo\akM{Akule} & 25/25 & 0.570 & 1.253 / 0.004 &  & 25/25 & 2.184 & 1.271 / 0.009 &  & 25/25 & 0.587 & 1.437 / 0.003 &  & 25/25 & 1.064 & 1.224 / 0.005 \\
        \akGroupLine
        6 & MPD & 19/25 & 9.100 & 1.120 / 0.006 &  & 1/25 & 9.200 & 1.110 / 0.008 &  & 0/25 & \multicolumn{2}{c}{\akFail} &  & 0/25 & \multicolumn{2}{c}{\akFail} \\ \akLine
        6 & MMD & 25/25 & 27.100 & 1.120 / 0.006 &  & 10/25 & 36.500 & 1.150 / 0.006 &  & 8/25 & 48.500 & 1.180 / 0.007 &  & 6/25 & 55.500 & 1.160 / 0.007 \\ \akLine
        6 & SMD & 25/25 & 254.300 & 1.100 / 0.006 &  & 25/25 & 287.300 & 1.130 / 0.007 &  & 25/25 & 274.300 & 1.160 / 0.008 &  & 25/25 & 181.000 & 1.190 / 0.007 \\ \akLine
        6 & DGD & 25/25 & 12.200 & 1.210 / 0.011 &  & 25/25 & 12.300 & 1.260 / 0.012 &  & 25/25 & 64.600 & 1.300 / 0.011 &  & 25/25 & 38.900 & 1.440 / 0.013 \\ \akLine
        \akOurs 6 & \AkuleInlineLogo\akM{Akule} & 25/25 & 2.042 & 1.301 / 0.007 &  & 25/25 & 4.067 & 1.268 / 0.010 &  & 25/25 & 1.449 & 1.430 / 0.006 &  & 25/25 & 2.553 & 1.256 / 0.007 \\
        \akGroupLine
        9 & MPD & 18/25 & 9.900 & 1.080 / 0.008 &  & 0/25 & \multicolumn{2}{c}{\akFail} &  & 0/25 & \multicolumn{2}{c}{\akFail} &  & 0/25 & \multicolumn{2}{c}{\akFail} \\ \akLine
        9 & MMD & 25/25 & 41.400 & 1.120 / 0.006 &  & 7/25 & 51.400 & 1.160 / 0.006 &  & 2/25 & 61.600 & 1.160 / 0.007 &  & 2/25 & 142.400 & 1.150 / 0.007 \\ \akLine
        9 & SMD & 25/25 & 504.000 & 1.120 / 0.006 &  & 25/25 & 582.100 & 1.130 / 0.008 &  & 24/25 & 607.800 & 1.170 / 0.009 &  & 24/25 & 239.400 & 1.220 / 0.008 \\ \akLine
        9 & DGD & 25/25 & 17.400 & 1.260 / 0.013 &  & 25/25 & 18.300 & 1.300 / 0.014 &  & 25/25 & 80.400 & 1.310 / 0.013 &  & 25/25 & 59.900 & 1.490 / 0.014 \\ \akLine
        \akOurs 9 & \AkuleInlineLogo\akM{Akule} & 25/25 & 3.397 & 1.300 / 0.008 &  & 25/25 & 8.557 & 1.289 / 0.011 &  & 25/25 & 3.547 & 1.436 / 0.008 &  & 25/25 & 4.740 & 1.289 / 0.010 \\
        \akGroupLine
        12 & MPD & 13/25 & 8.700 & 1.100 / 0.009 &  & 0/25 & \multicolumn{2}{c}{\akFail} &  & 0/25 & \multicolumn{2}{c}{\akFail} &  & 0/25 & \multicolumn{2}{c}{\akFail} \\ \akLine
        12 & MMD & 25/25 & 56.000 & 1.120 / 0.006 &  & 2/25 & 62.500 & 1.150 / 0.006 &  & 1/25 & 95.100 & 1.220 / 0.008 &  & 1/25 & 122.600 & 1.140 / 0.008 \\ \akLine
        12 & DGD & 24/25 & 25.000 & 1.280 / 0.014 &  & 25/25 & 32.900 & 1.350 / 0.016 &  & 25/25 & 131.500 & 1.330 / 0.013 &  & 25/25 & 84.000 & 1.500 / 0.015 \\ \akLine
        \akOurs 12 & \AkuleInlineLogo\akM{Akule} & 25/25 & 6.110 & 1.302 / 0.009 &  & 25/25 & 15.513 & 1.291 / 0.012 &  & 25/25 & 6.644 & 1.464 / 0.011 &  & 25/25 & 11.941 & 1.304 / 0.012 \\
        \akGroupLine
        15 & MPD & 3/25 & 10.300 & 1.090 / 0.009 &  & 0/25 & \multicolumn{2}{c}{\akFail} &  & 0/25 & \multicolumn{2}{c}{\akFail} &  & 0/25 & \multicolumn{2}{c}{\akFail} \\ \akLine
        15 & MMD & 25/25 & 70.900 & 1.120 / 0.006 &  & 1/25 & 72.400 & 1.170 / 0.007 &  & 0/25 & \multicolumn{2}{c}{\akFail} &  & 0/25 & \multicolumn{2}{c}{\akFail} \\ \akLine
        15 & DGD & 24/25 & 41.400 & 1.310 / 0.014 &  & 24/25 & 37.000 & 1.350 / 0.017 &  & 23/25 & 187.000 & 1.360 / 0.014 &  & 25/25 & 115.500 & 1.520 / 0.016 \\ \akLine
        \akOurs 15 & \AkuleInlineLogo\akM{Akule} & 25/25 & 8.991 & 1.336 / 0.011 &  & 25/25 & 38.405 & 1.315 / 0.013 &  & 25/25 & 13.931 & 1.484 / 0.012 &  & 24/25 & 15.946 & 1.327 / 0.013 \\
        \akGroupLine
        18 & MPD & 2/25 & 9.400 & 1.090 / 0.008 &  & 0/25 & \multicolumn{2}{c}{\akFail} &  & 0/25 & \multicolumn{2}{c}{\akFail} &  & 0/25 & \multicolumn{2}{c}{\akFail} \\ \akLine
        18 & MMD & 24/25 & 86.300 & 1.130 / 0.006 &  & 2/25 & 87.100 & 1.180 / 0.007 &  & 0/25 & \multicolumn{2}{c}{\akFail} &  & 0/25 & \multicolumn{2}{c}{\akFail} \\ \akLine
        18 & DGD & 24/25 & 65.700 & 1.330 / 0.015 &  & 19/25 & 47.800 & 1.370 / 0.017 &  & 21/25 & 324.800 & 1.380 / 0.015 &  & 25/25 & 219.000 & 1.550 / 0.016 \\ \akLine
        \akOurs 18 & \AkuleInlineLogo\akM{Akule} & 25/25 & 12.058 & 1.333 / 0.011 &  & 25/25 & 37.381 & 1.335 / 0.017 &  & 22/25 & 76.923 & 1.487 / 0.013 &  & 24/25 & 66.399 & 1.372 / 0.017 \\
        \akAccentLine
    \end{tabular}}
    \par\vspace{6pt}
    \def\akPanel{\dimexpr0.106\textwidth\relax}
    \setlength{\fboxsep}{1.5pt}
    \def\akFail{{\bfseries\color{red!80!black}[FAIL]}}
    \begin{tabular}{@{}c@{\hspace{2pt}}c@{\hspace{10pt}}c@{\hspace{2pt}}c@{\hspace{10pt}}c@{\hspace{2pt}}c@{\hspace{10pt}}c@{\hspace{2pt}}c@{}}
        \multicolumn{2}{c}{\akH{Basic}} & \multicolumn{2}{c}{\akH{Dense}} & \multicolumn{2}{c}{\akH{Shelf}} & \multicolumn{2}{c}{\akH{Room}} \\[1pt]
        \colorbox{white}{\includegraphics[width=\akPanel,trim=0.00 1.68 5.79 4.11,clip]{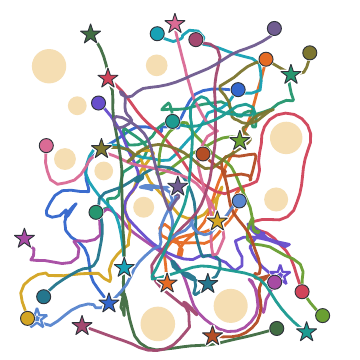}} & \colorbox{akTint}{\includegraphics[width=\akPanel]{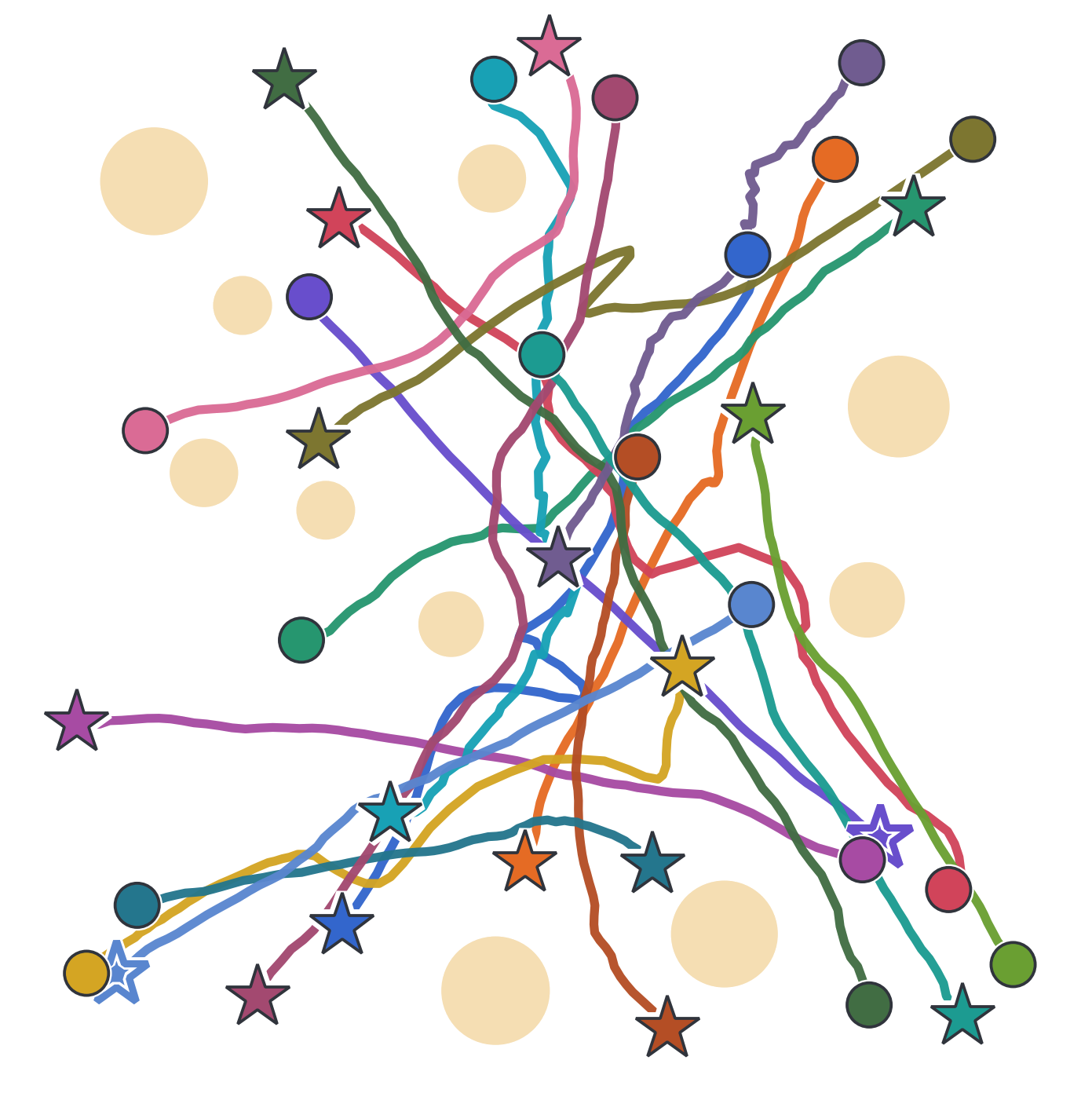}} & \colorbox{white}{\includegraphics[width=\akPanel]{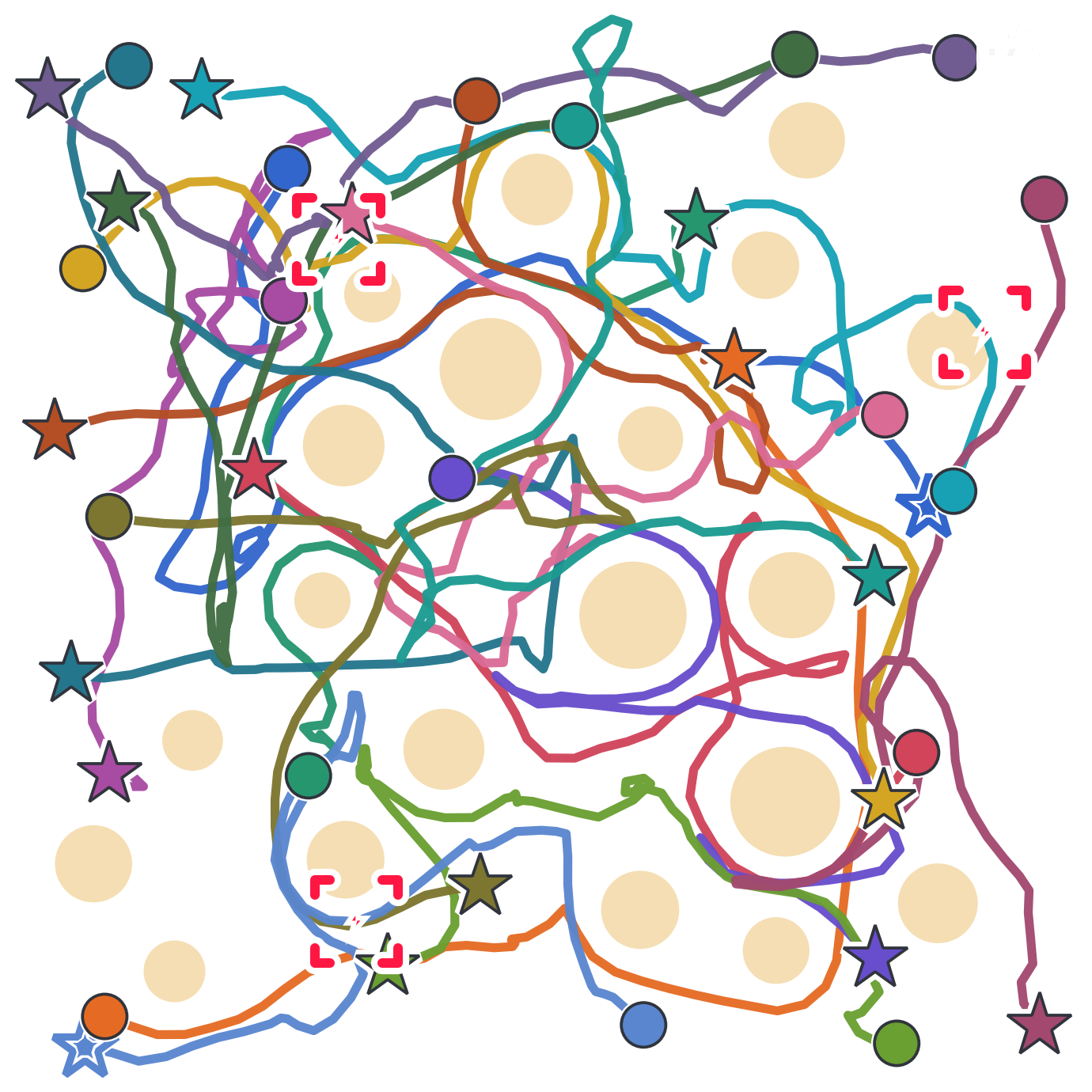}} & \colorbox{akTint}{\includegraphics[width=\akPanel]{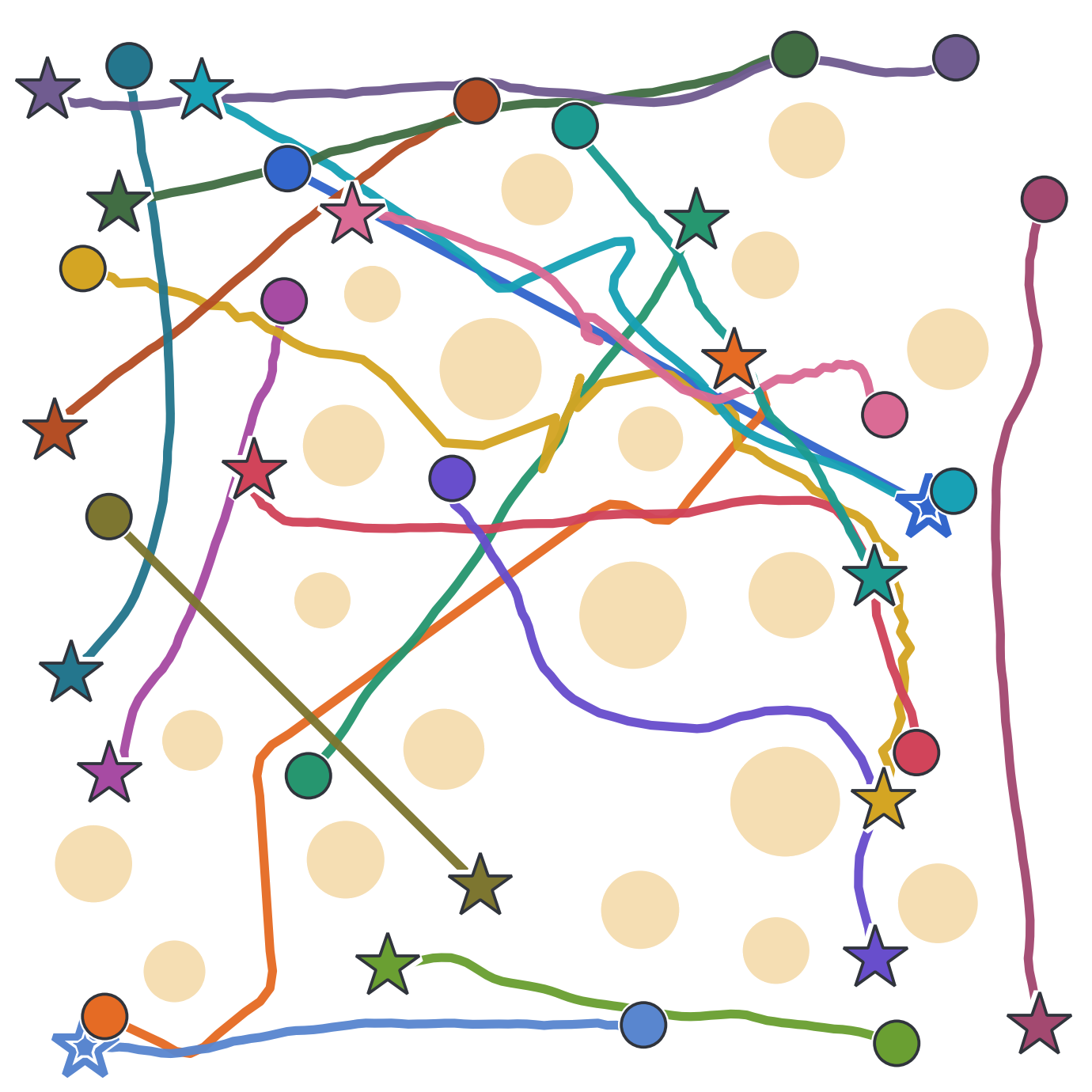}} & \colorbox{white}{\includegraphics[width=\akPanel,trim=0.22 0.00 2.82 3.04,clip]{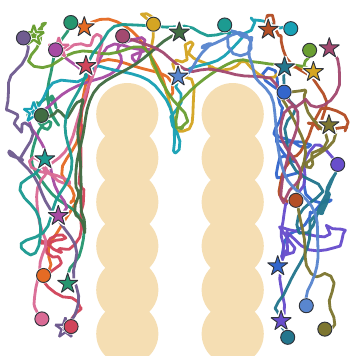}} & \colorbox{akTint}{\includegraphics[width=\akPanel]{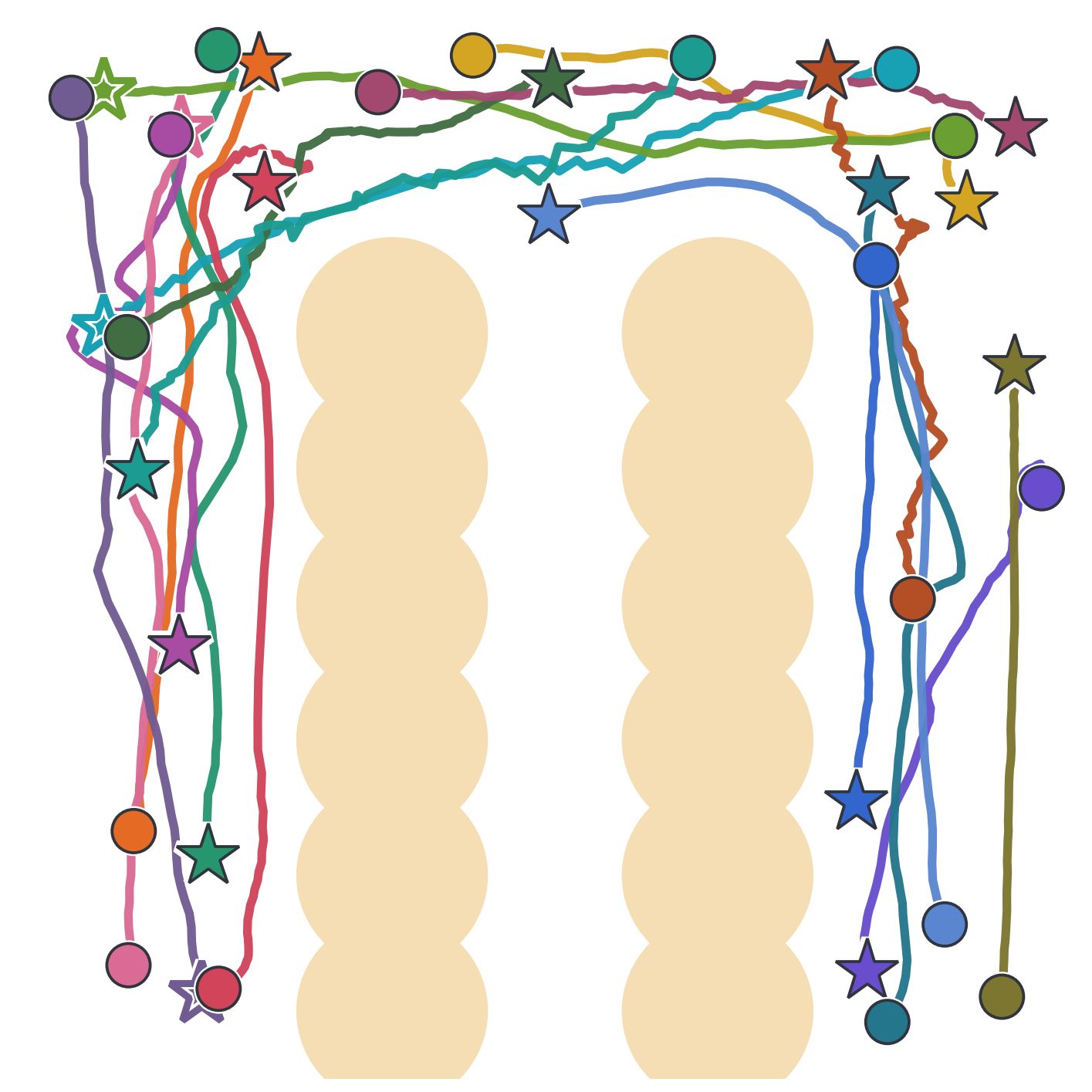}} & \colorbox{white}{\includegraphics[width=\akPanel]{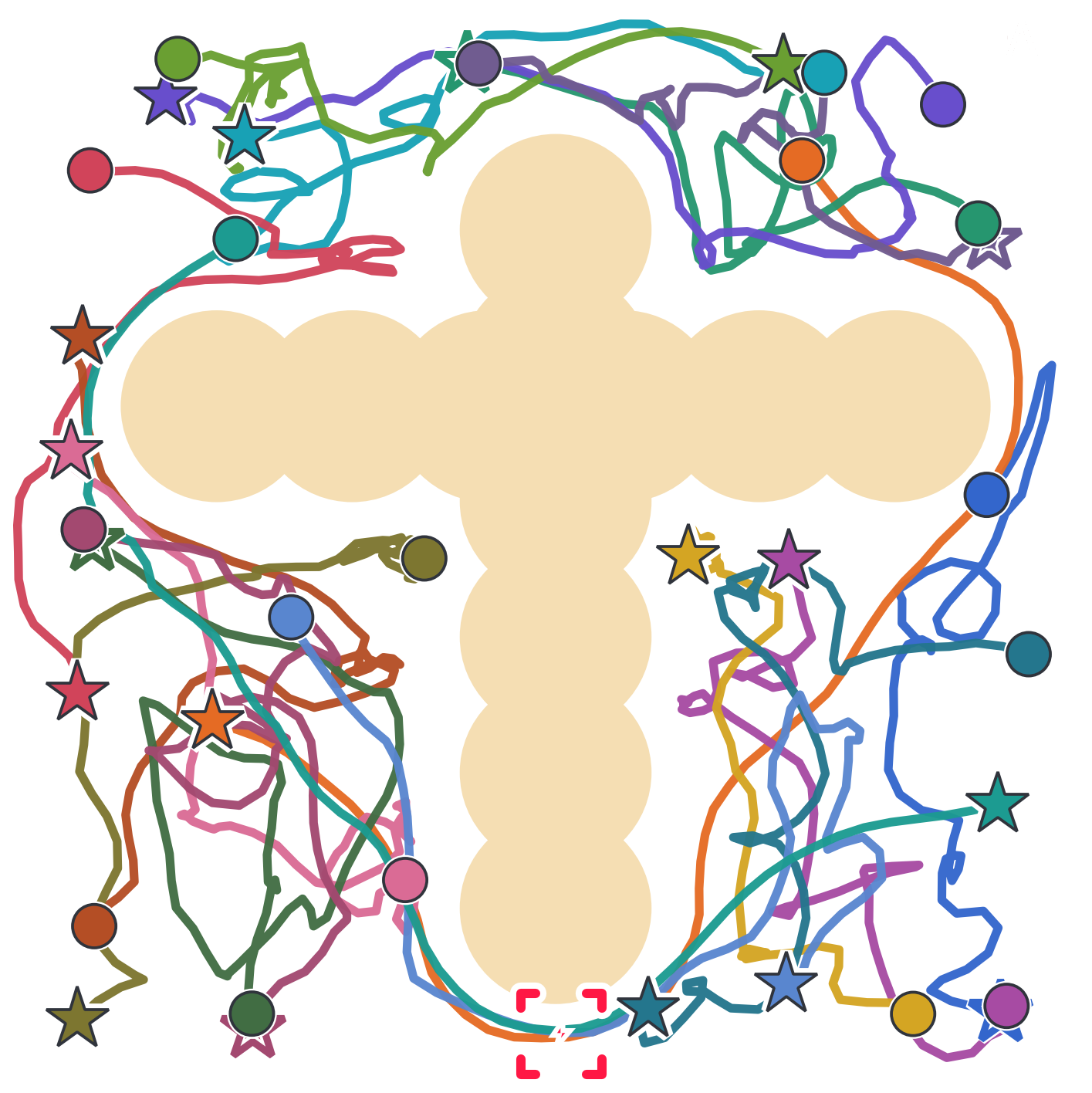}} & \colorbox{akTint}{\includegraphics[width=\akPanel]{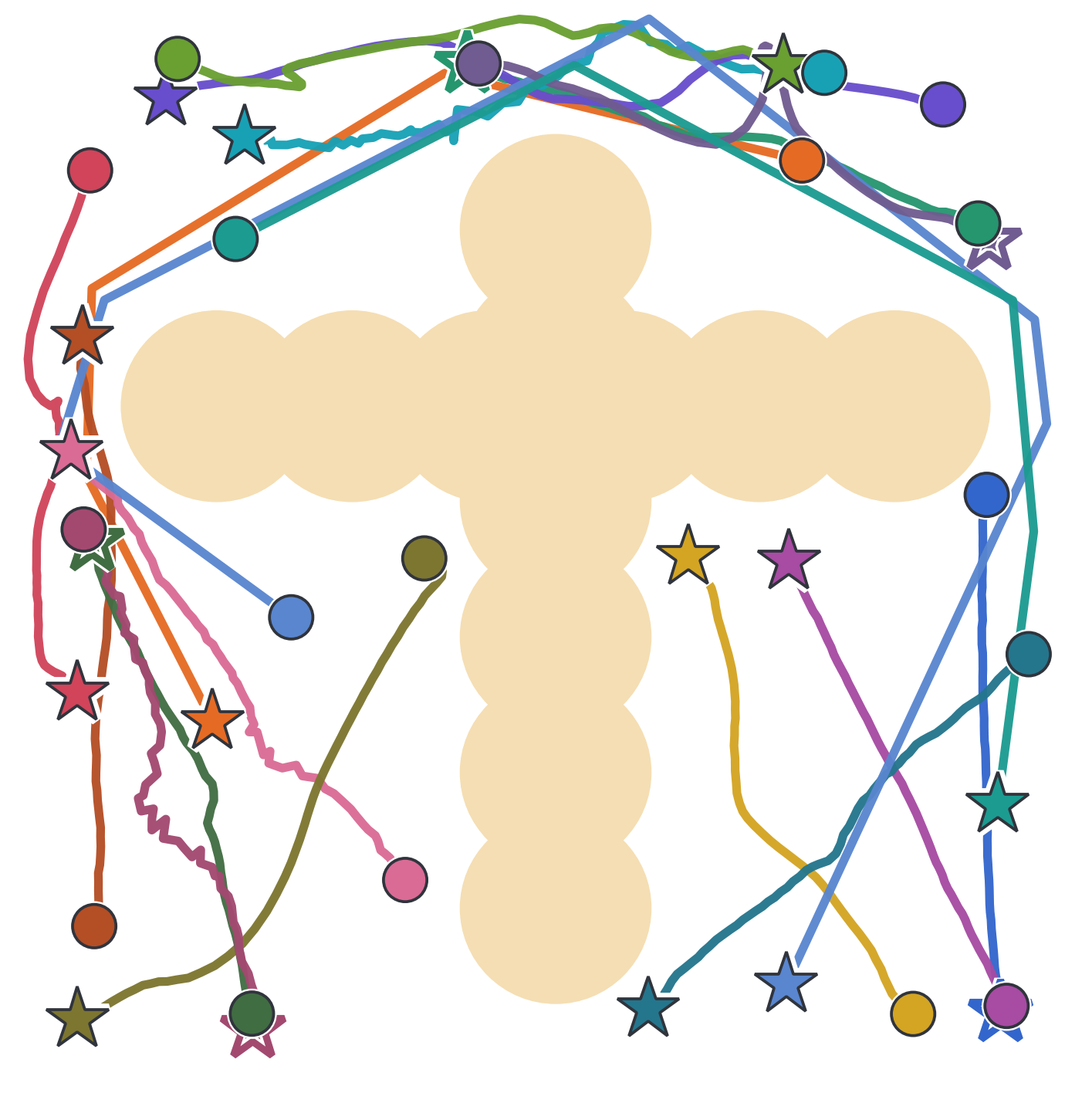}} \\[1pt]
        SMD & \AkuleInlineLogo\akM{Akule} & SMD~\akFail & \AkuleInlineLogo\akM{Akule} & SMD & \AkuleInlineLogo\akM{Akule} & SMD~\akFail & \AkuleInlineLogo\akM{Akule} \\
    \end{tabular}
\end{table*}

%% file: tables/tab_mmd_maps.tex
\begin{table}[!ht]
    \centering
    \caption{Highways and Conveyor (10 tasks per $N$; all methods share the environment-specific unary prior and \method is adapted at $N{=}20$ only), with one $N{=}20$ task per environment shown below. Scarlet brackets mark selected collisions.}
    \label{tab:highways}\label{tab:conveyor}\label{fig:qual_obstacle}
    \akStyle
    \def\akFail{{\bfseries\color{red!80!black}[FAIL]}}
    \akFit{\begin{tabular}{c l c r c@{}p{3\tabcolsep}@{}c r c}
         & & \multicolumn{3}{c}{\akH{Highways}} & & \multicolumn{3}{c}{\akH{Conveyor}} \\
        \arrayrulecolor{akGroup}\cmidrule(lr){3-5}\cmidrule(lr){7-9}\arrayrulecolor{akRule}
        \akH{$N$} & \akH{Method} & \akH{Success} & \akH{Plan (s)} & \akH{Quality (P/A)} & & \akH{Success} & \akH{Plan (s)} & \akH{Quality (P/A)} \\
        \akAccentLine
        3 & MPD & 4/10 & 2.152 & 1.935/0.038 &  & 4/10 & 2.287 & 3.019/0.093 \\ \akLine
        3 & MMD & 10/10 & 2.642 & 1.846/0.038 &  & 10/10 & 2.923 & 3.157/0.103 \\ \akLine
        \akOurs 3 & \AkuleInlineLogo\akM{Akule} & 10/10 & 0.905 & 1.841/0.039 &  & 10/10 & 1.058 & 3.125/0.104 \\
        \akGroupLine
        6 & MPD & 0/10 & \multicolumn{2}{c}{\akFail} &  & 1/10 & 4.510 & 2.882/0.094 \\ \akLine
        6 & MMD & 10/10 & 5.276 & 1.905/0.037 &  & 10/10 & 5.818 & 2.987/0.102 \\ \akLine
        \akOurs 6 & \AkuleInlineLogo\akM{Akule} & 10/10 & 1.392 & 1.873/0.038 &  & 10/10 & 2.284 & 3.284/0.125 \\
        \akGroupLine
        9 & MPD & 0/10 & \multicolumn{2}{c}{\akFail} &  & 0/10 & \multicolumn{2}{c}{\akFail} \\ \akLine
        9 & MMD & 10/10 & 8.358 & 1.978/0.041 &  & 10/10 & 10.402 & 3.353/0.124 \\ \akLine
        \akOurs 9 & \AkuleInlineLogo\akM{Akule} & 10/10 & 3.072 & 2.045/0.045 &  & 10/10 & 4.712 & 3.483/0.138 \\
        \akGroupLine
        12 & MPD & 0/10 & \multicolumn{2}{c}{\akFail} &  & 0/10 & \multicolumn{2}{c}{\akFail} \\ \akLine
        12 & MMD & 10/10 & 11.102 & 1.972/0.042 &  & 10/10 & 14.013 & 3.135/0.113 \\ \akLine
        \akOurs 12 & \AkuleInlineLogo\akM{Akule} & 10/10 & 4.244 & 2.020/0.046 &  & 10/10 & 6.935 & 3.424/0.137 \\
        \akGroupLine
        15 & MPD & 0/10 & \multicolumn{2}{c}{\akFail} &  & 0/10 & \multicolumn{2}{c}{\akFail} \\ \akLine
        15 & MMD & 10/10 & 15.022 & 2.188/0.047 &  & 10/10 & 18.913 & 3.370/0.129 \\ \akLine
        \akOurs 15 & \AkuleInlineLogo\akM{Akule} & 10/10 & 7.209 & 2.143/0.049 &  & 10/10 & 11.822 & 3.670/0.156 \\
        \akGroupLine
        20 & MPD & 0/10 & \multicolumn{2}{c}{\akFail} &  & 0/10 & \multicolumn{2}{c}{\akFail} \\ \akLine
        20 & MMD & 10/10 & 22.434 & 2.146/0.047 &  & 6/10 & 42.965 & 3.856/0.179 \\ \akLine
        \akOurs 20 & \AkuleInlineLogo\akM{Akule} & 10/10 & 15.384 & 2.179/0.051 &  & 6/10 & 38.151 & 4.113/0.209 \\
        \akAccentLine
    \end{tabular}}
    \par\vspace{5pt}
    \def\akPanel{\dimexpr0.30\columnwidth\relax}
    \setlength{\fboxsep}{1.5pt}
    \def\akFail{{\bfseries\color{red!80!black}[FAIL]}}
    \begin{tabular}{@{}c@{\hspace{4pt}}c@{\hspace{4pt}}c@{}}
        \multicolumn{3}{c}{\akH{Highways, $N{=}20$}} \\[1pt]
        \colorbox{white}{\includegraphics[width=\akPanel]{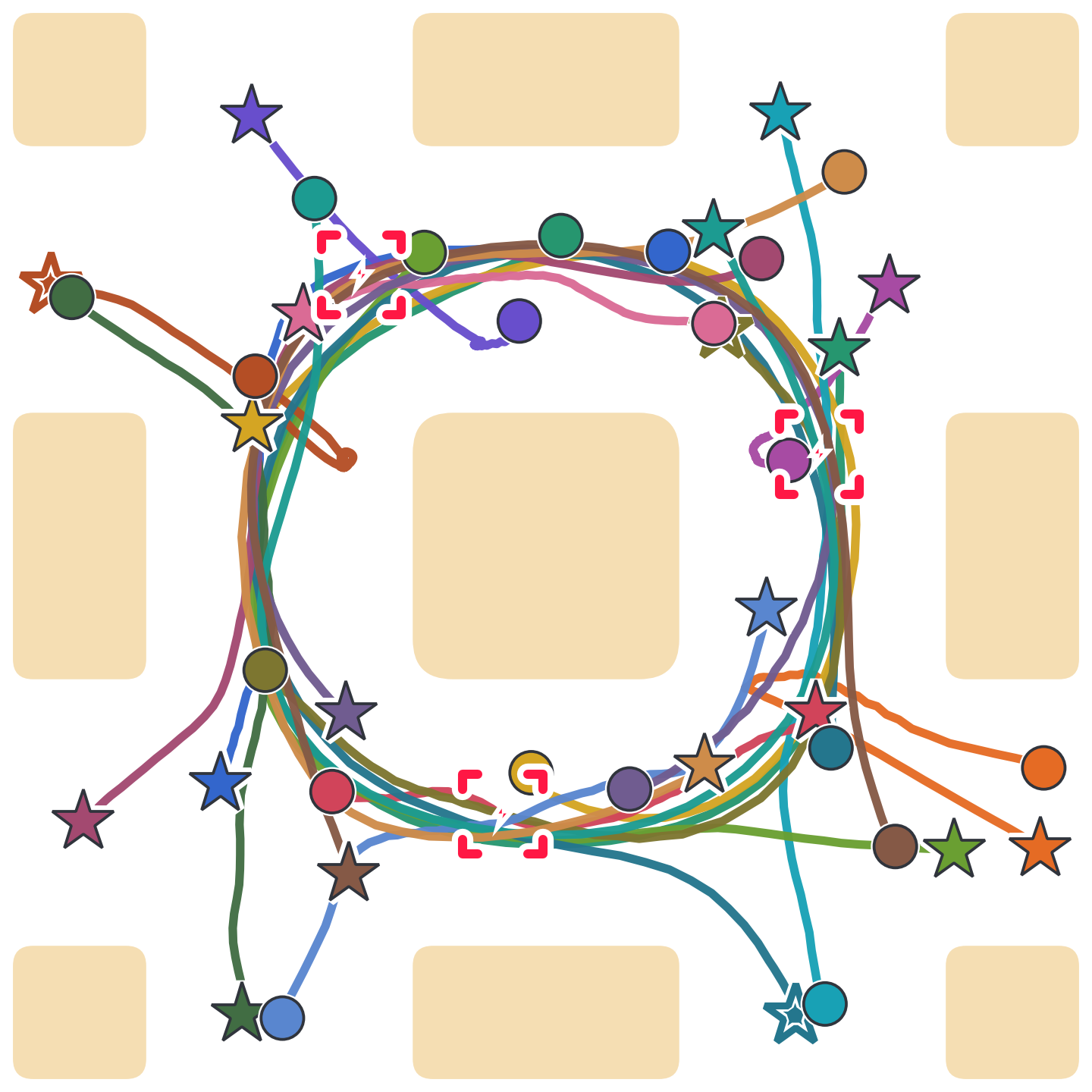}} & \colorbox{white}{\includegraphics[width=\akPanel,trim=0.00 0.00 0.00 0.00,clip]{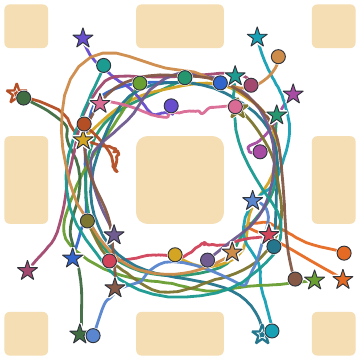}} & \colorbox{akTint}{\includegraphics[width=\akPanel]{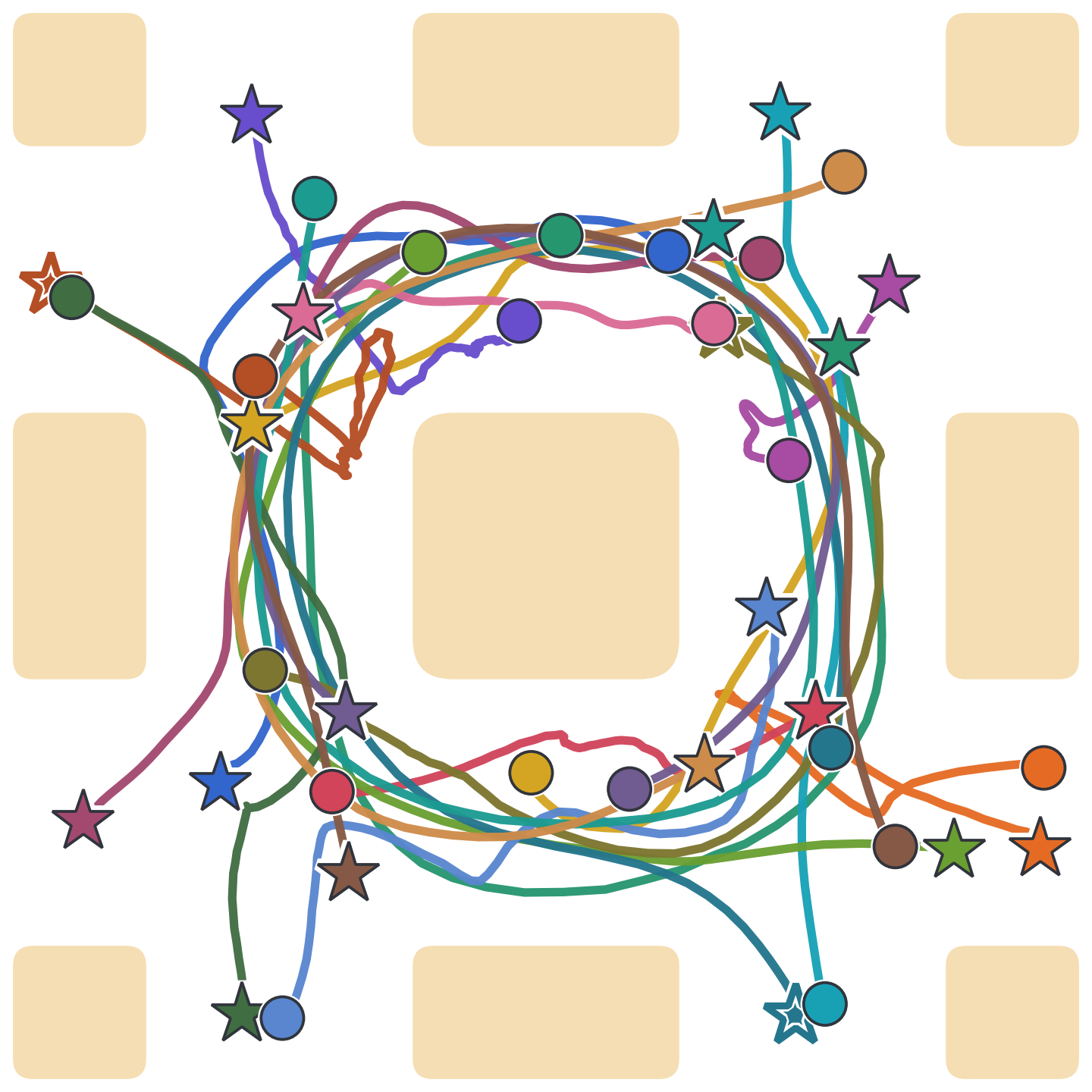}} \\[1pt]
        MPD~\akFail & MMD & \AkuleInlineLogo\akM{Akule} \\[4pt]
        \multicolumn{3}{c}{\akH{Conveyor, $N{=}20$}} \\[1pt]
        \colorbox{white}{\includegraphics[width=\akPanel]{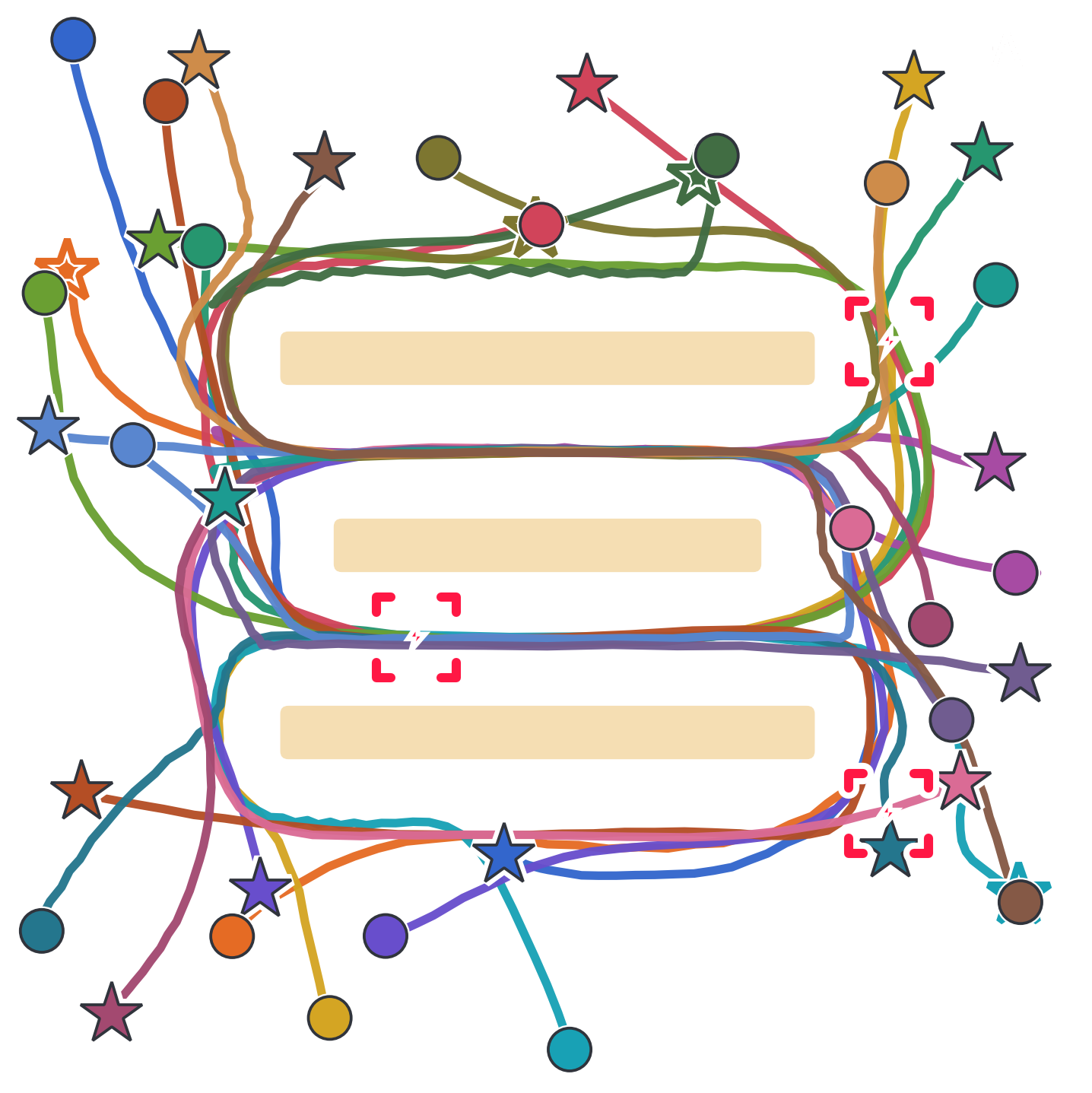}} & \colorbox{white}{\includegraphics[width=\akPanel,trim=0.00 0.00 0.47 0.47,clip]{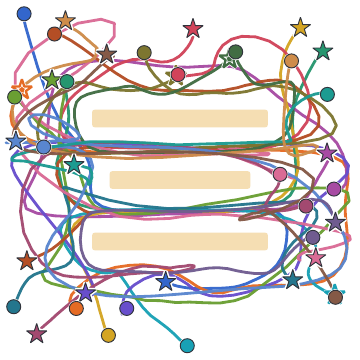}} & \colorbox{akTint}{\includegraphics[width=\akPanel]{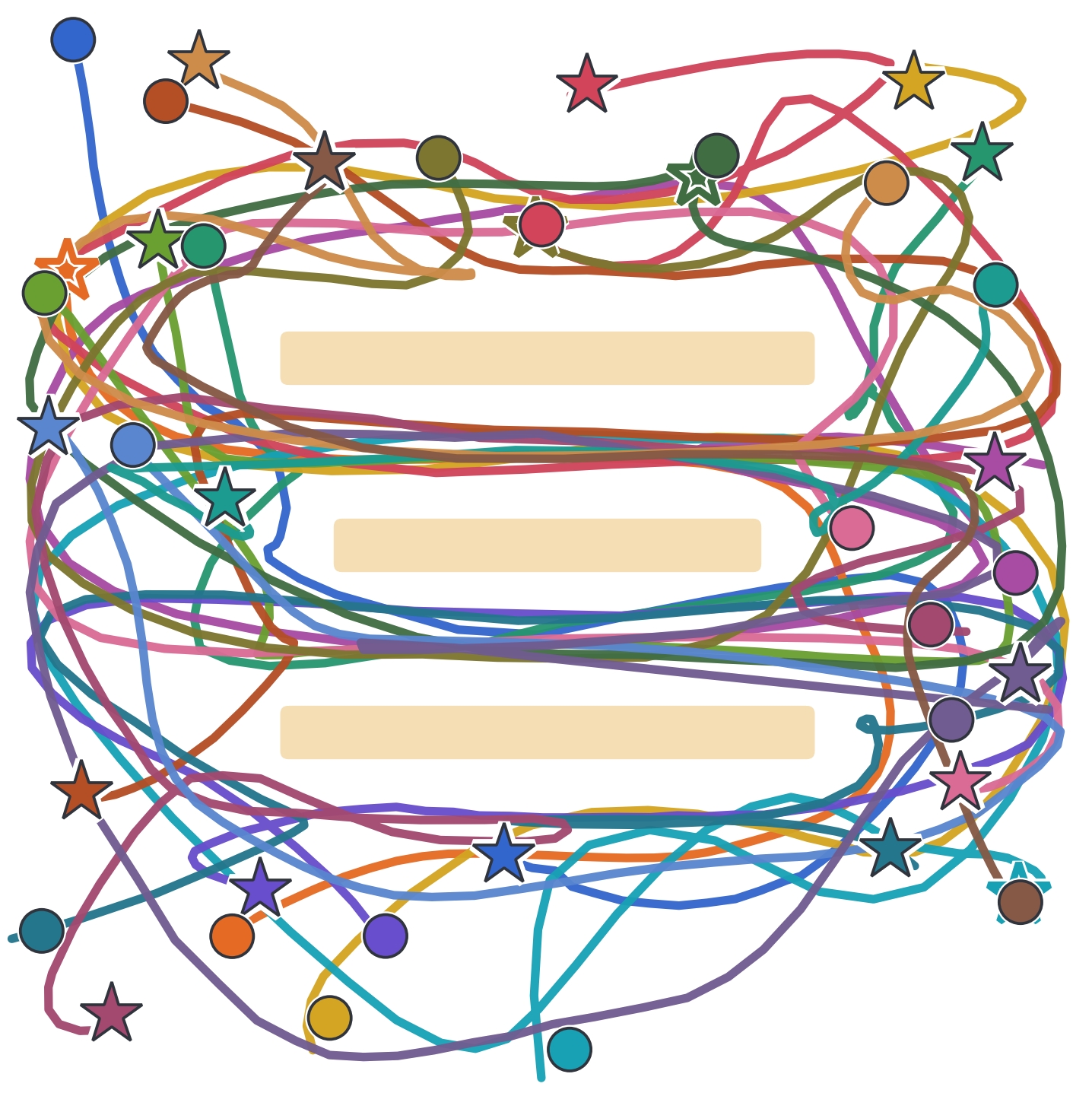}} \\[1pt]
        MPD~\akFail & MMD & \AkuleInlineLogo\akM{Akule} \\
    \end{tabular}
\end{table}

%% file: tables/tab_scaledweave.tex
\begin{table}[!ht]
    \centering
    \caption{ScaledWeave (10 scenes per $N$): the Weave geometry scaled to very large teams, with \method trained at $N{=}256$ and evaluated unchanged at smaller $N$.}
    \label{tab:scaledweave}
    \akStyle
    \setlength{\tabcolsep}{7pt}\arrayrulecolor{akRule}\begin{tabular}{c l c r}
        \akH{$N$} & \akH{Method} & \akH{Success} & \akH{Plan (s)} \\
        \akAccentLine
        40 & MPD & 8/10 & 219.7 \\ \akLine
        40 & MMD-xECBS & 10/10 & 176.4 \\ \akLine
        \akOurs 40 & \AkuleInlineLogo\akM{Akule (sparse)} & 10/10 & 32.1 \\
        \akGroupLine
        64 & MPD & 0/10 & $>$240 \\ \akLine
        64 & MMD-xECBS & 0/10 & $>$240 \\ \akLine
        \akOurs 64 & \AkuleInlineLogo\akM{Akule (sparse)} & 10/10 & 38.7 \\
        \akGroupLine
        100 & MPD & 0/10 & $>$240 \\ \akLine
        100 & MMD-xECBS & 0/10 & $>$240 \\ \akLine
        \akOurs 100 & \AkuleInlineLogo\akM{Akule (sparse)} & 10/10 & 48.5 \\
        \akGroupLine
        128 & MPD & 0/10 & $>$240 \\ \akLine
        128 & MMD-xECBS & 0/10 & $>$240 \\ \akLine
        \akOurs 128 & \AkuleInlineLogo\akM{Akule (sparse)} & 10/10 & 68.7 \\
        \akGroupLine
        256 & MPD & 0/10 & $>$240 \\ \akLine
        256 & MMD-xECBS & 0/10 & $>$240 \\ \akLine
        \akOurs 256 & \AkuleInlineLogo\akM{Akule (sparse)} & 8/10 & 143.4 \\
        \akAccentLine
    \end{tabular}
    \arrayrulecolor{black}
\end{table}

%% file: tables/fig_kf123.tex
\begin{figure*}[t]
    \centering
    \akStyle
    \begin{minipage}[t]{0.44\textwidth}
        \centering
        {\akH{(a) KF\#1}}\par\vspace{1pt}
        \includegraphics[width=\linewidth]{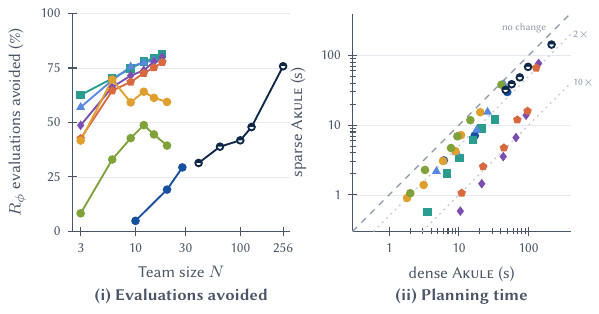}
    \end{minipage}\hfill
    \begin{minipage}[t]{0.27\textwidth}
        \centering
        {\akH{(b) KF\#2}}\par\vspace{1pt}
        \includegraphics[width=\linewidth]{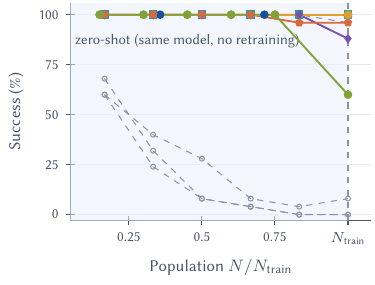}
    \end{minipage}\hfill
    \begin{minipage}[t]{0.27\textwidth}
        \centering
        {\akH{(c) KF\#3}}\par\vspace{1pt}
        \includegraphics[width=\linewidth]{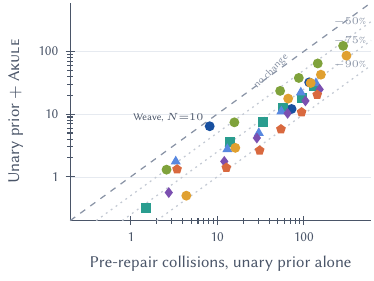}
    \end{minipage}
    \par\vspace{3pt}
    \includegraphics[width=0.9\textwidth]{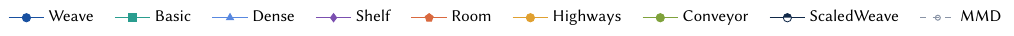}
    \caption{(a)~Sparse against dense \method on 44 matched tasks: (i)~fraction of $R_\phi$ evaluations skipped, (ii)~complete planning time. (b)~Success against population relative to the single training population ($N_{\rm train}=28$ on Weave, $18$ on the map families, $20$ on Highways and Conveyor). (c)~Pre-repair collisions per scene with and without \method's corrections, on matched scenes and noise before any repair (39 populations, 2{,}000 paired scenes); dotted guides mark 50/75/90\% reductions.}
    \label{fig:kf1}\label{fig:kf2}\label{fig:kf3}\label{fig:kf123}
\end{figure*}

%% file: tables/fig_kf4.tex
\begin{figure}[!ht]
    \centering
    \includegraphics[width=\columnwidth]{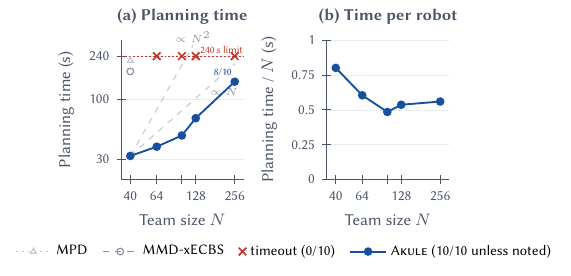}
    \caption{ScaledWeave (10 scenes per $N$): (a) complete planning time, with MPD and MMD-xECBS timing out on every scene from $N{=}64$; (b) planning time per robot.}
    \label{fig:kf4}
\end{figure}

%% file: tables/tab_ablations.tex
\begin{table}[!t]
    \centering

\caption{Ablation of $U_{\phi}$ and $G_{\omega}$ on Weave and Shelf. Conf.\ is mean pre-repair collision pair-times, and $R_{\rm ft}$ save is the fraction of pairwise corrections avoided relative to dense.}

    \label{tab:akule_ablations}
    \akStyle
    \akFit{\begin{tabular}{l<{\hspace{\akSlack}} c r r r c r}
        \akH{Method} &
        \akH{Success} &
        \akH{Conf.} &
        \akH{Repair (s)} &
        \akH{Plan (s)} &
        \akH{Quality (P/A)} &
        \akH{$R_{\rm ft}$ save} \\
        \akAccentLine

        \multicolumn{7}{l}{\textit{Canonical Weave, $N=28$}} \\
        \akLine
        \akOurs \AkuleInlineLogo\akM{Akule}
            & 30/30 & 32.133 & 22.901 & 29.944
            & 1.949/2.813 & 29.369\% \\ \akLine
        Akule (dense)
            & 30/30 & 32.797 & 25.163 & 43.571
            & 1.950/2.824 & 0\% \\ \akLine
        Raw/calibrated sum ($\alpha=1$)
            & 30/30 & 94.200 & 43.573 & 50.756
            & 1.935/2.602 & 29.740\% \\ \akLine
        Geometric selector
            & 30/30 & 59.167 & 32.766 & 39.929
            & 1.940/2.719 & 29.381\% \\ \akLine
        Collision-risk selector
            & 30/30 & 62.967 & 34.948 & 40.113
            & 1.949/2.794 & 29.381\% \\

        \akGroupLine
        \multicolumn{7}{l}{\textit{Shelf, $N=18$}} \\
        \akLine
        \akOurs \AkuleInlineLogo\akM{Akule} & 22/25 & 24.800 & 73.632 & 76.923 & 1.487 / 0.013 & 80.0\% \\
        \akLine
Akule (dense) & 24/25 & 19.200 & 71.051 & 81.495 & 1.512 / 0.014 & 0.0\% \\
\akLine
Raw sum ($\alpha=1$) & 21/25 & 61.960 & 83.179 & 86.835 & 1.556 / 0.016 & 81.8\% \\
\akLine
Calibrated sum ($\alpha=0$) & 20/25 & 152.800 & 103.958 & 107.744 & 1.517 / 0.018 & 79.6\% \\
\akLine
Geometric selector & 21/25 & 68.160 & 88.469 & 92.077 & 1.505 / 0.015 & 79.8\% \\
\akLine
Collision-risk selector & 21/25 & 43.240 & 79.909 & 82.403 & 1.504 / 0.013 & 79.8\% \\

        \akAccentLine
    \end{tabular}}
\end{table}

%% file: supplement/sections/architecture.tex
\section{Architecture and Implementation}
\label{sec:architecture}
This supplement specifies the implementation behind the sparse interaction model, its training dependencies, and the boundaries of the reported measurements. We retain the main paper's notation: $f_\theta$ is the frozen unary diffusion prior, $R_\phi$ the pairwise correction model, $U_\psi$ the selector, and $G_\omega$ the composer. A directed pair $(i,j)$ means that robot $j$ influences robot $i$. All three interaction modules share parameters across robots. Physical trajectory support $h$ and reverse-diffusion index $t$ are distinct: the standard trajectory has $H=64$ supports, while the reverse chain has $T=25$ transitions, indexed $24,\ldots,0$.

\subsection{State representation and frozen unary prior}
The grouped trajectory tensor is $B\times64\times N\times4$. Its physical channels are two positions and two stored displacements. They are not interchangeable with velocities in metres per second. The canonical codec is an unclipped affine transformation, $x^{\mathrm{phys}}=s\odot x^{\mathrm{norm}}+b$, with $s$ and $b$ fixed by the pretrained normalizer. Highways and Conveyor use a velocity bridge: the native prior's velocity channels are multiplied by $\Delta t=5/64$ to obtain displacement channels and divided by $\Delta t$ when encoded again. The conditioning tensor contains the start and goal four-channel states, hence eight endpoint values per robot. The start and goal positions and prescribed endpoint motion channels are reinstated after the reverse updates.

The canonical unary prior is the released MPD temporal U-Net~\cite{carvalho2023mpd,shaoul2025mmd}. It maps $64\times4$ noisy single-robot states to equally sized noise predictions. The base width is 32 and the channel multipliers are $(1,2,4,8)$, giving widths $(32,64,128,256)$. Each down stage has two residual temporal blocks, followed by downsampling except at the deepest stage; the bottleneck has two residual blocks, and three up stages concatenate encoder skip features. A residual block contains two kernel-5 convolutions with GroupNorm and Mish, an additive time projection, and a channel-matching residual connection. Downsampling uses a kernel-3 stride-2 convolution; upsampling uses a kernel-4 stride-2 transposed convolution. The output applies a kernel-5 block and a $1\times1$ convolution to four channels. The 32-dimensional time encoder uses a sinusoidal embedding and an MLP. The frozen configuration uses epsilon prediction and the 25-step exponential variance schedule. Highways and Conveyor load their own released unary checkpoints and normalizers with the same four-level width configuration.

For SMD maps, the frozen unary model adds goal-conditioned spatial features to the temporal backbone. A $64\times64$ raster contains occupancy, clipped signed distance, and start/goal Gaussian heatmaps with standard deviation $0.07$ m. Five spatial levels have widths $(16,24,32,48,64)$, two kernel-3 convolutions with SiLU per level, and stride-2 convolutions between levels. Bilinear samples from all five levels are concatenated and projected from 184 channels to 32 local features. Mean and maximum pooling of the deepest level produce 128 features projected to a 32-dimensional global token. Four exact local geometry features augment the local vector to 36 channels: scaled support clearance, its two-dimensional unit gradient, and scaled clearance at the following segment midpoint (the final support reuses its own clearance). Learned local and global projections inject this context into each temporal residual block. These parameters are adapted before coordination training and remain frozen during $R/G/U$ training and evaluation.

\subsection{Clean-state implementation of the pairwise correction}
The main paper expresses coordination as an additive epsilon correction. The implementation instead predicts a smooth \emph{physical clean-position} correction and converts the composed clean state through the usual diffusion posterior. Let $b_t^i$ denote the decoded, endpoint-conditioned unary clean estimate. The 29-channel pair sequence concatenates
\begin{equation}
\begin{split}
 \chi^{R}_{ij,t}(h)=\big[&b_t^i(h),b_t^j(h),e_i,e_j, p_t^j(h)-p_t^i(h),\\
 &\|p_t^j(h)-p_t^i(h)\|,t/24,h/63\big],
\end{split}
\end{equation}
where $e_i,e_j\in\mathbb R^8$. Thus the descriptor is computed from unary clean estimates of the noisy reverse state, not by directly feeding the normalized noisy states to every interaction module.

$R_\phi$ applies a $1\times1$ temporal convolution from 29 to 64 channels and SiLU, followed by four residual kernel-5 convolutions at dilations $1,2,4,8$, each with SiLU. Adaptive average pooling produces 12 control locations; a $1\times1$ head maps 64 channels to two position components. The decoder uses a cubic B-spline matrix constructed with 16 controls, retaining its 12 interior columns. Removing the first and last two controls enforces zero position correction and endpoint derivative. The deployed primitive subtracts the same network's null-partner prediction:
\begin{equation}
 d_{ij,t}=R_\phi(b_t^i,b_t^j,e_i,e_j,t)
             -R_\phi(b_t^i,0,e_i,0,t), \qquad d_{ij,t}\in\mathbb R^{64\times2}.
\end{equation}
This centering is part of the forward computation. An ``interaction evaluation'' in the reported savings counts an admitted directed primitive; it does not count the real and null network branches as two separate graph edges.

\subsection{Selector descriptor and contextual pooling}
The selector builds a 27-channel sequence for every candidate directed pair. It contains scaled ego and partner states (eight values), relative position and displacement (four), normalized separation minus one and an approach dot product (two), five geometric scene summaries for each endpoint robot (ten), a support-state scalar (one), and physical and reverse time (two). The position scale is $0.95$ m, the displacement scale is $0.5\Delta t$, and relative distance is scaled by $0.105$ m. At each support, the five scene summaries are nearest normalized separation, summed and maximum inverse-quadratic proximity, and a two-dimensional proximity-weighted relative direction. The support-state input is zero for one-shot deployed scoring. An elementwise $\operatorname{asinh}$ limits feature dynamic range without clipping.

A shared linear layer $27\to32$ with SiLU and a kernel-5 convolution $32\to32$ with SiLU encode the sequence. Within eight blocks of eight supports, mean and maximum pooling yield 64 features. These are concatenated with 24 transformed endpoint features $(e_i,e_j,e_j-e_i)$ and mapped $88\to48$ with SiLU. Mean and maximum pooling over the eight blocks, followed by $96\to48$ with SiLU, yields an edge token $h_{ij}$.

The abstract $\operatorname{Pool}$ in the main paper is implemented by learned-query attention pooling. Four shared inducing queries and four attention heads summarize (i) outgoing edges of the ego, (ii) incoming edges of the partner, and (iii) all directed edges. Self-edges are masked. Each four-query output is flattened and mapped $192\to48\to48$ with SiLU. A residual context update $192\to48\to48$ lets each edge incorporate its three set summaries; pooling is then repeated. The final contextual head is $192\to48\to1$ with SiLU. The frozen selector also has a shared local residual head $48\to192\to1$, whose output is added to the contextual logit. The extra per-agent density head supported by the code is not enabled in the canonical checkpoint. Deployment admits precisely $z_{ij,t}>0$, with no fixed degree cap.

\subsection{Composer and signed temporal weights}
$G_\omega$ encodes each admitted edge using the 29 pair features, the two residual components divided by $0.1$, and the gate, giving 32 input channels. Its temporal stack has width 112, a kernel-1 entry convolution, and four residual kernel-5 layers at dilations $1,2,4,8$. SiLU follows each convolution. Adaptive average pooling produces eight temporal blocks.

Two residual relational layers, each $336\to112\to112$ with SiLU, combine an edge token with normalized incident-edge sums at both robots. Gate-weighted aggregation excludes rejected interactions. For each ego, the subsequent set representation concatenates mean, maximum, and degree-square-root-normalized sums of 112-dimensional tokens; three corresponding residual-magnitude summaries; admitted degree divided by $N-1$; available degree divided by $N$; and population divided by 100. This $342$-dimensional vector maps to 112 features with SiLU. A node MLP $14\to112\to112$ consumes the four block-averaged state values, eight endpoint values, and two time values. Scene mean/max node pooling and population yield a $225\to112$ scene token.

The edge head concatenates the edge token, two node tokens, two set tokens, one scene token, and two time scalars. It maps $674\to112\to1$ with SiLU. For each admitted pair it outputs eight coefficients
\begin{equation}
 \beta_{ij,t,k}=s_t\tanh(\ell_{ij,t,k}),\qquad
 w_{ij,t}(h)=\sum_{k=1}^{8}B^{(8)}_{hk}\beta_{ij,t,k}.
\end{equation}
$B^{(8)}$ is the saved cubic B-spline basis, and $s_t>0$ is a fixed, training-calibrated trust bound. The main paper's scalar $\alpha_{ij,t}$ suppresses this dependence on physical support. The deployed update is
\begin{align}
 \delta p_i(h)&=\sum_{j\ne i}m_{ij,t}\,w_{ij,t}(h)d_{ij,t}(h),\\
 \widetilde b_t&=b_t+[\delta p,D\delta p],
\end{align}
where $D\delta p$ is the forward support difference with zero hard endpoint channels. Differencing the \emph{weighted sum} accounts for variation of $w$ along the trajectory. The codec then encodes $\widetilde b_t$ and applies the DDPM posterior. In normalized coordinates the corresponding epsilon change is $-\sqrt{\bar\alpha_t/(1-\bar\alpha_t)}\,\delta x_0$~\cite{ho2020ddpm}; physical position corrections must first pass through the codec and displacement construction.

\input{supplement/tables/architecture}
\subsection{Equivariance and computational boundary}
The specifications in Table~\ref{tab:architecture} are independent of population. No learned module uses a robot's array index as a feature. Shared pair encoders, masked set attention, symmetric reductions, and scatter-add composition commute with a simultaneous permutation of robots and their endpoints. The resulting trajectory update and directed mask therefore permute with the input. This is architectural permutation equivariance, up to floating-point reduction order. It does not imply rotation invariance, and the physical axes remain meaningful.

The selector still scores $N(N-1)$ candidates (the vectorized encoder also forms masked diagonal tokens). Sparse admission saves evaluations of $R_\phi$ and the admitted-edge composer; it does not make candidate scoring subquadratic. If no edge is admitted, composition returns a zero correction. The graph is recomputed at every interaction-aware reverse step, without a monotonic sparsification constraint.

\subsection{Complete planner and repair interface}
For Weave and SMD the complete proposal algorithm is: (1) sample the saved initial noise and condition endpoints; (2) for $t=24,\ldots,0$, predict unary clean states, encode selector descriptors, admit directed edges, evaluate centered pair corrections on those edges, compose signed corrections, and apply the conditioned posterior with the prescribed noise; (3) decode the final proposal; (4) perform the benchmark's static cleanup and hard repair; and (5) validate the returned complete trajectory. The posterior noise is zero at $t=0$.

For Highways and Conveyor, step (2) is replaced by independent candidate generation and selection followed by a \emph{single} $t=0$ interaction correction. Section~\ref{sec:protocol} specifies this distinction. Akule supplies a root trajectory to downstream conflict-based search~\cite{shaoul2025mmd}. The repair backend detects remaining conflicts, branches on constraints, and regenerates affected robot trajectories. SMD additionally uses targeted static repair with the native geometry and endpoint adapter; low-level RRT fallbacks are part of that implementation. Successful proposal generation alone is not complete-planner success, and the learned model provides no collision-free guarantee before repair.

%% file: supplement/tables/architecture.tex
\begin{table}[!htbp]
\caption{Frozen canonical interaction modules. Parameter counts exclude fixed spline bases and scale buffers.}
\label{tab:architecture}\centering\akStyle
\akFit{\begin{tabular}{l l r}
\akH{Module}&\akH{Output per directed pair}&\akH{Parameters}\\\akAccentLine
$R_\phi$ (width 64)&$64\times2$ correction&84,226\\\akLine
$U_\psi$ (width 48)&one logit&87,938\\\akLine
$G_\omega$ (width 112)&eight signed coefficients&509,601\\\akAccentLine
\end{tabular}}
\end{table}

%% file: supplement/sections/learning.tex
\section{Learning and Teacher Generation}
\label{sec:learning}
Learning follows the dependency order unary prior $\rightarrow R_\phi\rightarrow G_\omega\rightarrow U_\psi$. A later stage consumes frozen earlier modules. Physical trajectory teachers define useful coordination outcomes, not a uniquely correct interaction graph. We describe the optimization and supervision separately from checkpoint selection: a training stage can be executed without its last checkpoint being the deployed model.

\subsection{Matched pair optimization and residual supervision}
For a common unary clean estimate $b$, both pair optimizations start from the same positions, preserve the same endpoints, and use the same base objective. The interaction solve includes pair clearance; the null solve sets its weight to zero. The position variables are optimized by Adam at $0.01$ for 180 updates with projection to the workspace. Motion channels are reconstructed as forward position differences. The best observed objective is retained separately for each example. In the pair-target configuration,
\begin{equation}
 J=50C+0.2Q_{\mathrm{GP}}+0.2L+0.05Q_{\mathrm{trust}}+100Q_{\mathrm{box}},
 \label{eq:pairloss}
\end{equation}
with the $50C$ term removed in the null solve. Specifically,
\begin{align}
 C&=\frac1N\sum_{h,i<j}\left[\frac{(0.110-\|p_i(h)-p_j(h)\|)_+}{0.105}\right]^2,\\
 L&=\frac{1}{N\ell_0}\sum_{h,i}\|p_i(h+1)-p_i(h)\|,\\
 Q_{\mathrm{trust}}&=\frac{\operatorname{mean}_{h,i,d}[(p-b_p)^2]}{0.01},\\
 Q_{\mathrm{box}}&=\frac1N\sum_{h,i,d}\left[\frac{(|p_{id}(h)|-0.95)_+}{0.05}\right]^2.
\end{align}
The GP regularizer uses $v=w/\Delta t$, $e_p=p_{h+1}-p_h-\Delta t\,v_h$, and $e_v=v_{h+1}-v_h$. It averages over robots the sum over supports and coordinates of
\begin{equation}
 \frac{12}{\Delta t^3}e_p^2-\frac{12}{\Delta t^2}e_pe_v+
 \frac{4}{\Delta t}e_v^2,
\end{equation}
then divides by the fixed GP scale. The canonical residual checkpoint records $\ell_0=1.7466968662$ and GP scale $43.9410842702$. These differ from the composition-teacher scales below; they must not be exchanged.

If $p^{\mathrm{real}}$ and $p^{\mathrm{null}}$ are the optimized positions, the stored regression target is
\begin{equation}
 y=\Pi(p^{\mathrm{real}}-p^{\mathrm{null}}),\qquad
 \Pi=B_{\mathrm{int}}B_{\mathrm{int}}^{\dagger},
\end{equation}
where $B_{\mathrm{int}}$ contains the 12 interior cubic-spline columns. The noise-space target in the main paper follows by applying the codec and the DDPM clean/noise correspondence. In practice, supervision is mean squared \emph{position} error on the centered primitive, divided by the stored scale $1.415236620232463\times10^{-4}$. This avoids claiming that a physical displacement target is already a normalized epsilon target. Both pair directions are supervised. The canonical $R$ optimizer is Adam with learning rate $10^{-4}$, $\beta=(0.9,0.999)$, $\epsilon=10^{-8}$, and zero weight decay. Its replay fraction is zero. The checkpoint records update 6,000 and freezes the unary prior, codec, $G$, and $U$ during this residual adaptation.

The SMD residual corpus uses 16 training and two selection scenes per family over Empty, Basic, Dense, Shelf, and Room. Five unary reverse states are collected per scene. At each state up to eight unordered pairs are sampled from each minimum-distance stratum: below $0.110$ m, $[0.110,0.250)$ m, and at least $0.250$ m. Actual retained counts depend on pair qualification. The selected target policy is obstacle-agnostic: the matched pair optimization has an empty obstacle list, while the source unary trajectories remain map-conditioned. This separates the learned pair correction from the obstacle prior; it does not claim that all target trajectories satisfy the original map's obstacles. The trainer uses batches of 64: six examples from each of ten family/near--far bins and four rotating extra examples. Adam uses $10^{-4}$, with gradient norm clipping at 5 and rejection of nonfinite loss/gradients; checkpoints are saved every 500 updates. The selected SMD residual is update 6,500, selected with an additional disjoint composition panel rather than pair MSE alone.

\subsection{Offline dense composition teachers}
\label{app:dense_teachers}
The canonical teacher corpus contains 25,000 qualified geometries partitioned into 20,000 training, 2,500 selection, and 2,500 shadow geometries, with eight deterministic noise realizations per geometry. A frozen dense parent proposal is produced by a 25-step rollout. At its terminal clean-state interface, the optimizer adjusts eight signed coefficients for each directed frozen pair field. This is the concrete implementation of the main paper's trajectory-space composition optimization; it is not a separate free-trajectory replacement of the learned proposal.

The parameterization is $\beta=s_0q$ with $q\in[-1,1]$. The 25 trust scales are calibrated on training states alone using a $99.5$th percentile coefficient-magnitude rule and a $1.25$ safety multiplier; the terminal scale is $s_0=2.0591923185$. The complete scale vector is included in the compact supporting data. No scale is fit to final-test outcomes.

The optimizer uses float32 composition and gradients, transported to a float64 SciPy L-BFGS-B solver. Four penalty levels are $2\times10^3$, $2\times10^5$, $2\times10^7$, and $2\times10^9$, each with at most 80 iterations and 320 objective evaluations, $\texttt{ftol}=10^{-10}$, $\texttt{gtol}=10^{-8}$, and at most 30 line-search steps. For positions $p(q)$ it minimizes
\begin{equation}
\begin{split}
 J_\lambda(q)={}&\lambda C_{\mathrm{swept},0.120}+0.2Q_{\mathrm{GP}}+0.2L\\
 &+0.05Q_{\mathrm{trust}}+1000Q_{\mathrm{box}}+0.01Q_{\mathrm{guide}},
\end{split}
\end{equation}
where $C_{\mathrm{swept},0.120}$ sums squared clearance deficits divided by $0.01^2N$ at exact closest points of synchronized linear support intervals. $Q_{\mathrm{guide}}=\operatorname{mean}[(p-p^{\mathrm{guide}})^2]/0.01$. The GP and length scales here are $97.0556594884$ and $1.8758727280$. The guide is secondary, and the teacher always remains a composition of the frozen pair fields. The Weave guide is generated with ORCA and refined by 480 Adam updates at $0.01$.

Restart order is fixed: the initial signed fit, zero coefficients, a committed robust start, and, when needed, a guide fit and three seeded perturbations. A swept-constraint projection can refine an otherwise numerically stable candidate. Qualification requires sampled and swept separation of at least $0.120$ m, valid endpoints and workspace, finite values, and the stored motion envelope. Path, GP, acceleration, and jerk cannot exceed $1.10$ times their initial-reference values. Correction RMS is bounded by the larger of $1.5$ times its reference value and that value plus $0.03$ m. At most 20\% of coefficients may saturate; float32/float64 composition disagreement must have RMS at most $0.001$ m and maximum at most $0.003$ m. Qualified candidates are ordered by the original physical objective, guide fit, restart name, and penalty phase. Failed attempts do not enter the qualified corpus.

SMD teacher generation uses the same frozen-field principle with the map-specific physical contract. Its final training/selection manifest contains 204/26 qualified geometries per family, or 1,020/130 in total, with eight noise realizations. Geometry, obstacle, unary, residual, and teacher hashes are stored together. Thus these corpus sizes describe qualified examples and should not be confused with the number of attempted optimizations or the evaluation panel sizes.

\subsection{Composer transition bootstrap}
A teacher-consistent reverse chain is constructed from the optimized clean trajectory and saved noise. For each sampled step, the unary-only next state is subtracted from both the predicted and teacher next states. Let $\Delta y_t$ and $\Delta y_t^\star$ be these physical transition differences, separated into position and displacement channels $c\in\{p,w\}$. The implemented loss is
\begin{equation}
 \mathcal L_G^{\mathrm{trans}}=
 \mathbb E\left[w_t\sum_{c\in\{p,w\}}
 \frac{\operatorname{mean}[(\Delta y_{t,c}-\Delta y_{t,c}^{\star})^2]}
 {\max(10^{-6},\operatorname{mean}[(\Delta y_{t,c}^{\star})^2])}\right].
 \label{eq:transition}
\end{equation}
The denominator is detached. This is the normalized physical posterior-transition implementation of the abstract denoising-matching loss. The saved residual-confidence timestep weights are normalized to mean one. Timesteps are stratified over three portions of the 25-step schedule.

The composer is initialized freshly with a normally distributed final-head weight (standard deviation $0.01$) and zero bias. AdamW uses learning rates $3\times10^{-5}$ for the temporal/relational trunk and $10^{-4}$ for the remaining context/head parameters, weight decay $10^{-5}$, and gradient norm clipping at 5. A training update uses one geometry/noise condition. Fixed 16-scene sentinel and 128-scene selection panels monitor the canonical fit; the full corpus's shadow split is not used for checkpoint selection.

\subsection{Complete-rollout refinement}
The next stage propagates the composer's own outputs through all 25 reverse steps. Its loss is
\begin{align}
 \mathcal L_G^{\mathrm{roll}}={}&\operatorname{Huber}_{0.05}(p_G,p^\star)
 +\lambda_c C_{0.120}(p_G)+10^{-3}\mathcal L_G^{\mathrm{trans}},\\
 C_{0.120}(p)&=\frac1N\sum_{h,i<j}
 \left[\frac{(0.120-\|p_i(h)-p_j(h)\|)_+}{0.01}\right]^2,
\end{align}
with $\lambda_c=3.9508342635550774\times10^{-6}$. AdamW is reset and both learning rates are divided by ten. Autograd checkpointing limits memory, while gradients pass through the reverse states and frozen modules' computations; frozen parameters receive no optimizer updates. Nonfinite gradients fail the update, and the trainable gradient norm is clipped at 5. The canonical schedule cycles through 4,096 training geometries; the selected checkpoint is refinement update 3,290.

The SMD recipe also evaluates transition and rollout candidates, but its final repair-based selection retains transition-stage update 14,500. It would be inaccurate to infer a deployed rollout-refined composer solely from the existence of a refinement script. Selection prioritizes native complete-planner success, then repair and total time, then constraint-tree, low-level, and RRT workload on held-out scenes.

\subsection{Support optimization, selector bootstrap, and refinement}
With the unary, residual, and composer frozen, an offline tensor of edge logits is optimized with Adam at $0.03$ for at most 50 updates. Off-diagonal logits initialize to $0.01$. The objective adds expected mean directed degree to the physical teacher-matching and clearance loss:
\begin{equation}
 \mathcal L_U=\operatorname{Huber}_{0.05}(p_U,p^\star)
 +\lambda_c C_{0.120}(p_U)
 +\lambda_s\frac1{TN}\sum_{t,i\ne j}\sigma(z_{ij,t}),
\end{equation}
where $\lambda_s=1.885424111516031\times10^{-5}$. Hard masks are used forward, with
\begin{equation}
 m^{\mathrm{ST}}=\operatorname{sg}\!\left(\mathbf1[z>0]-\sigma(z)\right)+\sigma(z)
\end{equation}
for backpropagation. Dense fields provide gradient credit to currently rejected edges during optimization; inference evaluates admitted fields only. The gate multiplies a residual contribution once, avoiding an accidental squared straight-through gate. Candidate supports are checked at updates $0,5,10,25,50$ against goal completion and a 10\% motion-quality envelope. The support optimizer seeks a useful physical/sparsity trade-off, not an identifiable ground-truth graph.

The canonical target manifest records 256 attempted conditions and 130 valid targets. The first 128 valid conditions give $128\times25=3,200$ supervised states. Selector bootstrap uses a Huber loss with threshold $0.1$, separately averaged over retained and removed outgoing edges and eligible robots, weighting retained edges twice as strongly. AdamW uses $5\times10^{-4}$, weight decay $10^{-5}$, batches of eight, and gradient clipping at 5. Three sweeps give 1,200 optimizer updates; this is the deployed canonical selector checkpoint.

SMD generates 26 valid support conditions per family (130 total), from 154 attempts. Its selector additionally undergoes complete-rollout straight-through refinement with the unary, $R$, and $G$ frozen. It uses AdamW at $5\times10^{-5}$, weight decay $10^{-5}$, and clip norm 5. Update 1,000 is selected from the fixed validation frontier through complete-planner evaluation. The refinement uses the same physical loss and degree penalty. The shadow panel is consulted only after freezing the selected checkpoint.

\subsection{Map adaptation and available compute measurements}
The Highways/Conveyor adaptation interface operates at $N=20$ on terminal selected-candidate states. The saved training contract specifies 128 training and 32 validation scenes per map, a shared composer without environment-ID features, bounded coefficient teachers with 240 Adam steps at $0.025$ and two starts, and a native adherence guard. Transition bootstrap uses the canonical optimizer for 4,000 updates and terminal-rollout refinement for 400; validation is separate from the final ten-scene-per-population panels. This terminal interface has posterior coefficients $c_1=1,c_2=0$ and zero posterior noise. It must not be interpreted as training or applying 25 interaction corrections to these map candidates.

The final batched map pipeline continues the shared composer for 1,500 updates (1,000 transition updates followed by 500 physical-refinement updates) on 128 qualified training scenes per map. It freezes composer update 1,500 and selector update 200. Selector adaptation uses the same balanced Huber loss, AdamW at $5\times10^{-4}$, decay $10^{-5}$, clip norm 5, and batches containing four Highways and four Conveyor states. The budget is 1,000 updates with checks every 100 and stopping after two nonimproving checks; complete-planner selection uses five disjoint validation tasks per map. The frozen residual remains unchanged.

End-to-end training and teacher-generation compute is not uniformly recorded, so we do not provide a total training-cost claim. Two canonical checkpoint records contain partial elapsed times: 3,820.15 s through the selected $G$ refinement checkpoint and 11.89 s for the three-sweep $U$ fit. These exclude prior training, teacher generation, and other selection runs, and are not a matched hardware comparison. The SMD unary-intervention replay's 130.14 s for 600 paired scenes is an evaluation measurement, not training cost.

%% file: supplement/sections/protocol.tex
\section{Experimental Protocol}
\label{sec:protocol}
The benchmarks share the sparse-interaction decomposition but use different proposal interfaces and native validation contracts. Table~\ref{tab:protocol} summarizes the evaluated grids. Population transfer means reusing a frozen coordination model at the stated test population; it does not imply an unchanged environment-specific unary prior or identical downstream planning software across benchmark families.
\begin{table*}[t]
\caption{Evaluation populations and proposal interfaces. Scene counts are per population, except the three explicitly listed Weave counts. The 39-condition intervention study comprises the first seven environments (2,000 paired scenes). ScaledWeave is a separate ten-scene-per-population family. Source: frozen scene and runtime manifests.}
\label{tab:protocol}\centering\akStyle
\akFit{\begin{tabular}{l l l l}
\akH{Environment}&\akH{Evaluated populations $N$}&\akH{Scenes per $N$}&\akH{Interaction interface}\\\akAccentLine
Canonical Weave&10, 20, 28&512, 256, 512&25 reverse steps\\\akLine
Highways&3, 6, 9, 12, 15, 20&10&one correction at $t=0$\\\akLine
Conveyor&3, 6, 9, 12, 15, 20&10&one correction at $t=0$\\\akLine
SMD Basic / Dense / Shelf / Room&3, 6, 9, 12, 15, 18&25 per environment&25 reverse steps\\\akLine
ScaledWeave&40, 64, 100, 128, 256&10&separate scaled geometry and physical clock\\\akAccentLine
\end{tabular}}
\end{table*}

\subsection{Scenes, geometry, and physical representation}
\paragraph{Canonical Weave.} Robots exchange opposite sides of the fixed obstacle-free square $[-1,1]^2$ m. The construction starts on a perturbed angular ring and snaps a dominant coordinate to the square-side boundary; the goal reflects that coordinate. Robot radius is $0.05$ m, and the frozen center bound is $0.95$ m. The coordination model is trained at $N=28$ and transferred to $N=10,20$. Weave's fixed construction is retained rather than enlarged for higher populations. The $N=28$ panel is the predeclared first 512 executed conditions of a 1,024-condition frozen manifest. The smaller-population panels contain 512 and 256 scenes, respectively. Learned/unary comparisons share deterministic noise; the native MMD path uses the recorded compatible seed convention.

\paragraph{Highways and Conveyor.} These use the released MMD obstacle geometry, disk radius $0.05$ m, and workspace $[-1,1]^2$ m. Obstacles are rounded boxes with the exact saved centers, full sizes, and corner radii, not rectangular approximations inferred from images. Each of the six populations has ten frozen start/goal tasks per map. Adaptation is performed at $N=20$ where applicable; the evaluation does not fit a separate coordination network at each smaller $N$. The environment-specific single-robot prior produces 64 independent candidates per robot. Its frozen runtime uses 64 supports, 20 guide steps with guidance starting at reverse index 13, one final noise-free step, and three noising/three denoising steps for constrained local inference. The supporting runtime configuration includes the exact four-channel normalizer limits. Akule batches candidate generation across robots and selects candidates before applying one terminal interaction correction. The reference MPD/MMD paths may generate robot candidates sequentially. In particular, MMD's roots already incorporate soft inter-robot guidance; they are not independent unary trajectories.

Consequently, the map experiments are not 25-step interaction rollouts. Their priors use the diffusion schedule during candidate generation, but $U/R/G$ operate once on the selected candidates at $t=0$. Candidate batching and guidance change sampler cost independently of interaction selection. Comparisons of complete planning time evaluate these implemented pipelines, while the matched unary-intervention study isolates changes to the same selected candidates.

\paragraph{SMD maps.} Basic, Dense, Shelf, and Room use the released instance obstacle circles~\cite{liang2025smd}. Shelf and Room partitions are represented by the actual circle unions, not substituted rectangles. A shared map-conditioned unary and the frozen $R/G/U$ stack are trained/adapted using $N=18$ contexts and transferred across the six test populations. Each final cell contains 25 held-out instances with radius $0.05$ m. Training also includes Empty, but the seven-environment intervention table reports the four named obstacle families. The figure data include exact layout and scene identities. SMD representative baseline reruns are kept distinct from any published aggregate reference, and a displayed example is not itself a 25-scene estimate.

All seven intervention environments record 64 trajectory supports at $\Delta t=5/64=0.078125$ s, with nominal duration 5 s. The first-to-last support span is $63\Delta t$, which is distinct from the nominal horizon. Endpoints remain conditioned throughout the 25-step exponential diffusion schedule in the full-rollout implementation.

\paragraph{ScaledWeave.} The square workspace has half-extent $s(N)$ metres, where
\begin{equation}
 s(N)=\max\!\left(1,\frac{1.001(0.150)N\sqrt2}{0.8(2\pi)(1-0.005)}\right).
\end{equation}
Construction radius is $0.8s$ and the snapped coordinate is $0.87s$. Angular phase is sampled within $\pm0.005$ radians and per-robot jitter within $\pm0.0025(2\pi/N)$. The minimum required endpoint separation is $0.150$ m; the disk radius remains $0.05$ m. This preserves boundary packing rather than constant areal density. The saved proposal study uses nominal duration $5s(N)$ seconds and physical timestep $5s(N)/64$ seconds, with 64 supports and 25 reverse steps. Positions and displacements are mapped through the similarity frame; the physical disk radius remains fixed. 

\subsection{Success, failure, and latency}
\paragraph{Complete-planner success.} A run must return a complete team solution and meet its benchmark's final native validation and endpoint requirements within the applicable time budget. On canonical Weave, the frozen contract requires all goals within $0.05$ m and zero sampled pair-time conflicts at strict physical separation below $0.100$ m. Native MMD search also has its own conflict predicate; search-node conflict counts are not interchangeable with the physical pair-time diagnostic.

The released SMD sampled pair/obstacle predicate compares squared distance with the squared sum of radii minus $0.001$. Goal and endpoint checks also apply. This is distinct from a strict disk-overlap diagnostic and from a swept-path validation. We retain native success as the SMD benchmark contract and label physical diagnostics separately. No sampled check proves continuous collision freedom between supports. Swept checks used for teacher qualification are a stronger and separately stated requirement.

\paragraph{Latency.} Let $M$ be the attempted scene count, $\ell$ its planning limit, and $t_s$ its recorded planning time. The reported all-attempt mean is
\begin{equation}
 \overline t=\frac1M\sum_{s=1}^{M}
 \begin{cases}\ell,&\text{run }s\text{ reaches its configured timeout},\\
 t_s,&\text{otherwise, including terminal failures}.
 \end{cases}
\end{equation}
Canonical Weave uses a 240 s native limit (255 s external watchdog); SMD Akule complete-planner evaluation uses 900 s. Highways/Conveyor main-panel reporting uses a 60 s budget; longer diagnostic requests are not additional successful attempts under that budget. ScaledWeave artifacts must retain their own configured budget and proposal/complete-planner scope, rather than inheriting the map limit. Failure remains in the denominator. Model loading, compilation, warmup, serialization, and visualization are excluded where the native benchmark timer excludes them; proposal generation, static repair, and hard search belong to online planning. New backend measurements in Section~\ref{sec:additional} use an explicitly fixed 240 s contract.

\paragraph{Trajectory quality.} Mean path length averages summed Euclidean support-to-support distances over robots and then successful scenes. Canonical acceleration RMS is the root mean square norm of second position differences divided by $\Delta t^2$; jerk analogously uses third differences divided by $\Delta t^3$. SMD's reported per-step acceleration statistic is the mean norm of second position differences without this physical-time rescaling. Native MMD quality follows its released evaluator and velocity convention. These differently normalized quantities should not be compared across benchmark families as though they were one physical statistic. Undefined quality or makespan for failed runs is omitted, never set to zero. Makespan, when available, is sustained goal arrival computed from saved supports, not a physical robot execution measurement.

\subsection{Frozen models and baseline scope}
The supporting provenance data identify the four canonical checkpoints and four SMD checkpoints by SHA256; changing population does not change those checkpoint identities. Map evaluation also pins the environment-specific priors, normalizers, and adapted composer/selector files in its final model manifest. Baseline names refer to executable pipelines: independent MPD proposals, native MMD with its specified CBS-family search, and simultaneous projected diffusion (SMD). A saved MPD \emph{proposal} with collisions may be displayed even when MPD plus subsequent repair succeeds; captions identify that stage. The main manuscript\textquotesingle s SMD-map MPD, MMD, SMD, and DGD aggregate references come from the published DGD benchmark~\cite{liang2025dgd}; their acceleration formula is not supplied by that aggregate table. The displayed SMD trajectories are local representative reruns, not reconstructed published trajectories. Published-only aggregates are not treated as locally paired reruns or used for hardware speed ratios. The supplementary component timings and conflict counts come from the frozen local panels and explicitly state their scope.

%% file: supplement/sections/quantitative.tex
\section{Additional Quantitative Results}
\label{sec:quantitative}
\subsection{Matched unary intervention}
We compare Akule with its \emph{own} unary process before any smoothing, static cleanup, or hard repair. This controls for the input scene and unary prior, and avoids attributing differences between candidate-generation pipelines to interaction learning. Canonical Weave compares saved unary-only and interaction-aware 25-step rollouts. Highways/Conveyor compare the same selected independent candidate trajectories before and after the terminal correction. SMD uses matched 25-step proposal-only replays with identical initial and posterior noise tensors and frozen checkpoints; no hard repair is invoked. The replay verification records agreement with the frozen Akule pair-conflict, residual-evaluation, and native static-invalid counts for all 600 SMD cases.

For positions $p_i(h)$ in metres, collision pair-times are
\begin{equation}
 C(p)=\sum_{h=0}^{63}\sum_{1\leq i<j\leq N}
 \mathbf1\!\left[\|p_i(h)-p_j(h)\|_2<0.100\right].
 \label{eq:collision}
\end{equation}
Distances are computed from float64 positions at the recorded supports. An unordered pair contributes once per colliding support. This is neither a unique-pair count nor an interpolated collision-duration measure, and it excludes obstacle contacts. Equality at $0.100$ m does not count. The intervention is paired by scene identity and inputs; the full-rollout arms additionally share noise. Complete-planner failures remain eligible because the quantity measured is the pre-repair proposal.

Table~\ref{tab:unary_largest} gives the largest population in each of seven environments. Table~\ref{tab:unary_all} preserves all 39 conditions and all 2,000 paired scenes, with no missing pairs. The reported percentage is a reduction in the \emph{mean} count, not the average of per-scene percentages (which would be undefined on zero-conflict unary scenes).
\input{supplement/tables/unary_largest}
\input{supplement/tables/unary_all}

Every condition has a positive mean reduction. At the largest populations, reductions range from 56.9\% on Conveyor to 85.8\% on Room. This does not imply per-scene dominance: Weave $N=10$ has 33/512 scenes with increased conflicts, Conveyor $N=6$ has 1/10, and Room $N=3$ has 1/25. The weaker mean improvements---21.8\% at Weave $N=10$ and 47.0\% at Dense $N=3$---are retained. The table supports a proposal-stage coordination benefit under the stated pairing, but does not by itself establish obstacle feasibility, complete success, or a fixed latency improvement.

\subsection{Extended proposal-stage measurements}
Table~\ref{tab:stages} separates independent candidates, interaction-corrected proposals, static cleanup, and roots entering hard search. Static cleanup can increase pair conflicts because it repairs obstacles one robot at a time. On Highways $N=20$, for example, Akule's mean increases from 85.7 after interaction to 86.6 after static cleanup. MMD's soft-guided roots contain fewer residual conflicts than Akule's corrected map proposals: 13.7 versus 85.7 on that condition. At Highways $N=3,6$, all ten MMD roots are conflict-free. These results are consistent with MMD spending computation on sequential soft coordination before hard search.
\input{supplement/tables/stages}

Conveyor $N=20$ has only six complete MMD roots. The other four attempts return incomplete teams and cannot supply full-team root conflict counts. Their planning failures and recorded computation remain part of complete-planner accounting, while root comparisons explicitly restrict to available complete roots. We do not impute missing agents, zero conflicts, or ten complete MMD proposals. The independent unary candidates used in Table~\ref{tab:unary_all} are all available and are a different baseline from these soft-guided roots.

%% file: supplement/tables/unary_largest.tex
\begin{table}[!htbp]
\centering\caption{Largest-population unary intervention results. Mean physical collision pair-times before postprocessing; reduction and F/E/M follow Table~\ref{tab:unary_all}.}
\label{tab:unary_largest}\akStyle
\akFit{\begin{tabular}{l r r r r r r}
\akH{Environment} & \akH{$N$} & \akH{Scenes} & \akH{Unary} & \akH{Akule} & \akH{$\Delta$} & \akH{F/E/M}\\\akAccentLine
Weave & 28 & 512 & 115.613 & \cellcolor{akTint}32.133 & 72.2\% & 512/0/0\\\akLine
Highways & 20 & 10 & 311.600 & \cellcolor{akTint}85.700 & 72.5\% & 10/0/0\\\akLine
Conveyor & 20 & 10 & 285.300 & \cellcolor{akTint}122.900 & 56.9\% & 10/0/0\\\akLine
SMD Basic & 18 & 25 & 129.960 & \cellcolor{akTint}28.120 & 78.4\% & 25/0/0\\\akLine
SMD Dense & 18 & 25 & 138.240 & \cellcolor{akTint}31.160 & 77.5\% & 25/0/0\\\akLine
SMD Shelf & 18 & 25 & 152.800 & \cellcolor{akTint}24.800 & 83.8\% & 25/0/0\\\akLine
SMD Room & 18 & 25 & 143.160 & \cellcolor{akTint}20.400 & 85.8\% & 25/0/0\\\akLine
\akAccentLine
\end{tabular}}
\end{table}

%% file: supplement/tables/unary_all.tex
\begin{table*}[t]
\centering\caption{All 39 evaluated population conditions; no population or paired scene is omitted. Mean unordered physical collision pair-times before smoothing, static cleanup, and hard repair. Reduction is $100(1-\bar C_{\mathrm{Akule}}/\bar C_{\mathrm{unary}})$. F/E/M counts scenes with fewer/equal/more conflicts after interaction. Provenance: frozen paired-intervention export; 64 supports, strict distance $<0.100$ m.}
\label{tab:unary_all}\akStyle
\akFit{\begin{tabular}{l r r r r r r}
\akH{Environment} & \akH{$N$} & \akH{Pairs} & \akH{Unary} & \akH{Akule} & \akH{Reduction} & \akH{F/E/M}\\\akAccentLine
Canonical Weave & 10 & 512 & 8.195 & \cellcolor{akTint}6.410 & 21.8\% & 430/49/33\\\akLine
Canonical Weave & 20 & 256 & 72.379 & \cellcolor{akTint}12.082 & 83.3\% & 256/0/0\\\akLine
Canonical Weave & 28 & 512 & 115.613 & \cellcolor{akTint}32.133 & 72.2\% & 512/0/0\\\akLine
Highways & 3 & 10 & 4.400 & \cellcolor{akTint}0.500 & 88.6\% & 7/3/0\\\akLine
Highways & 6 & 10 & 16.200 & \cellcolor{akTint}2.900 & 82.1\% & 9/1/0\\\akLine
Highways & 9 & 10 & 66.100 & \cellcolor{akTint}17.700 & 73.2\% & 10/0/0\\\akLine
Highways & 12 & 10 & 120.300 & \cellcolor{akTint}31.400 & 73.9\% & 10/0/0\\\akLine
Highways & 15 & 10 & 157.800 & \cellcolor{akTint}42.600 & 73.0\% & 10/0/0\\\akLine
Highways & 20 & 10 & 311.600 & \cellcolor{akTint}85.700 & 72.5\% & 10/0/0\\\akLine
Conveyor & 3 & 10 & 2.600 & \cellcolor{akTint}1.300 & 50.0\% & 5/5/0\\\akLine
Conveyor & 6 & 10 & 15.800 & \cellcolor{akTint}7.400 & 53.2\% & 9/0/1\\\akLine
Conveyor & 9 & 10 & 53.400 & \cellcolor{akTint}23.500 & 56.0\% & 10/0/0\\\akLine
Conveyor & 12 & 10 & 87.700 & \cellcolor{akTint}37.800 & 56.9\% & 10/0/0\\\akLine
Conveyor & 15 & 10 & 145.200 & \cellcolor{akTint}64.400 & 55.6\% & 10/0/0\\\akLine
Conveyor & 20 & 10 & 285.300 & \cellcolor{akTint}122.900 & 56.9\% & 10/0/0\\\akLine
SMD Basic & 3 & 25 & 1.520 & \cellcolor{akTint}0.320 & 78.9\% & 6/19/0\\\akLine
SMD Basic & 6 & 25 & 14.040 & \cellcolor{akTint}3.560 & 74.6\% & 22/3/0\\\akLine
SMD Basic & 9 & 25 & 34.040 & \cellcolor{akTint}7.440 & 78.1\% & 25/0/0\\\akLine
SMD Basic & 12 & 25 & 57.080 & \cellcolor{akTint}12.520 & 78.1\% & 25/0/0\\\akLine
SMD Basic & 15 & 25 & 95.880 & \cellcolor{akTint}17.960 & 81.3\% & 25/0/0\\\akLine
SMD Basic & 18 & 25 & 129.960 & \cellcolor{akTint}28.120 & 78.4\% & 25/0/0\\\akLine
SMD Dense & 3 & 25 & 3.320 & \cellcolor{akTint}1.760 & 47.0\% & 9/16/0\\\akLine
SMD Dense & 6 & 25 & 13.120 & \cellcolor{akTint}2.760 & 79.0\% & 19/6/0\\\akLine
SMD Dense & 9 & 25 & 30.280 & \cellcolor{akTint}5.000 & 83.5\% & 24/1/0\\\akLine
SMD Dense & 12 & 25 & 55.120 & \cellcolor{akTint}11.040 & 80.0\% & 25/0/0\\\akLine
SMD Dense & 15 & 25 & 92.600 & \cellcolor{akTint}22.120 & 76.1\% & 25/0/0\\\akLine
SMD Dense & 18 & 25 & 138.240 & \cellcolor{akTint}31.160 & 77.5\% & 25/0/0\\\akLine
SMD Shelf & 3 & 25 & 2.760 & \cellcolor{akTint}0.560 & 79.7\% & 14/11/0\\\akLine
SMD Shelf & 6 & 25 & 12.080 & \cellcolor{akTint}1.760 & 85.4\% & 23/2/0\\\akLine
SMD Shelf & 9 & 25 & 28.720 & \cellcolor{akTint}4.160 & 85.5\% & 25/0/0\\\akLine
SMD Shelf & 12 & 25 & 65.120 & \cellcolor{akTint}10.280 & 84.2\% & 25/0/0\\\akLine
SMD Shelf & 15 & 25 & 104.560 & \cellcolor{akTint}16.280 & 84.4\% & 25/0/0\\\akLine
SMD Shelf & 18 & 25 & 152.800 & \cellcolor{akTint}24.800 & 83.8\% & 25/0/0\\\akLine
SMD Room & 3 & 25 & 3.440 & \cellcolor{akTint}1.320 & 61.6\% & 11/13/1\\\akLine
SMD Room & 6 & 25 & 12.760 & \cellcolor{akTint}1.400 & 89.0\% & 21/4/0\\\akLine
SMD Room & 9 & 25 & 30.960 & \cellcolor{akTint}2.640 & 91.5\% & 25/0/0\\\akLine
SMD Room & 12 & 25 & 54.560 & \cellcolor{akTint}5.760 & 89.4\% & 25/0/0\\\akLine
SMD Room & 15 & 25 & 94.520 & \cellcolor{akTint}10.760 & 88.6\% & 25/0/0\\\akLine
SMD Room & 18 & 25 & 143.160 & \cellcolor{akTint}20.400 & 85.8\% & 25/0/0\\\akLine
\akAccentLine
\end{tabular}}
\end{table*}

%% file: supplement/tables/stages.tex
\begin{table}[!htbp]
\centering\caption{Map proposal-stage mean collision pair-times. Akule uses ten scenes per cell; the Conveyor $N=20$ MMD root mean uses its six available complete roots.}
\label{tab:stages}\akStyle
\akFit{\begin{tabular}{l r r r r r r}
\akH{Map} & \akH{$N$} & \akH{Unary} & \akH{Corrected} & \akH{After static} & \akH{MMD root} & \akH{Roots}\\\akAccentLine
Highways & 3 & 4.4 & \cellcolor{akTint}0.5 & 0.4 & 0.0 & 10/10\\\akLine
Highways & 6 & 16.2 & \cellcolor{akTint}2.9 & 2.8 & 0.0 & 10/10\\\akLine
Highways & 9 & 66.1 & \cellcolor{akTint}17.7 & 18.6 & 1.2 & 10/10\\\akLine
Highways & 12 & 120.3 & \cellcolor{akTint}31.4 & 30.0 & 2.7 & 10/10\\\akLine
Highways & 15 & 157.8 & \cellcolor{akTint}42.6 & 42.6 & 4.2 & 10/10\\\akLine
Highways & 20 & 311.6 & \cellcolor{akTint}85.7 & 86.6 & 13.7 & 10/10\\\akLine
Conveyor & 3 & 2.6 & \cellcolor{akTint}1.3 & 1.1 & 0.4 & 10/10\\\akLine
Conveyor & 6 & 15.8 & \cellcolor{akTint}7.4 & 7.1 & 1.5 & 10/10\\\akLine
Conveyor & 9 & 53.4 & \cellcolor{akTint}23.5 & 26.0 & 7.0 & 10/10\\\akLine
Conveyor & 12 & 87.7 & \cellcolor{akTint}37.8 & 36.7 & 9.2 & 10/10\\\akLine
Conveyor & 15 & 145.2 & \cellcolor{akTint}64.4 & 68.1 & 16.8 & 10/10\\\akLine
Conveyor & 20 & 285.3 & \cellcolor{akTint}122.9 & 145.4 & 34.0 & 6/10\\\akLine
\akAccentLine
\end{tabular}}
\end{table}

%% file: supplement/sections/ablations.tex
\section{Ablation Details}
\label{sec:ablations}
The final ablation panel uses Canonical Weave $N=28$ (30 instances) and Shelf $N=18$ (25 instances). The variants are inference-time substitutions with frozen unary and pairwise models, the same diffusion schedule, and the same downstream repair interface. They are not separately retrained neural architectures. Table~\ref{tab:ablations} reproduces the final manuscript's ablation results, including success, pre-repair conflicts, repair and complete latency, successful-trajectory quality, and directed-interaction savings.
\input{supplement/tables/ablations}

\paragraph{Composition substitutions.} Full Akule uses learned hard selection and signed contextual composition. Dense Akule opens every non-self directed edge while retaining the learned composer. Raw summation retains the learned selector and gives every admitted primitive coefficient one. Calibrated summation retains the selector and uses a single constant from $\{0,0.25,0.5,0.75,1,1.5,2\}$, selected on disjoint validation scenes by complete proposal rollouts.

Weave validation lexicographically minimizes mean pre-repair pair conflicts, then acceleration and path length. The first criterion is smallest at $\alpha=1$ (94.6875 validation pair-times), narrowly ahead of $\alpha=0.75$ (95.21875), so calibrated and raw summation are the same intervention and share measurements. Shelf first minimizes obstacle-invalid robots, then pair conflicts, acceleration, and path length. The $\alpha=0$ validation candidates have zero obstacle-invalid robots, whereas every positive grid coefficient has a positive mean. The rule therefore selects $\alpha=0$ even though its pair-conflict mean is 165.5769. This is a physical-feasibility-first validation decision, not a choice minimizing complete-planner latency. A zero coefficient removes the composed correction but does not imply that an unoptimized implementation skipped computing its admitted residuals.

\paragraph{Selector substitutions.} For each reverse step separately, let
\begin{equation}
 K_t=\operatorname{round}\!\left(\frac1{|\mathcal V|}
       \sum_{s\in\mathcal V}|E_t^{(s)}|\right)
\end{equation}
be the learned selector's mean directed edge count on the held-out validation panel. Both heuristic selectors rank the complete set of directed non-self candidates \emph{globally} and admit the top $K_t$; they do not select $K_t$ partners for every robot. Geometric selection ranks by physical start-position distance. Collision-risk selection ranks by minimum time-aligned separation along the current step's estimated clean unary trajectories. Both use the learned composer on the resulting support. Separate $K_t$ values preserve the learned mean budget profile over reverse time; geometric rankings can remain fixed while $K_t$ changes.

Calibration uses 32 Weave validation scenes and 26 Shelf validation scenes, disjoint from the evaluation scenes. The supporting data contain all scalar scores and per-step $K_t$ values. An evaluation outcome does not change the calibration or frozen model. Matching mean counts does not force per-scene graph equality and does not imply that the heuristic has the same repair workload.

On the final manuscript panel, raw summation increases Weave conflicts from 32.133 to 94.200 and Shelf conflicts from 24.800 to 61.960. Dense composition also illustrates an efficiency--success trade-off: dense Shelf solves 24/25 instances versus 22/25 for sparse Akule, while sparse selection avoids 80.0\% of directed residual evaluations. The benefits of sparsity should therefore be assessed jointly with success and downstream repair, rather than from residual savings alone. The calibrated Shelf unary-like proposal has higher complete latency than full Akule under this panel's reported results.

%% file: supplement/tables/ablations.tex
\begin{table}[t]
    \centering

\caption{\textbf{Ablation of selection and composition} on Canonical Weave ($N{=}28$, 30 instances) and Shelf ($N{=}18$, 25 instances). Conf.\ is mean pre-repair collision pair-times; Repair and Plan are repair and complete-planner latency (Plan over all runs); Quality is path length / acceleration over successful plans (Weave: physical RMS; Shelf: per-step second difference); $R_{\rm ft}$ save is the fraction of pairwise corrections avoided relative to dense Akule. Source: the current manuscript's final ablation table.}

    \label{tab:ablations}
    \akStyle
    \akFit{\begin{tabular}{l<{\hspace{\akSlack}} c r r r c r}
        \akH{Method} &
        \akH{Success} &
        \akH{Conf.} &
        \akH{Repair (s)} &
        \akH{Plan (s)} &
        \akH{Quality (P/A)} &
        \akH{$R_{\rm ft}$ save} \\
        \akAccentLine

        \multicolumn{7}{l}{\textit{Canonical Weave, $N=28$}} \\
        \akLine
        \akOurs \AkuleInlineLogo\akM{Akule}
            & 30/30 & 32.133 & 22.901 & 29.944
            & 1.949/2.813 & 29.369\% \\ \akLine
        Akule (dense)
            & 30/30 & 32.797 & 25.163 & 43.571
            & 1.950/2.824 & 0\% \\ \akLine
        Raw/calibrated sum ($\alpha=1$)
            & 30/30 & 94.200 & 43.573 & 50.756
            & 1.935/2.602 & 29.740\% \\ \akLine
        Geometric selector
            & 30/30 & 59.167 & 32.766 & 39.929
            & 1.940/2.719 & 29.381\% \\ \akLine
        Collision-risk selector
            & 30/30 & 62.967 & 34.948 & 40.113
            & 1.949/2.794 & 29.381\% \\

        \akGroupLine
        \multicolumn{7}{l}{\textit{Shelf, $N=18$}} \\
        \akLine
        \akOurs \AkuleInlineLogo\akM{Akule} & 22/25 & 24.800 & 73.632 & 76.923 & 1.487 / 0.013 & 80.0\% \\
        \akLine
Akule (dense) & 24/25 & 19.200 & 71.051 & 81.495 & 1.512 / 0.014 & 0.0\% \\
\akLine
Raw sum ($\alpha=1$) & 21/25 & 61.960 & 83.179 & 86.835 & 1.556 / 0.016 & 81.8\% \\
\akLine
Calibrated sum ($\alpha=0$) & 20/25 & 152.800 & 103.958 & 107.744 & 1.517 / 0.018 & 79.6\% \\
\akLine
Geometric selector & 21/25 & 68.160 & 88.469 & 92.077 & 1.505 / 0.015 & 79.8\% \\
\akLine
Collision-risk selector & 21/25 & 43.240 & 79.909 & 82.403 & 1.504 / 0.013 & 79.8\% \\

        \akAccentLine
    \end{tabular}}
\end{table}

%% file: supplement/sections/analysis.tex
\section{Interaction and Computational Analysis}
\label{sec:analysis}
\subsection{Dynamic support, additions, and removals}
At a reverse step the mean selected directed degree is $|E_t|/N$, and the fraction of pairwise evaluations avoided is $1-|E_t|/[N(N-1)]$. Across a full rollout we sum actual admitted evaluations and divide by $TN(N-1)$. For maps, the dense comparison is $N(N-1)$ because interaction occurs once. Edge additions and removals between consecutive reverse steps are normalized by the possible directed support $N(N-1)$; the symmetric difference divided by the union supplies a complementary turnover statistic.

\begin{figure*}[t]
\centering\includegraphics[width=\textwidth]{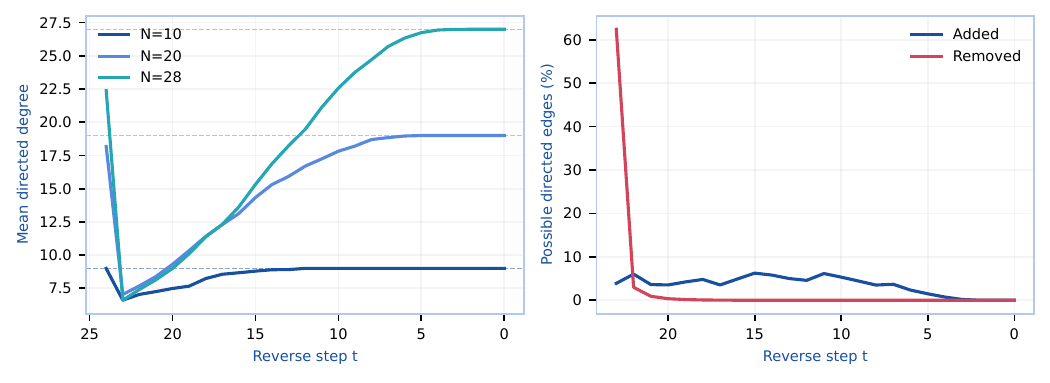}
\caption{Frozen Weave interaction profiles over reverse steps $24\to0$. Left: mean selected directed degree; dashed lines show dense degree $N-1$. Right: mean added and removed fractions of possible directed edges at $N=28$. Each curve uses the full corresponding panel (512/256/512 scenes). Source: saved interaction-degree and edge-transition exports.}
\label{fig:interaction}
\end{figure*}

Figure~\ref{fig:interaction} shows the frozen Weave profiles. Support is not monotonically pruned. At $N=28$, mean degree falls from 22.388 at $t=24$ to 6.596 at $t=23$, then grows to the full directed degree 27 at $t=1,0$. The first transition adds 3.90\% and removes 62.39\% of possible directed edges; 76.35\% of the union changes. Every profiled Weave trajectory, at all three populations, has at least one degree increase during denoising. Thus the savings arise from selectively reduced support at intermediate steps, not from a progressively vanishing graph. Highways and Conveyor have a single observed interaction step, so additions/removals over reverse time are undefined for those proposal interfaces.

\subsection{Signed coefficient usage}
Table~\ref{tab:signed} distinguishes negative spline coefficients from negative evaluated temporal weights. These differ because a support's weight interpolates several controls. On Weave $N=28$, 42.96\% of admitted coefficients and 44.46\% of evaluated weights are negative. The corresponding median negative magnitudes are 0.276 and 0.296. The map interface also uses negative values. Such statistics establish that the trained model uses sign reversal; they do not establish that sign freedom is necessary. That causal claim would require a controlled retrained nonnegative-composer comparison, which is not supplied by an inference-time absolute-value or clipping intervention.
\input{supplement/tables/signed}

\subsection{Sampler and interaction computation}
Table~\ref{tab:timing} reports exclusive CUDA-event intervals from frozen-proposal replay. The profiling contract uses serialized GPU work and checks proposal agreement rather than selecting replays for favorable timing. The Weave component intervals include its 25-step process; map unary intervals include candidate generation and guidance, while interaction is terminal. Host overhead and downstream search are outside these intervals.
\input{supplement/tables/timing}

Akule's map sampler batches independent candidates, while reference pipelines may generate candidates sequentially and MMD additionally performs soft coordination. A sampler-cost difference therefore cannot be assigned solely to $U/R/G$, and saved CUDA measurements on different GPUs are not matched wall-clock measurements. The replay environment is RTX 5090 for the Akule component table; original baseline/search timings retain their own hardware provenance. We do not add event intervals to older repair timers to synthesize a new complete latency.

\subsection{Downstream search workload}
Table~\ref{tab:repair} reports actual saved hard-repair timers, completed constraint-tree expansions, and generated hard constraints. At Weave $N=28$, Akule has 24.27 mean expansions, compared with 49.74 for MPD and 41.19 for MMD-xECBS. The two children of a normal conflict expansion add two hard constraints, explaining the relationship between these columns; generated constraint points are a separate quantity. Partial or interrupted expansions can break exact proportionality.

The map findings are less favorable for downstream burden. On Highways $N=20$, Akule has 11.3 mean CT expansions versus MMD's 4.1. On the six Conveyor $N=20$ tasks with complete MMD roots, these means are 53.5 and 33.67, respectively. These observations accompany MMD's lower root conflict counts and should remain visible alongside Akule's batching and proposal-generation advantages. For missing roots, we report coverage rather than inventing a comparable search input. Reduction against an independent unary proposal, sampler cost, and downstream search cost are three distinct observations.
\input{supplement/tables/repair}

%% file: supplement/tables/signed.tex
\begin{table}[!htbp]
\centering\caption{Signed-composer statistics on admitted edges. Fractions count negative coefficients/weights; medians summarize their negative magnitudes.}
\label{tab:signed}\akStyle
\akFit{\begin{tabular}{l r r r r r}
\akH{Environment} & \akH{$N$} & \akH{$\beta<0$} & \akH{$w<0$} & \akH{Med. $|\beta^-|$} & \akH{Med. $|w^-|$}\\\akAccentLine
Weave & 10 & 41.57\% & 45.51\% & 0.291 & 0.305\\\akLine
Weave & 20 & 43.05\% & 45.06\% & 0.281 & 0.299\\\akLine
Weave & 28 & 42.96\% & 44.46\% & 0.276 & 0.296\\\akLine
Highways & 20 & 30.88\% & 27.20\% & 0.144 & 0.133\\\akLine
Conveyor & 20 & 41.38\% & 34.03\% & 0.139 & 0.124\\\akLine
\akAccentLine
\end{tabular}}
\end{table}

%% file: supplement/tables/timing.tex
\begin{table}[!htbp]
\centering\caption{Akule CUDA-event component times (ms) on RTX 5090. Weave sums 25 steps; maps use one correction after candidate generation. Savings are relative to dense $R$ evaluation.}
\label{tab:timing}\akStyle
\akFit{\begin{tabular}{l r r r r r r r}
\akH{Environment} & \akH{$N$} & \akH{Scenes} & \akH{Unary} & \akH{U} & \akH{R} & \akH{G} & \akH{$R$ save}\\\akAccentLine
Weave & 10 & 512 & 64.20 & 38.48 & 12.80 & 27.73 & 4.85\%\\\akLine
Weave & 20 & 256 & 65.43 & 39.34 & 20.36 & 27.32 & 19.20\%\\\akLine
Weave & 28 & 512 & 64.18 & 39.08 & 31.09 & 36.73 & 29.37\%\\\akLine
Highways & 20 & 10 & 806.03 & 2.06 & 1.38 & 1.82 & 59.39\%\\\akLine
Conveyor & 20 & 10 & 919.07 & 2.09 & 1.78 & 2.05 & 38.71\%\\\akLine
\akAccentLine
\end{tabular}}
\end{table}

%% file: supplement/tables/repair.tex
\begin{table}[!htbp]
\centering\caption{Mean downstream repair workload. Weave uses full panels, Highways ten tasks, and Conveyor six matched complete-root tasks. CT counts tree expansions; constraints count generated branch constraints. Timers retain their recorded hardware.}
\label{tab:repair}\akStyle
\akFit{\begin{tabular}{l r r r r r}
\akH{Env. / $N$} & \akH{Method} & \akH{Scenes} & \akH{CT} & \akH{Constraints} & \akH{Repair (s)}\\\akAccentLine
Weave 10 & MPD & 512 & 5.00 & 10.00 & 3.900\\\akLine
Weave 10 & MMD-xECBS & 512 & 4.63 & 9.25 & 3.502\\\akLine
\akOurs Weave 10 & \akM{Akule} & 512 & 3.94 & 7.88 & 2.977\\\akLine
Weave 20 & MPD & 256 & 21.44 & 42.88 & 26.583\\\akLine
Weave 20 & MMD-xECBS & 256 & 14.16 & 28.31 & 15.588\\\akLine
\akOurs Weave 20 & \akM{Akule} & 256 & 7.20 & 14.40 & 6.936\\\akLine
Weave 28 & MPD & 512 & 49.74 & 99.47 & 77.496\\\akLine
Weave 28 & MMD-xECBS & 512 & 41.19 & 82.38 & 57.567\\\akLine
\akOurs Weave 28 & \akM{Akule} & 512 & 24.27 & 48.55 & 29.738\\\akLine
Highways 20 & MMD-xECBS & 10 & 4.10 & 8.20 & 3.771\\\akLine
\akOurs Highways 20 & \akM{Akule} & 10 & 11.30 & 22.60 & 14.115\\\akLine
Conveyor 20 & MMD-xECBS & 6 & 33.67 & 66.50 & 24.693\\\akLine
\akOurs Conveyor 20 & \akM{Akule} & 6 & 53.50 & 104.67 & 42.192\\\akLine
\akAccentLine
\end{tabular}}
\end{table}

%% file: supplement/sections/qualitative.tex
\input{supplement/sections/qual_weave}
\input{supplement/sections/qual_highways}
\input{supplement/sections/qual_conveyor}
\input{supplement/sections/qual_basic}
\input{supplement/sections/qual_dense}
\input{supplement/sections/qual_shelf}
\input{supplement/sections/qual_room}

%% file: supplement/sections/qual_weave.tex
\begin{figure*}[!tp]
\begin{minipage}{\textwidth}
\input{supplement/sections/qualitative_intro}
\subsection{Canonical Weave}
The canonical Weave examples show opposing flows in the fixed workspace at all three evaluated populations. The MPD panels show independent proposals before repair; MMD and Akule show saved downstream outputs. Figure~\ref{fig:qual_weave} presents the corresponding comparisons.
\end{minipage}
\medskip
\centering\akStyle
\def\akPanel{\dimexpr.215\textwidth\relax}
\begin{tabular}{@{}c@{\hspace{3pt}}c@{\hspace{3pt}}c@{}}
\multicolumn{3}{c}{\akH{Canonical Weave, $N=10$}}\\[1pt]
\akQualOther{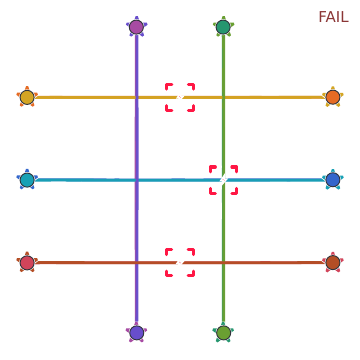} & \akQualOther{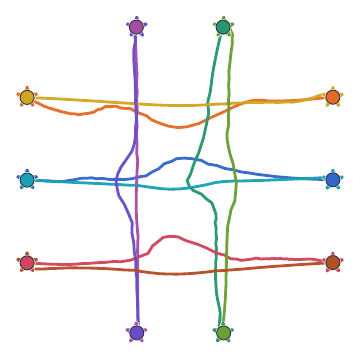} & \akQualOurs{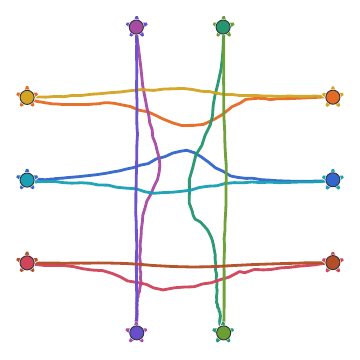}\\[1pt]
MPD~\akQualFail & MMD-xECBS & \AkuleInlineLogo\akM{Akule}\\
\end{tabular}\par\smallskip
\begin{tabular}{@{}c@{\hspace{3pt}}c@{\hspace{3pt}}c@{}}
\multicolumn{3}{c}{\akH{Canonical Weave, $N=20$}}\\[1pt]
\akQualOther{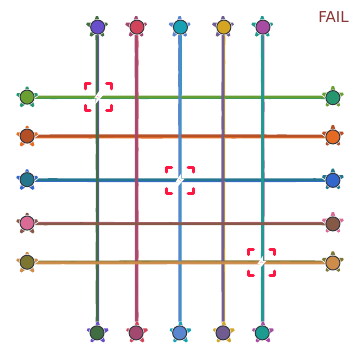} & \akQualOther{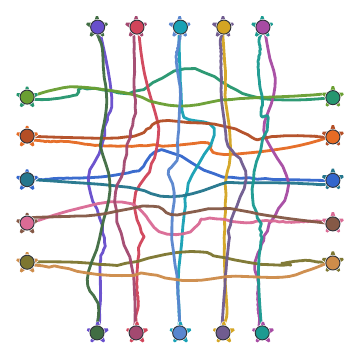} & \akQualOurs{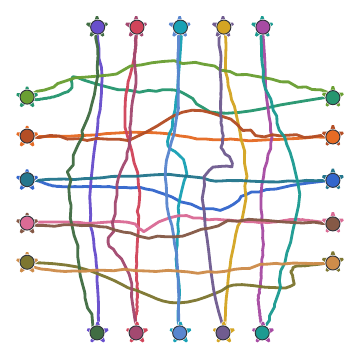}\\[1pt]
MPD~\akQualFail & MMD-xECBS & \AkuleInlineLogo\akM{Akule}\\
\end{tabular}\par\smallskip
\begin{tabular}{@{}c@{\hspace{3pt}}c@{\hspace{3pt}}c@{}}
\multicolumn{3}{c}{\akH{Canonical Weave, $N=28$}}\\[1pt]
\akQualOther{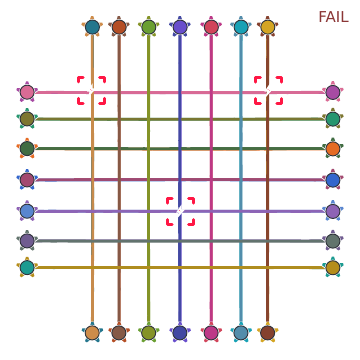} & \akQualOther{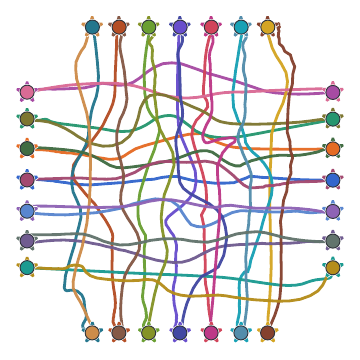} & \akQualOurs{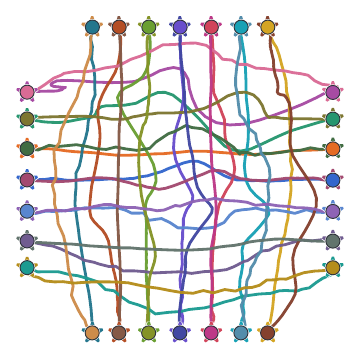}\\[1pt]
MPD~\akQualFail & MMD-xECBS & \AkuleInlineLogo\akM{Akule}\\
\end{tabular}\par\smallskip
\caption{Canonical Weave at $N=10, 20, 28$. Independent MPD proposals are shown before repair. Same task across methods at each population. Scene IDs, in population order: 510001, 510000, 105. The shared plotting conventions are defined in Section~\ref{sec:qualitative}.}
\label{fig:qual_weave}
\end{figure*}

%% file: supplement/sections/qualitative_intro.tex
\section{Qualitative Results}
\label{sec:qualitative}
The following comparisons cover all 39 evaluated population conditions in the seven environments. Each comparison uses the same saved task across methods, with consistent robot colors and spatial limits. Discs mark starts, stars mark goals, and wheat-colored regions show obstacles. A blue box highlights Akule. Native failures are labeled \textcolor{red!80!black}{[FAIL]}; a drawable trajectory alone does not establish complete-planner success.

Scarlet brackets and an impact symbol mark up to three spatially separated conflicts in failed baseline panels. Callouts use simultaneous saved supports and the reference renderer's pair/obstacle predicates; geometric crossings at different times are not marked. Panels with no eligible conflict receive no callout. These annotations illustrate individual interactions; aggregate results are reported in Section~\ref{sec:quantitative}.

The SMD examples use the 13th instance of each population's 25-scene panel. Reused largest-population high-fidelity panels retain their original display treatment; other panels are rendered directly from saved supports. Scene IDs appear in each caption, and the supporting manifest maps every panel to its input array or retained asset.

%% file: supplement/sections/qual_highways.tex
\begin{figure*}[!tp]
\begin{minipage}{\textwidth}
\subsection{Highways}
Highways illustrates the interaction of several robot streams around the repeated rounded-box obstacles. Each population uses one shared start/goal task across MPD, MMD, and Akule. Figure~\ref{fig:qual_highways} presents the corresponding comparisons.
\end{minipage}
\medskip
\centering\akStyle
\def\akPanel{\dimexpr.149\textwidth\relax}
\begin{tabular}{@{}c@{\hspace{3pt}}c@{\hspace{3pt}}c@{\hspace{10pt}}c@{\hspace{3pt}}c@{\hspace{3pt}}c@{}}
\multicolumn{3}{c}{\akH{Highways, $N=3$}} & \multicolumn{3}{c}{\akH{Highways, $N=6$}}\\[1pt]
\akQualOther{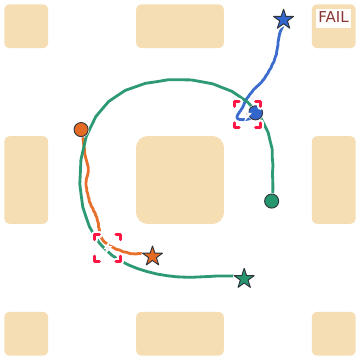} & \akQualOther{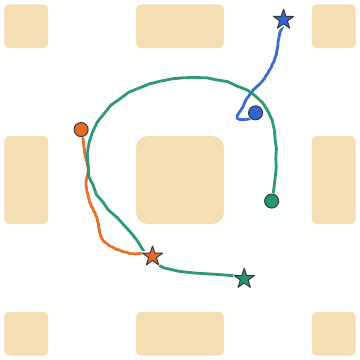} & \akQualOurs{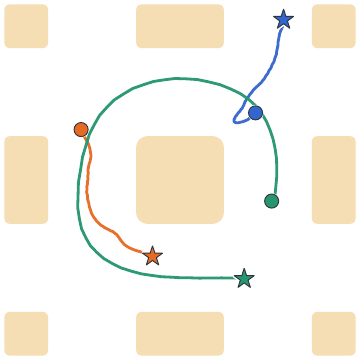} & \akQualOther{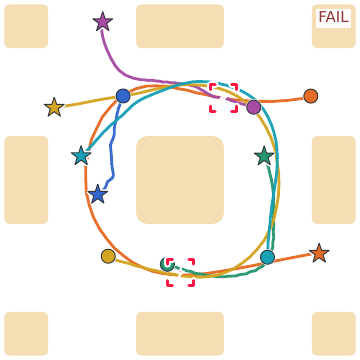} & \akQualOther{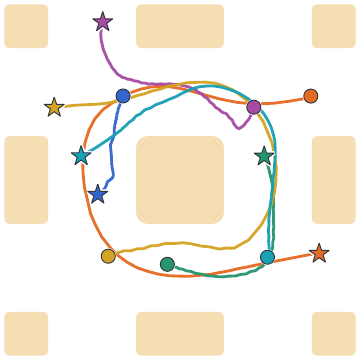} & \akQualOurs{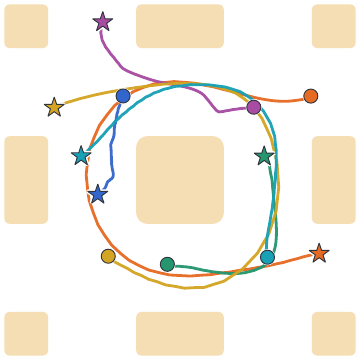}\\[1pt]
MPD~\akQualFail & MMD & \AkuleInlineLogo\akM{Akule} & MPD~\akQualFail & MMD & \AkuleInlineLogo\akM{Akule}\\
\end{tabular}\par\smallskip
\begin{tabular}{@{}c@{\hspace{3pt}}c@{\hspace{3pt}}c@{\hspace{10pt}}c@{\hspace{3pt}}c@{\hspace{3pt}}c@{}}
\multicolumn{3}{c}{\akH{Highways, $N=9$}} & \multicolumn{3}{c}{\akH{Highways, $N=12$}}\\[1pt]
\akQualOther{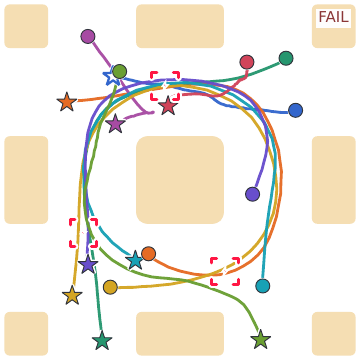} & \akQualOther{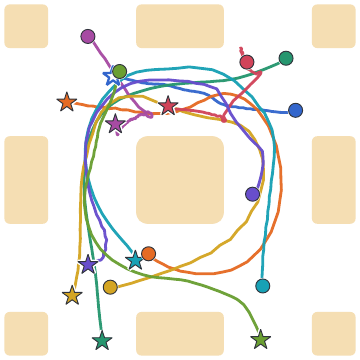} & \akQualOurs{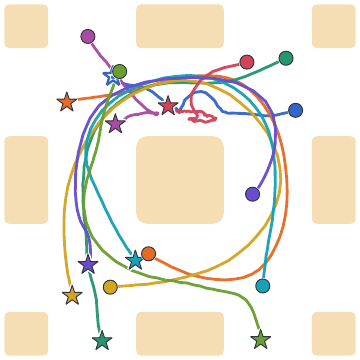} & \akQualOther{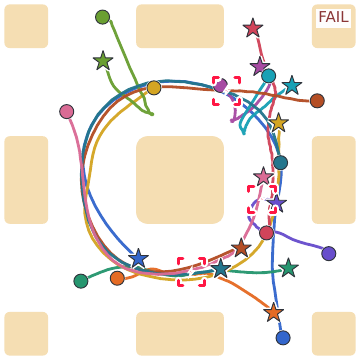} & \akQualOther{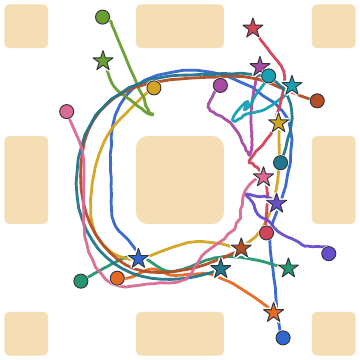} & \akQualOurs{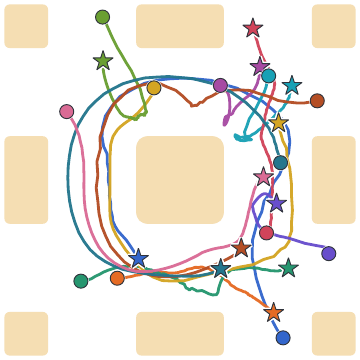}\\[1pt]
MPD~\akQualFail & MMD & \AkuleInlineLogo\akM{Akule} & MPD~\akQualFail & MMD & \AkuleInlineLogo\akM{Akule}\\
\end{tabular}\par\smallskip
\begin{tabular}{@{}c@{\hspace{3pt}}c@{\hspace{3pt}}c@{\hspace{10pt}}c@{\hspace{3pt}}c@{\hspace{3pt}}c@{}}
\multicolumn{3}{c}{\akH{Highways, $N=15$}} & \multicolumn{3}{c}{\akH{Highways, $N=20$}}\\[1pt]
\akQualOther{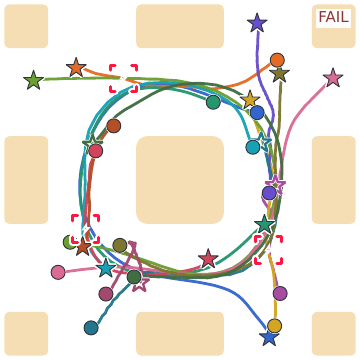} & \akQualOther{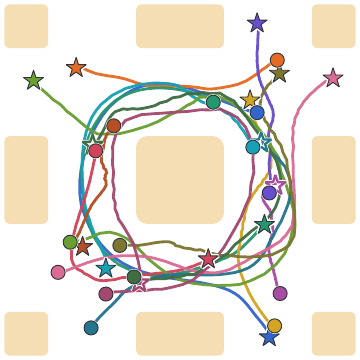} & \akQualOurs{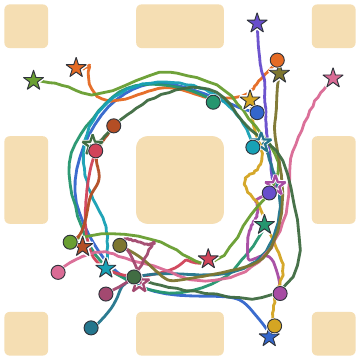} & \akQualOther{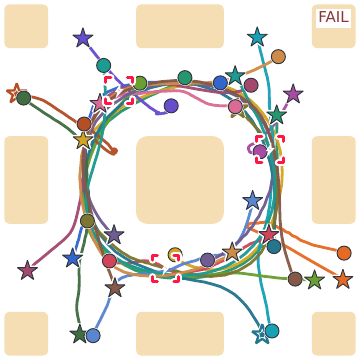} & \akQualOther{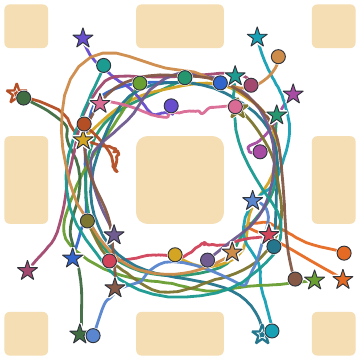} & \akQualOurs{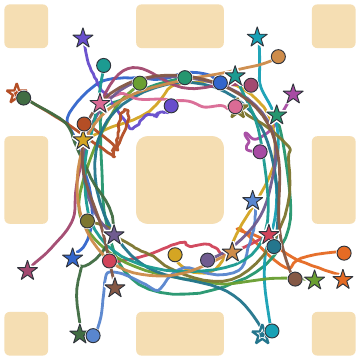}\\[1pt]
MPD~\akQualFail & MMD & \AkuleInlineLogo\akM{Akule} & MPD~\akQualFail & MMD & \AkuleInlineLogo\akM{Akule}\\
\end{tabular}\par\smallskip
\caption{Highways at $N=3, 6, 9, 12, 15, 20$. Same task across methods at each population. Scene IDs, in population order: highways\_n3\_trial02, highways\_n6\_trial04, highways\_n9\_trial00, highways\_n12\_trial05, highways\_n15\_trial03, highways\_n20\_trial06. The shared plotting conventions are defined in Section~\ref{sec:qualitative}.}
\label{fig:qual_highways}
\end{figure*}

%% file: supplement/sections/qual_conveyor.tex
\begin{figure*}[!tp]
\begin{minipage}{\textwidth}
\subsection{Conveyor}
Conveyor compares the saved trajectories around its directional corridor geometry, with the same task and robot colors across methods. Figure~\ref{fig:qual_conveyor} presents the corresponding comparisons.
\end{minipage}
\medskip
\centering\akStyle
\def\akPanel{\dimexpr.149\textwidth\relax}
\begin{tabular}{@{}c@{\hspace{3pt}}c@{\hspace{3pt}}c@{\hspace{10pt}}c@{\hspace{3pt}}c@{\hspace{3pt}}c@{}}
\multicolumn{3}{c}{\akH{Conveyor, $N=3$}} & \multicolumn{3}{c}{\akH{Conveyor, $N=6$}}\\[1pt]
\akQualOther{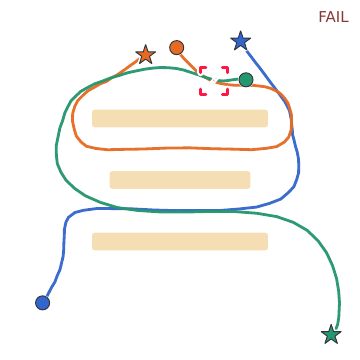} & \akQualOther{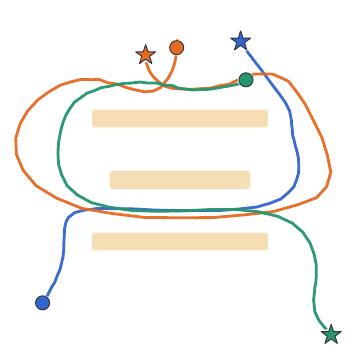} & \akQualOurs{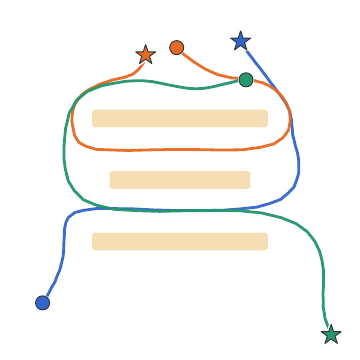} & \akQualOther{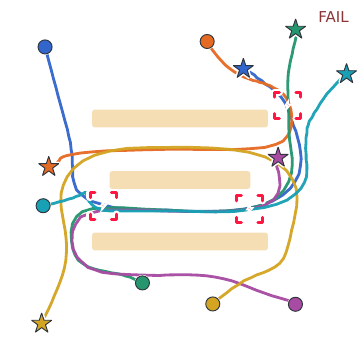} & \akQualOther{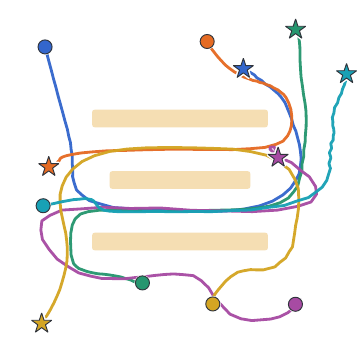} & \akQualOurs{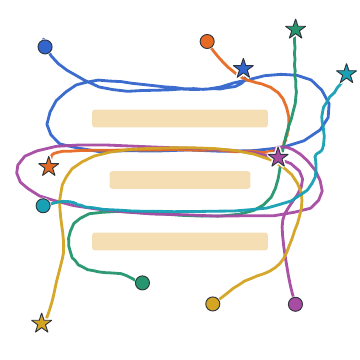}\\[1pt]
MPD~\akQualFail & MMD & \AkuleInlineLogo\akM{Akule} & MPD~\akQualFail & MMD & \AkuleInlineLogo\akM{Akule}\\
\end{tabular}\par\smallskip
\begin{tabular}{@{}c@{\hspace{3pt}}c@{\hspace{3pt}}c@{\hspace{10pt}}c@{\hspace{3pt}}c@{\hspace{3pt}}c@{}}
\multicolumn{3}{c}{\akH{Conveyor, $N=9$}} & \multicolumn{3}{c}{\akH{Conveyor, $N=12$}}\\[1pt]
\akQualOther{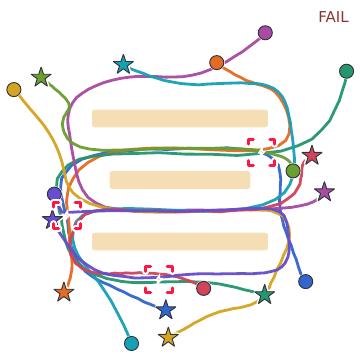} & \akQualOther{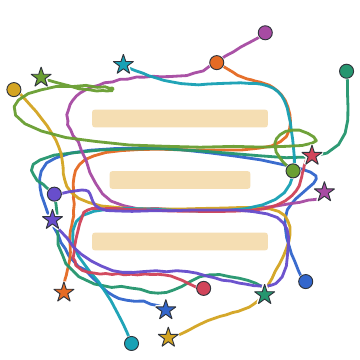} & \akQualOurs{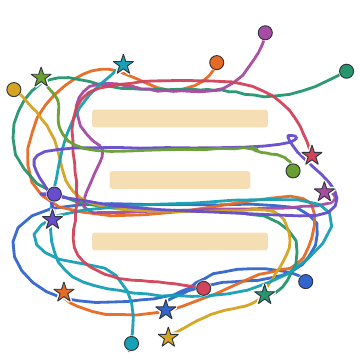} & \akQualOther{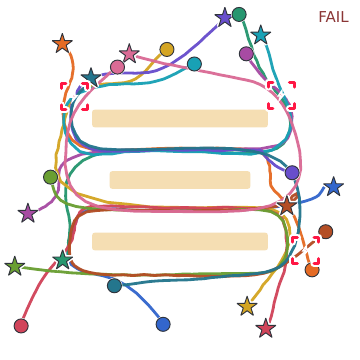} & \akQualOther{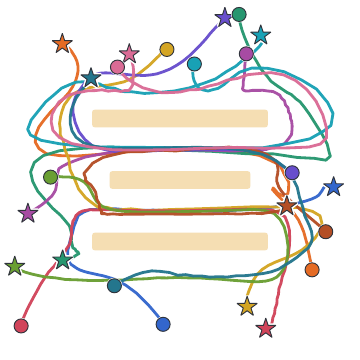} & \akQualOurs{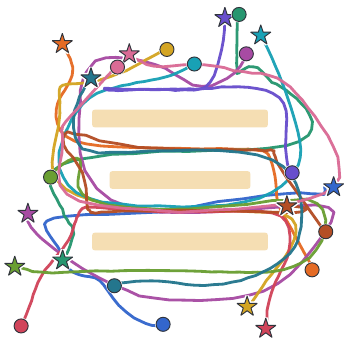}\\[1pt]
MPD~\akQualFail & MMD & \AkuleInlineLogo\akM{Akule} & MPD~\akQualFail & MMD & \AkuleInlineLogo\akM{Akule}\\
\end{tabular}\par\smallskip
\begin{tabular}{@{}c@{\hspace{3pt}}c@{\hspace{3pt}}c@{\hspace{10pt}}c@{\hspace{3pt}}c@{\hspace{3pt}}c@{}}
\multicolumn{3}{c}{\akH{Conveyor, $N=15$}} & \multicolumn{3}{c}{\akH{Conveyor, $N=20$}}\\[1pt]
\akQualOther{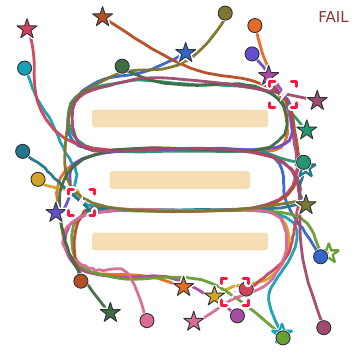} & \akQualOther{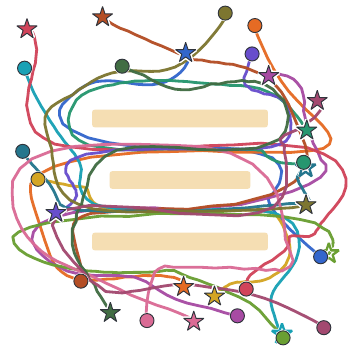} & \akQualOurs{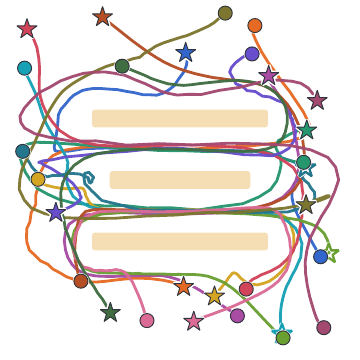} & \akQualOther{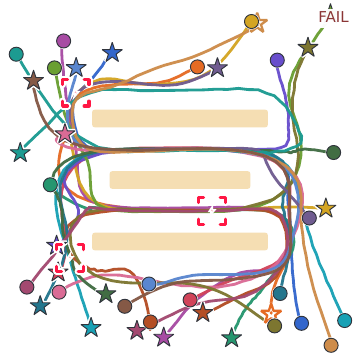} & \akQualOther{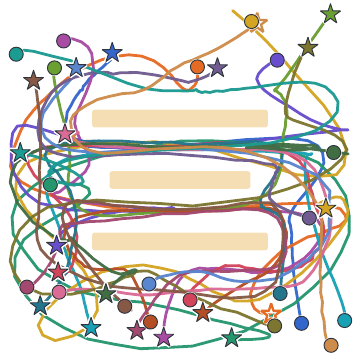} & \akQualOurs{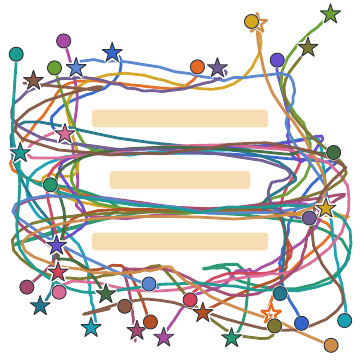}\\[1pt]
MPD~\akQualFail & MMD & \AkuleInlineLogo\akM{Akule} & MPD~\akQualFail & MMD & \AkuleInlineLogo\akM{Akule}\\
\end{tabular}\par\smallskip
\caption{Conveyor at $N=3, 6, 9, 12, 15, 20$. Same task across methods at each population. Scene IDs, in population order: conveyor\_n3\_trial00, conveyor\_n6\_trial02, conveyor\_n9\_trial06, conveyor\_n12\_trial07, conveyor\_n15\_trial05, conveyor\_n20\_trial05. The shared plotting conventions are defined in Section~\ref{sec:qualitative}.}
\label{fig:qual_conveyor}
\end{figure*}

%% file: supplement/sections/qual_basic.tex
\begin{figure*}[!tp]
\begin{minipage}{\textwidth}
\subsection{Basic}
Basic illustrates coordination around scattered circular obstacles across the six evaluated populations. Figure~\ref{fig:qual_basic} presents the corresponding comparisons.
\end{minipage}
\medskip
\centering\akStyle
\def\akPanel{\dimexpr.215\textwidth\relax}
\begin{tabular}{@{}l@{\hspace{6pt}}c@{\hspace{5pt}}c@{\hspace{5pt}}c@{}}
 & \makebox[\akPanel]{\akH{$N=3$}} & \makebox[\akPanel]{\akH{$N=6$}} & \makebox[\akPanel]{\akH{$N=9$}}\\[1pt]
\raisebox{\dimexpr.5\akPanel-2pt\relax}{SMD} & \akQualOther{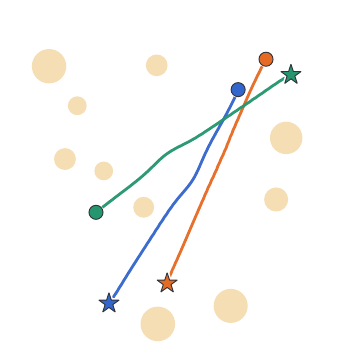} & \akQualOther{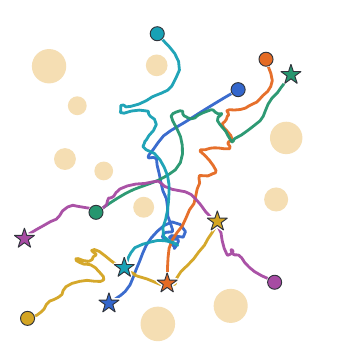} & \akQualOther{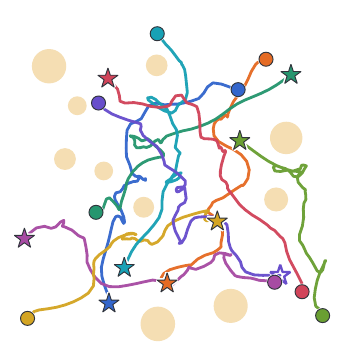}\\[3pt]
\raisebox{\dimexpr.5\akPanel-2pt\relax}{\AkuleInlineLogo\akM{Akule}} & \akQualOurs{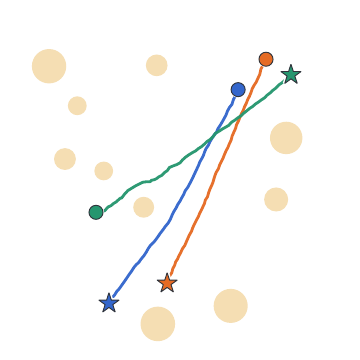} & \akQualOurs{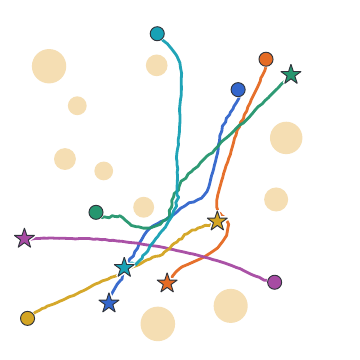} & \akQualOurs{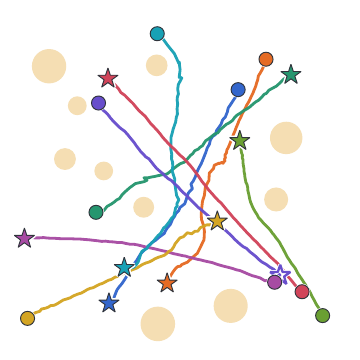}\\[3pt]
\end{tabular}\par\smallskip
\begin{tabular}{@{}l@{\hspace{6pt}}c@{\hspace{5pt}}c@{\hspace{5pt}}c@{}}
 & \makebox[\akPanel]{\akH{$N=12$}} & \makebox[\akPanel]{\akH{$N=15$}} & \makebox[\akPanel]{\akH{$N=18$}}\\[1pt]
\raisebox{\dimexpr.5\akPanel-2pt\relax}{SMD} & \akQualOther{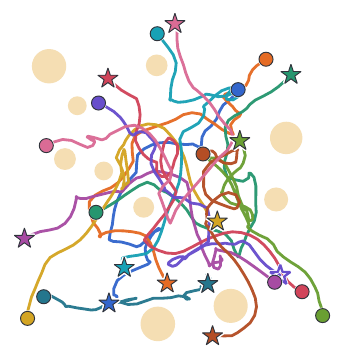} & \akQualOther{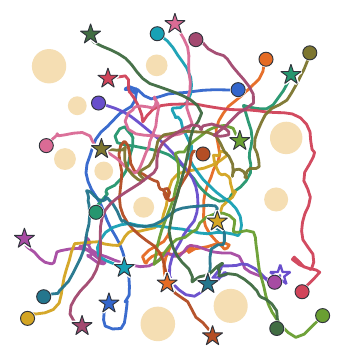} & \akQualOther{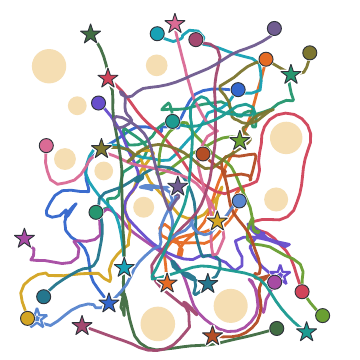}\\[3pt]
\raisebox{\dimexpr.5\akPanel-2pt\relax}{\AkuleInlineLogo\akM{Akule}} & \akQualOurs{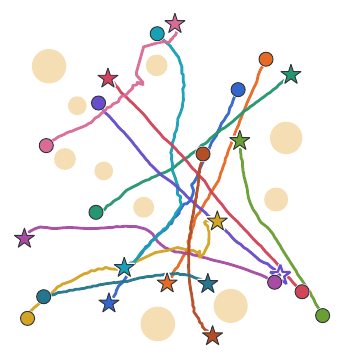} & \akQualOurs{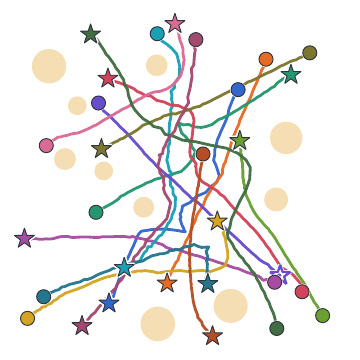} & \akQualOurs{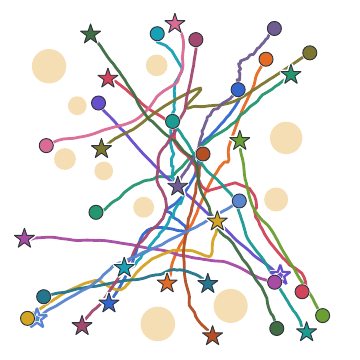}\\[3pt]
\end{tabular}\par\smallskip
\caption{Basic at $N=3, 6, 9, 12, 15, 18$. Same task across methods at each population. Scene IDs, in population order: 81003012, 81006012, 81009012, 81012012, 81015012, 81018012. The shared plotting conventions are defined in Section~\ref{sec:qualitative}.}
\label{fig:qual_basic}
\end{figure*}

%% file: supplement/sections/qual_dense.tex
\begin{figure*}[!tp]
\begin{minipage}{\textwidth}
\subsection{Dense}
Dense illustrates the more constrained obstacle field; native failure labels and selected callouts remain visible on unsuccessful baseline outputs. Figure~\ref{fig:qual_dense} presents the corresponding comparisons.
\end{minipage}
\medskip
\centering\akStyle
\def\akPanel{\dimexpr.215\textwidth\relax}
\begin{tabular}{@{}l@{\hspace{6pt}}c@{\hspace{5pt}}c@{\hspace{5pt}}c@{}}
 & \makebox[\akPanel]{\akH{$N=3$}} & \makebox[\akPanel]{\akH{$N=6$}} & \makebox[\akPanel]{\akH{$N=9$}}\\[1pt]
\raisebox{\dimexpr.5\akPanel-2pt\relax}{SMD} & \akQualOther{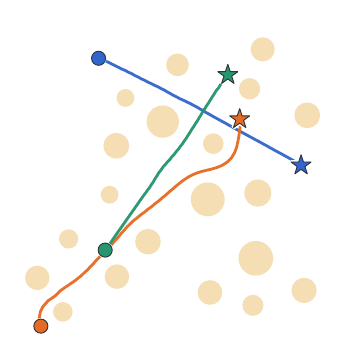} & \akQualOther{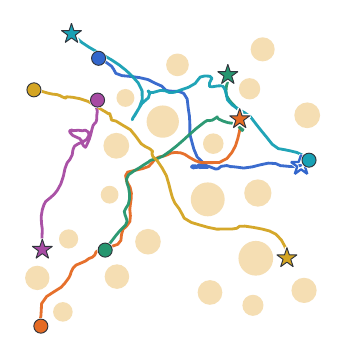} & \akQualOther{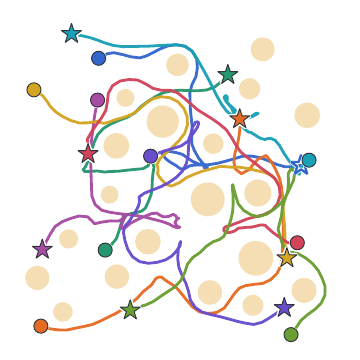}\\[3pt]
\raisebox{\dimexpr.5\akPanel-2pt\relax}{\AkuleInlineLogo\akM{Akule}} & \akQualOurs{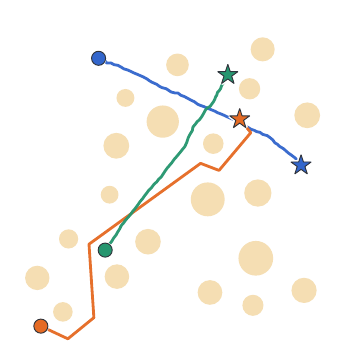} & \akQualOurs{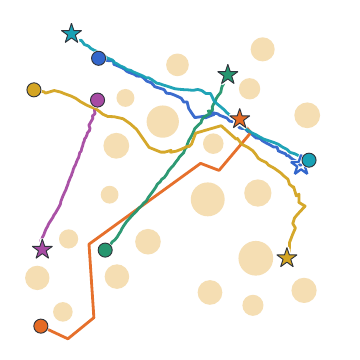} & \akQualOurs{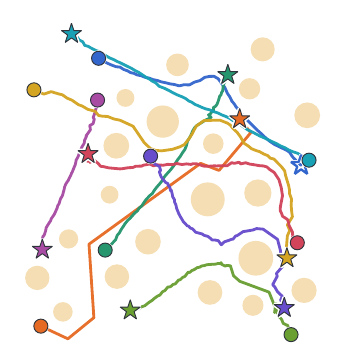}\\[3pt]
\end{tabular}\par\smallskip
\begin{tabular}{@{}l@{\hspace{6pt}}c@{\hspace{5pt}}c@{\hspace{5pt}}c@{}}
 & \makebox[\akPanel]{\akH{$N=12$}} & \makebox[\akPanel]{\akH{$N=15$}} & \makebox[\akPanel]{\akH{$N=18$}}\\[1pt]
\raisebox{\dimexpr.5\akPanel-2pt\relax}{SMD} & \akQualOther{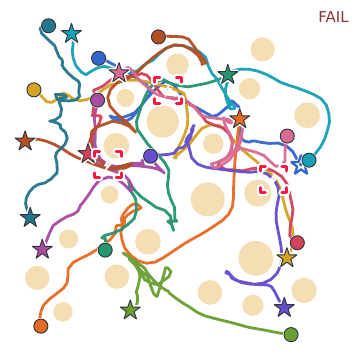} & \akQualOther{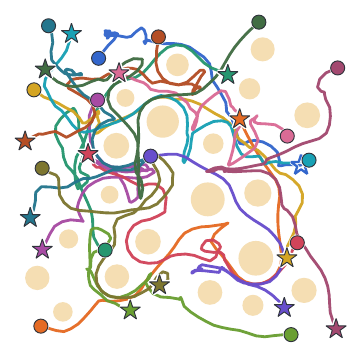} & \akQualOther{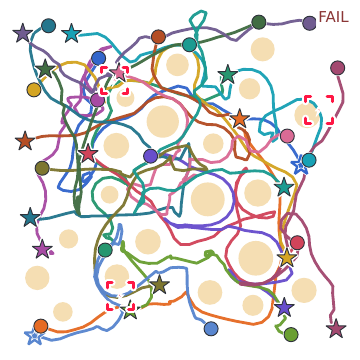}\\[3pt]
\raisebox{\dimexpr.5\akPanel-2pt\relax}{\AkuleInlineLogo\akM{Akule}} & \akQualOurs{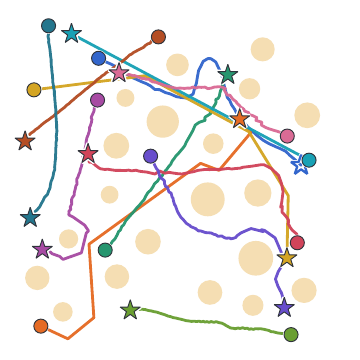} & \akQualOurs{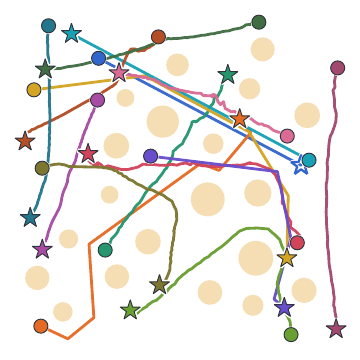} & \akQualOurs{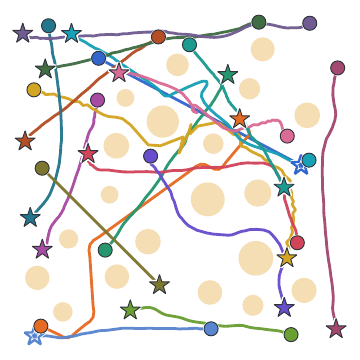}\\[3pt]
\end{tabular}\par\smallskip
\caption{Dense at $N=3, 6, 9, 12, 15, 18$. Same task across methods at each population. Scene IDs, in population order: 82003012, 82006012, 82009012, 82012012, 82015012, 82018012. The shared plotting conventions are defined in Section~\ref{sec:qualitative}.}
\label{fig:qual_dense}
\end{figure*}

%% file: supplement/sections/qual_shelf.tex
\begin{figure*}[!tp]
\begin{minipage}{\textwidth}
\subsection{Shelf}
Shelf illustrates how the recorded trajectories pass between the obstacle columns. The plots retain the evaluated circle-union geometry. Figure~\ref{fig:qual_shelf} presents the corresponding comparisons.
\end{minipage}
\medskip
\centering\akStyle
\def\akPanel{\dimexpr.215\textwidth\relax}
\begin{tabular}{@{}l@{\hspace{6pt}}c@{\hspace{5pt}}c@{\hspace{5pt}}c@{}}
 & \makebox[\akPanel]{\akH{$N=3$}} & \makebox[\akPanel]{\akH{$N=6$}} & \makebox[\akPanel]{\akH{$N=9$}}\\[1pt]
\raisebox{\dimexpr.5\akPanel-2pt\relax}{SMD} & \akQualOther{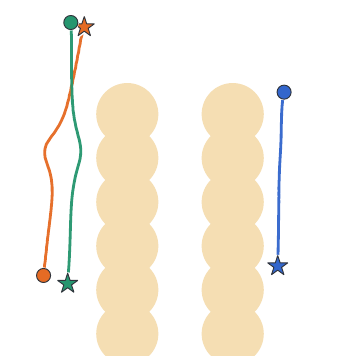} & \akQualOther{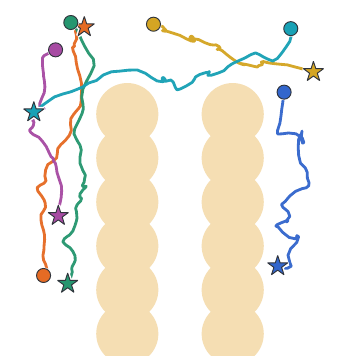} & \akQualOther{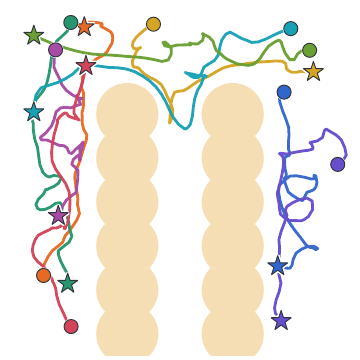}\\[3pt]
\raisebox{\dimexpr.5\akPanel-2pt\relax}{\AkuleInlineLogo\akM{Akule}} & \akQualOurs{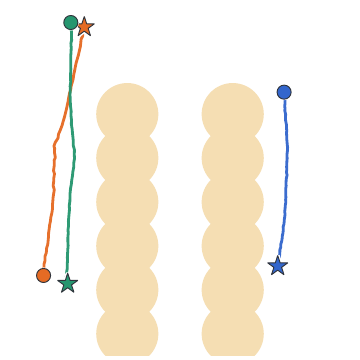} & \akQualOurs{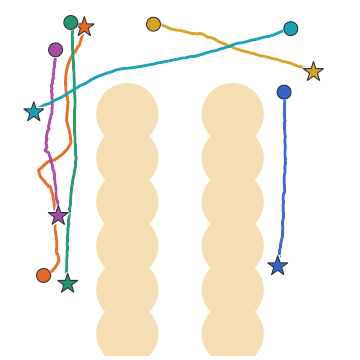} & \akQualOurs{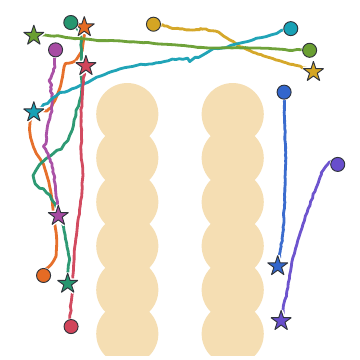}\\[3pt]
\end{tabular}\par\smallskip
\begin{tabular}{@{}l@{\hspace{6pt}}c@{\hspace{5pt}}c@{\hspace{5pt}}c@{}}
 & \makebox[\akPanel]{\akH{$N=12$}} & \makebox[\akPanel]{\akH{$N=15$}} & \makebox[\akPanel]{\akH{$N=18$}}\\[1pt]
\raisebox{\dimexpr.5\akPanel-2pt\relax}{SMD} & \akQualOther{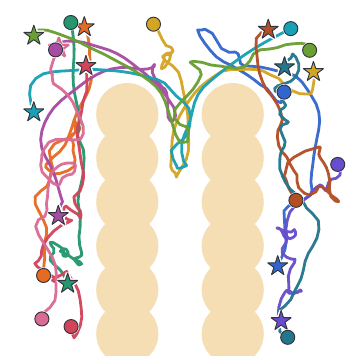} & \akQualOther{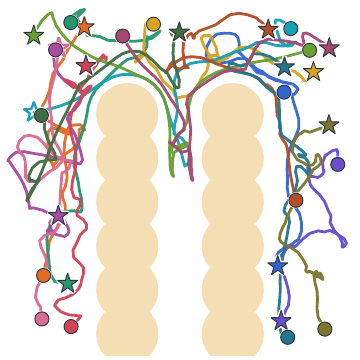} & \akQualOther{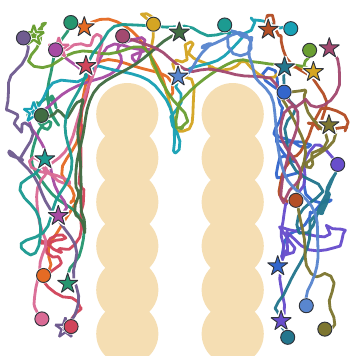}\\[3pt]
\raisebox{\dimexpr.5\akPanel-2pt\relax}{\AkuleInlineLogo\akM{Akule}} & \akQualOurs{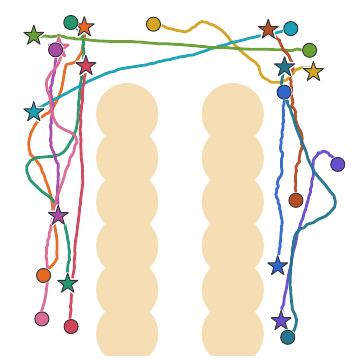} & \akQualOurs{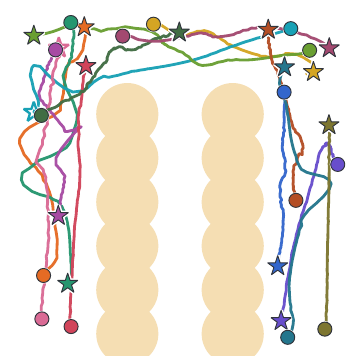} & \akQualOurs{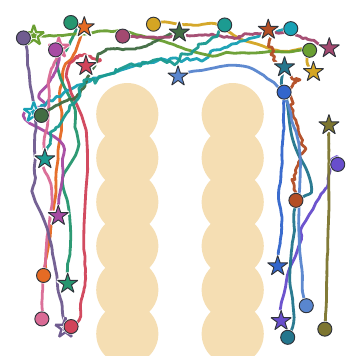}\\[3pt]
\end{tabular}\par\smallskip
\caption{Shelf at $N=3, 6, 9, 12, 15, 18$. Same task across methods at each population. Scene IDs, in population order: 83003012, 83006012, 83009012, 83012012, 83015012, 83018012. The shared plotting conventions are defined in Section~\ref{sec:qualitative}.}
\label{fig:qual_shelf}
\end{figure*}

%% file: supplement/sections/qual_room.tex
\begin{figure*}[!tp]
\begin{minipage}{\textwidth}
\subsection{Room}
Room illustrates the alternative passages around its circle-union partitions at each evaluated population. Figure~\ref{fig:qual_room} presents the corresponding comparisons.
\end{minipage}
\medskip
\centering\akStyle
\def\akPanel{\dimexpr.215\textwidth\relax}
\begin{tabular}{@{}l@{\hspace{6pt}}c@{\hspace{5pt}}c@{\hspace{5pt}}c@{}}
 & \makebox[\akPanel]{\akH{$N=3$}} & \makebox[\akPanel]{\akH{$N=6$}} & \makebox[\akPanel]{\akH{$N=9$}}\\[1pt]
\raisebox{\dimexpr.5\akPanel-2pt\relax}{SMD} & \akQualOther{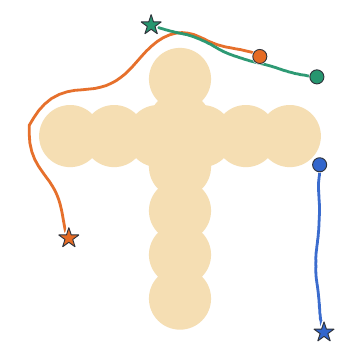} & \akQualOther{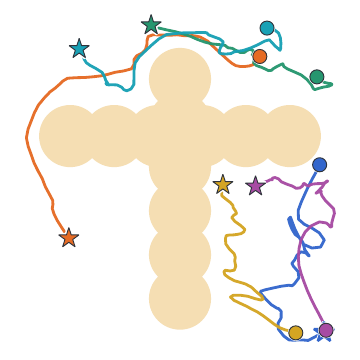} & \akQualOther{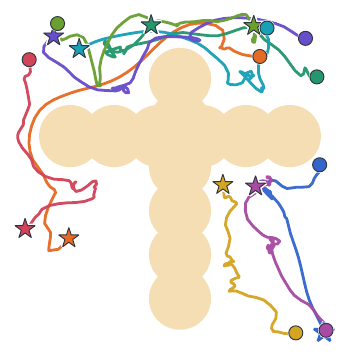}\\[3pt]
\raisebox{\dimexpr.5\akPanel-2pt\relax}{\AkuleInlineLogo\akM{Akule}} & \akQualOurs{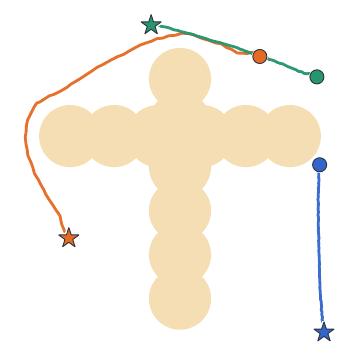} & \akQualOurs{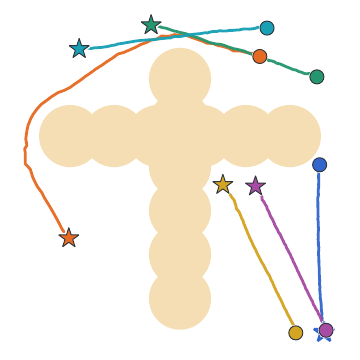} & \akQualOurs{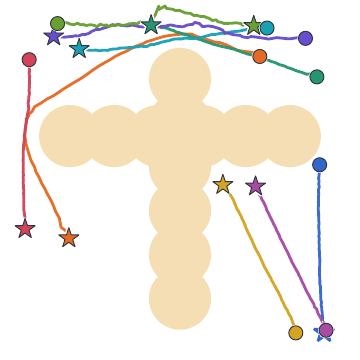}\\[3pt]
\end{tabular}\par\smallskip
\begin{tabular}{@{}l@{\hspace{6pt}}c@{\hspace{5pt}}c@{\hspace{5pt}}c@{}}
 & \makebox[\akPanel]{\akH{$N=12$}} & \makebox[\akPanel]{\akH{$N=15$}} & \makebox[\akPanel]{\akH{$N=18$}}\\[1pt]
\raisebox{\dimexpr.5\akPanel-2pt\relax}{SMD} & \akQualOther{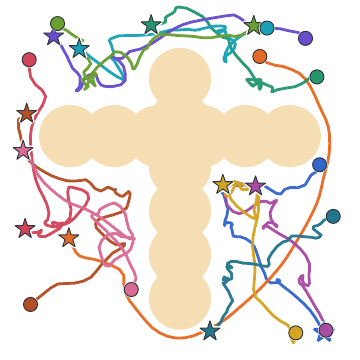} & \akQualOther{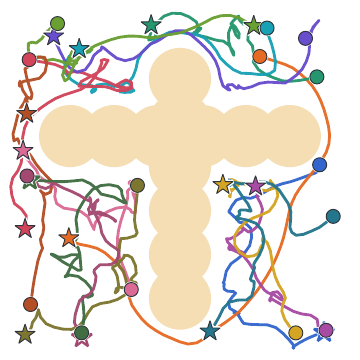} & \akQualOther{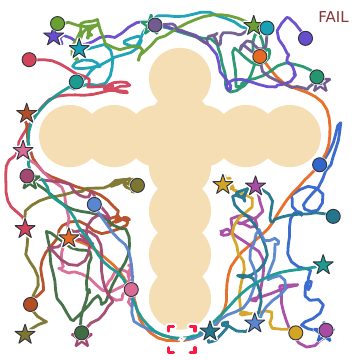}\\[3pt]
\raisebox{\dimexpr.5\akPanel-2pt\relax}{\AkuleInlineLogo\akM{Akule}} & \akQualOurs{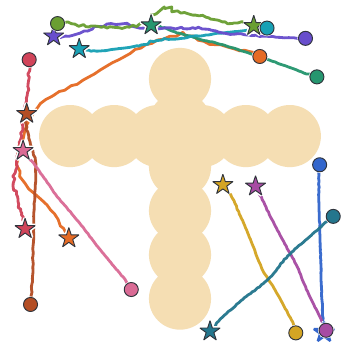} & \akQualOurs{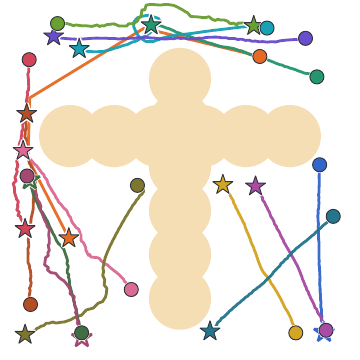} & \akQualOurs{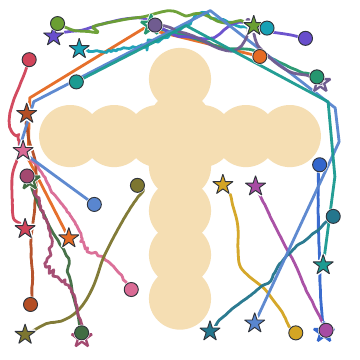}\\[3pt]
\end{tabular}\par\smallskip
\caption{Room at $N=3, 6, 9, 12, 15, 18$. Same task across methods at each population. Scene IDs, in population order: 84003012, 84006012, 84009012, 84012012, 84015012, 84018012. The shared plotting conventions are defined in Section~\ref{sec:qualitative}.}
\label{fig:qual_room}
\end{figure*}

%% file: supplement/sections/additional.tex
\section{Additional Studies and Limitations}
\label{sec:additional}
\subsection{Safety-preserving smoothing}
Saved-path smoothing can improve the regularity of a completed trajectory without regenerating its proposal or repeating hard search. Table~\ref{tab:smoothing} reports the existing Weave study at $N=10,20,28$: a candidate is accepted only when it passes the study's validation gate, and otherwise the original repaired trajectory is retained. Native validated success is preserved, while mean path length, acceleration RMS, and jerk RMS decrease. Length is in metres; acceleration and jerk use the physical timestep convention of Section~\ref{sec:protocol}. These saved-path measurements are separate from online proposal and search timings. The table reproduces the recorded summary and original planning fields; post-smoothing execution/response-to-completion measurements are not reported.
\input{supplement/tables/smoothing}

\subsection{Alternative downstream planners}
Akule's proposal interface can be combined with different CBS-family backends. We evaluate this interface on a fixed small panel of 20 canonical Weave $N=28$ tasks, using the first 20 frozen final-test conditions and unchanged pretrained models. Table~\ref{tab:backends} distinguishes newly executed ECBS measurements from existing CBS/xECBS records on those condition indices. Each attempt has a 240 s native planning limit and a 255 s external watchdog. We do not train or tune the coordination model for the substituted backend. Success requires the same complete-team, endpoint, and sampled physical collision checks as the canonical benchmark.

The comparison tests compatibility and sensitivity to the repair backend. It is a small-panel study, not a replacement for the full Weave benchmark. Historical and newly collected timers retain their own execution provenance; the table does not establish a matched-hardware latency speedup between cohorts. Mean latency includes all 20 attempts and assigns 240 s to timeouts. Successful-run quality is reported only where a valid result exists.
\input{supplement/tables/backends}

Plain CBS times out on all 20 tasks for both proposal interfaces. Newly executed Akule--ECBS solves 20/20, with mean all-attempt latency 77.45 s and 28.75 CT expansions; the existing Akule--xECBS subset also solves 20/20. These outcomes show that useful learned proposals do not remove dependence on the repair backend. Akule's mean physical root conflict count is 31.60 in each backend panel, whereas the native baseline's root construction depends on its planner configuration.

\subsection{DropRegion and reproducibility}
The available released DropRegion artifacts include trained network weights and model arguments, but the environment-specific trajectory dataset required by the official inference loader is absent. The loader reads dataset metadata and successful trajectory tensors to construct a four-channel limits normalizer. The saved network and diffusion-schedule tensors do not recover these data-dependent minima and maxima. Thus the available checkpoint alone does not establish the original mapping between physical position/velocity channels and normalized diffusion inputs and outputs.

A reproducible matched evaluation requires the original inference dataset or the corresponding normalization statistics and metadata. Substituting another map's limits or estimating new limits would change the pretrained model's inference protocol. We therefore omit DropRegion. This is an unavailable normalization dependency, not evidence of infeasibility or of Akule's performance on that environment.

\subsection{Method limitations}
\paragraph{Candidate scoring.} The selector evaluates directed candidates before hard admission. Its pointwise descriptors and contextual processing therefore remain quadratic in robot count, even when expensive pairwise corrections are evaluated sparsely. Savings in $R_\phi$ evaluations do not imply a linear-time complete sampler or planner.

\paragraph{Restricted correction span.} The composer combines frozen pairwise correction directions through a finite temporal basis. It cannot express every joint multi-robot update, and signed coefficients do not remove this representational restriction. Our observed sign statistics describe the learned solution; a retrained nonnegative model would be needed to assess the necessity of signed weights.

\paragraph{Offline optimization and transfer.} Training depends on optimized, qualified pair and composition targets. Their cost and acceptance criteria affect the available supervision, and several supports can meet the same physical objectives. There is no uniquely correct interaction graph. Transfer to different populations and maps depends on the prior, normalization, geometry, and training contexts; the evaluated grids do not establish arbitrary out-of-distribution performance.

\paragraph{Repair and validation.} Interaction-aware proposals can retain robot--robot or robot--obstacle conflicts and still require static cleanup and hard search. The map measurements show that a faster batched proposal need not reduce subsequent search effort relative to a soft-guided MMD root. Complete planning also remains sensitive to backend choice and time budgets. Finally, sampled success and ideal-tracking trajectory quality do not certify continuous-time safety, dynamic feasibility under unmodeled robot physics, or execution robustness.


%% file: supplement/tables/smoothing.tex
\begin{table}[!htbp]
\centering\caption{Saved-path smoothing on canonical Weave. Cells show before/after values. Native success is unchanged; rejected smoothing candidates retain the original trajectory. Plan includes the recorded smoothing overhead after processing.}
\label{tab:smoothing}\akStyle
\akFit{\begin{tabular}{r r r r r r r}
\akH{$N$}&\akH{Accepted}&\akH{Success}&\akH{Plan (s)}&\akH{Length}&\akH{Accel.}&\akH{Jerk}\\\akAccentLine
\akOurs 10&118/512&512/512&3.128/3.131&1.811/1.808&1.593/1.525&33.54/31.79\\\akLine
\akOurs 20&103/256&256/256&7.106/7.113&1.910/1.887&2.447/2.112&54.19/45.74\\\akLine
\akOurs 28&200/512&512/512&29.944/29.957&1.949/1.933&2.813/2.573&62.50/56.44\\\akAccentLine
\end{tabular}}
\end{table}

%% file: supplement/tables/backends.tex
\begin{table*}[t]
\caption{Alternative repair backends on the first 20 frozen Weave $N=28$ test conditions. H denotes existing frozen records and N a new RTX 5090 run; cohort timings are not a matched-hardware speed comparison. Latency includes all attempts (240 s for a timeout); root conflicts count physical pair-times. CT is mean constraint-tree expansions. Path length (m) and acceleration RMS (m/s$^2$) average successful cases only; dashes denote no successful cases. Source: compact per-attempt backend records and their source hashes.}
\label{tab:backends}\centering\akStyle
\akFit{\begin{tabular}{l l c r r r r r r r}
\akH{Backend}&\akH{Proposal}&\akH{Cohort}&\akH{Success}&\akH{Timeouts}&\akH{Latency (s)}&\akH{Root conflicts}&\akH{CT}&\akH{Length}&\akH{Accel.}\\\akAccentLine
CBS & MMD & H & 0/20 & 20 & 240.00 & 124.40 & 68.20 & -- & -- \\
\akLine
\akOurs CBS & \akM{Akule} & H & 0/20 & 20 & 240.00 & 31.60 & 90.05 & -- & -- \\
\akLine
ECBS & MMD & H & 19/20 & 1 & 167.96 & 80.30 & 51.35 & 1.920 & 2.164 \\
\akLine
\akOurs ECBS & \akM{Akule} & N & 20/20 & 0 & 77.45 & 31.60 & 28.75 & 1.977 & 2.897 \\
\akLine
xECBS & MMD & H & 20/20 & 0 & 84.73 & 80.30 & 41.75 & 1.890 & 2.215 \\
\akLine
\akOurs xECBS & \akM{Akule} & H & 20/20 & 0 & 28.90 & 31.60 & 23.45 & 1.954 & 2.873 \\
\akAccentLine
\end{tabular}}
\end{table*}